\documentclass[12pt]{amsart}
\usepackage{amsbsy,amssymb,amsmath,amsthm,amscd,amsfonts,latexsym,amstext,delarray,
amsmath,graphicx,url} 
\usepackage{tikz}
\usepackage{tikz-qtree}
\usepackage{qtree}
\usepackage[margin=.8in]{geometry}
\usepackage{color}
\usepackage[all]{xy}

\newtheorem{thm}{Theorem}[section]
\newtheorem{prop}[thm]{Proposition}

\newtheorem{lem}[thm]{Lemma}

\newtheorem{defn}[thm]{Definition}
\newtheorem{rem}[thm]{Remark}

\usepackage[normalem]{ulem}

\usepackage[]{mdframed}

\usepackage{bigstrut}

\numberwithin{equation}{section}

\def\cB{{\mathcal B}}
\def\cC{{\mathcal C}}

\def\cF{{\mathcal F}}

\def\cH{{\mathcal H}}
\def\cI{{\mathcal I}}

\def\cK{{\mathcal K}}
\def\cL{{\mathcal L}}
\def\cM{{\mathcal M}}

\def\cO{{\mathcal O}}
\def\cP{{\mathcal P}}

\def\cR{{\mathcal R}}
\def\cS{{\mathcal S}}
\def\cT{{\mathcal T}}

\def\cV{{\mathcal V}}

\def\bB{{\mathbb B}}

\def\bT{{\mathbb T}}

\def\fM{{\mathfrak M}}
\def\fT{{\mathfrak T}}
\def\fF{{\mathfrak F}}

\def\fH{{\mathfrak H}}
\def\fc{{\mathfrak c}}

\def\fh{{\mathfrak h}}
\def\fz{{\mathfrak z}}
\def\fs{{\mathfrak s}}

\def\fO{{\mathfrak O}}
\def\fm{{\mathfrak m}}

\title[Hypermagmas \& Colored Operads, Phases \& Theta]{Hypermagmas and Colored Operads: Heads, Phases, and Theta Roles}
\author[M.Marcolli, M.A.C.Huijbregts, R.K.Larson]{Matilde Marcolli, Riny Huijbregts, Richard K.~ Larson}
\date{2025}

\address{Department of Mathematics and Department of Computing and Mathematical Sciences, 
California Institute of Technology, CA 91125, USA}
\email{matilde@caltech.edu}
\address{Department of Linguistics, Utrecht University, Trans 10, 3512 JK Utrecht, The Netherlands}
\email{m.a.c.huijbregts@uu.nl} 
\address{Department of Linguistics, Stony Brook University, NY 11794-4376, USA}
\email{richard.larson@stonybrook.edu}

\begin{document}
\maketitle

\begin{abstract}
We show that head functions on syntactic objects extend the magma structure to a hypermagma, 
with the c-command relation compatible with the magma operation and the m-command relation
with the hypermagma. We then show that the structure of head and complement and specifier,
additional modifier positions, and the structure of phases in the Extended Projection 
can be formulated as a bud generating system of a colored operad, in a form similar to
the structure of theta roles. We also show that, due to the special form of the colored
operad generators, the filtering of freely generated syntactic objects by these coloring rules
can be equivalently formulated as a filtering in the course of structure formation via a colored 
Merge, which can in turn be related to the hypermagma structure. The rules on movement by
Internal Merge with respect to phases, the Extended Projection Principle, Empty Category
Principle, and Phase Impenetrability Condition are all subsumed into the form of the
colored operad generators. 
Movement compatibilities between the phase structure and the theta roles assignments can then
be formulated in terms of the respective colored operads and a transduction of colored operads.
\end{abstract}

\tableofcontents

\section{Introduction}

This paper, in combination with \cite{MarLar}, is aimed at providing a mathematical
model of the filtering of the freely formed syntactic objects generated by the action
of the free symmetric Merge on workspaces, by viability for interpretation at the
interfaces (Syntax-Semantic interface and Externalization). While \cite{MarLar}
focused on filtering on the bases of correct assignment of theta roles and the
theta criterion, as well as the dichotomy between External and Internal Merge
in their roles with respect to theta theory, in this paper we focus on filtering by
well formed head and complement structure and phases. We also analyze the
relation between the filtering by phases and the filtering by theta theory. 
As in the case of theta theory, we show that the filtering can equivalently
be interpreted as simultaneous to structure formation, in the formed of a
``colored Merge" that performs structure building and filtering at each step
rather than first unconstrained structure formation followed by filtering. Since
these two processes are in fact equivalent, adopting one or the other viewpoint
is more a matter of convenience (decoupling formation and filtering for deriving
formal arguments in abstract modeling, combining them for algorithmic optimization 
avoiding combinatorial explosions). As in the case of theta theory, the filtering
for phase structure is governed by the same algebraic formalism involving
colored operads, bud generating systems, and algebras over operads. To show
some specific examples of non-trivial interaction between the filtering by
theta theory and the filtering by phases we analyze the cases of passivization,
of exceptional case marking, and of double object constructions. We 
propose a formalism, based on a notion of transduction (or correspondence)
of colored operads, for the combined implementation of filtering by
theta roles and by phase structure. 

\smallskip

We start by considering the algebraic properties describing the
structure of head and complement, and phases, starting from the
formulation discussed in Chapter~1 of \cite{MCB}. We show that, when
one incorporates the structure of head, the magma of syntactic objects 
needs to be extended to a hypermagma that captures the behavior of
the syntactic head. This shows, for example, that while the c-command relation
is fully describable in terms of the magma of syntactic objects, the m-command
relation requires the hypermagma structure. 

We then show that the structure of head and
complement, and phases, can be formulated in terms of
coloring rules, and a bud generating system for a colored operad,
in much the same way as we have done for theta roles assignments in \cite{MarLar}.
This determines a coloring algorithm for the full abstract binary rooted trees of
syntactic objects that assigns the phase structure. 

We then compare the coloring by phases with the coloring by theta roles.
The compatibility between the two imposes restrictions on certain kinds
of movement explanations for theta role configurations, and generally for
the behavior of theta roles in relation to movement. 

A note for the readers: the initial part, in \S \ref{HyperHSec}, on the hypermagma
structure clarifies and provides further details about the properties of the head function 
discussed in \S 1.13 of \cite{MCB}. It is used again in this paper only in \S \ref{HypColorMergeSec}.
Thus, \S \ref{HyperHSec}, the implication on command structures in \S \ref{CommandSec}, and the
last part of \S \ref{HypColorMergeSec} can be skipped by the
reader interested in the linguistics applications to phases and labeling algorithms, 
more than in the details of the mathematical structures involved.  (However, note that
Definition~\ref{headfuncDef1} in \S \ref{HyperHSec} is used in the rest of the paper.)

\section{The algebraic structure of syntactic objects and head functions}\label{HyperHSec}

In this section we recall some notions of hyperstructures (in particular
the notion of hypermagma) associated to algebraic structure (in this
case the notion of magma). We recall how syntactic objects are
obtained as the free non-associative commutative magma over the
set of lexical items and syntactic features. We then show that, when
one takes into consideration also the presence of a syntactic head,
whose properties are encoded in the abstract notion of a head function,
this naturally leads from the magma to a hypermagma structure.

\subsection{Magmas and hypermagmas} 

Hyperstructures are generalizations of algebraic structures where
the operations are set valued, \cite{NakaRey}. In particular, we are interested here
in the notion of hypermagma, \cite{Marty}, \cite{Mit}, \cite{Smith}. 

\begin{defn}\label{hypmagmadef} {\rm
For a given set $X$, let $\cP(X)=2^X$
be the powerset of $X$. Then a hyperoperation on $X$ is a function
\begin{equation}\label{hyperop}
 \star : X \times X \to \cP(X) \, . 
\end{equation} 
Any such hyperoperation extends to an operation on the powerset
$$ \star: \cP(X) \times \cP(X) \to \cP(X) $$
by setting, for $A,B \subset X$, 
$$ A \star B =\bigcup_{(a,b)\in A\times B} a \star b \, . $$
A hyperoperation is {\em total} if \eqref{hyperop} takes values
in $\cP(X)^o \subset \cP(X)$, the set of non-empty subsets of $X$. 
It then extends to an operation on $\cP(X)^o$.
A pair $(X,\star)$ of a set $X$ together with a hyperoperation 
as in \eqref{hyperop} is called a {\em hypermagma}. A
{\em magma} is a {\em hypermagma} where $a\star b$ is
always a singleton, for all $a,b \in X$, namely \eqref{hyperop}
is an actual binary operation on $X$,
$$ \star : X \times X \to X \, . $$
 }\end{defn}
 
 The magma operation and the hypermagma hyperoperation
 are in general not associative: when associativity holds the
 resulting structure is referred to as a monoid, or hypermonoid. 
 
 \begin{rem}\label{hypmagcat} {\rm
 Hypermagmas form a category $\cH\cM$ with morphisms $\phi: (X,\star_X) \to (Y,\star_Y)$
 given by functions $\phi: X\to Y$ satisfying the {\em colax} condition
 $$ \phi( a\star_X b) \subseteq \phi(a) \star_Y \phi(b)\, . $$
 A strict morphism is one that also satisfies the {\em lax} condition
 $\phi( a\star_X b) \supseteq \phi(a) \star_Y \phi(b)$, so that  
 $\phi( a\star_X b) = \phi(a) \star_Y \phi(b)$.  }
 \end{rem}
 
 A hypermagma $(X,\star)$ is unital if there is a (unique) element $1\in X$ 
 such that $1\star x = x\star 1 = \{ x  \}$. Morphisms of unital hypermagmas
 also map the unit to the unit.
 
 We also recall some definitions form \cite{NakaRey} that will be
 useful in the following.
 
 \begin{defn}\label{short}{\rm 
 A morphism $\phi: (X,\star_X) \to (Y,\star_Y)$ of hypermagmas 
 is said to be {\em short} if it is surjective and 
 $$  \phi(\phi^{-1}(a)\star \phi^{-1}(b))= a\star b \,, \ \ \  \forall \, a,b\in Y \, ,  $$
 and  is said to be {\em coshort} if it is injective and 
$$ u \star v = \phi^{-1}(\phi(u)\star \phi(v))   \,  , \ \ \  \forall \, u,v \in X \, . $$
Given a hypermagma $(X,\star_X)$, a {\em strict subhypermagma} $(Y, \star_X)$
is a subset $Y\subset X$ such that $a\star_X b \subseteq Y$, for all $a,b\in Y$.
A {\em weak subhypermagma} $(Y, \star_Y)$ of $(X,\star_X)$ is a subset 
$Y\subset X$ with a hyperoperation $\star_Y$ satisfying 
\begin{equation}\label{starYweak}
a \star_Y b = (a\star_X b) \cap Y \, . 
\end{equation}
A coshort morphism is the same as the inclusion of its image as a weak subhypermagma.  
Any subset  $Y\subset X$ has an induced hyperoperation defined by the right-hand-side of
\eqref{starYweak} that makes is a weak subhypermagma. }
\end{defn} 

Note that, even if the hypermagma $(X,\star_X)$  is total, weak subhypermagmas need not
be total, as the intersections $(a\star_X b) \cap Y$ can be empty even if $a\star_X b$ is
never empty. 

\subsection{The magma of syntactic objects}

As shown in \cite{MCB}, syntactic objects can be described as the elements
of the free non-associative commutative magma 
$$ \cS\cO={\rm Magma}_{c,na}(\cS\cO_0,\fM) $$
generated by a finite set $\cS\cO_0$ of lexical items and syntactic features. 
This means that syntactic objects are obtained, starting with the elements of 
$\cS\cO_0$ by repeated applications of a single non-associative, commutative
binary operation $\fM$. Thus, one recursively obtains increasingly complex syntactic objects,
such as $\fM(\alpha,\beta)$, $\fM(\fM(\alpha,\beta), \gamma)$, 
$\fM(\delta, \fM(\alpha, \fM(\beta, \gamma)))$, etc.
The magma operation $\fM: \cS\cO\times \cS\cO\to \cS\cO$ acts as
$(T,T')\mapsto \fM(T,T')$. 

\smallskip 

It is a well known fact that the set obtained as the
free non-associative commutative magma on a given finite set $S$
is canonically isomorphic to the set $\fT_S$ of non-planar binary rooted trees
with leaves labelled by elements of the set $S$. As in \cite{MCB}
when we say binary rooted trees we always mean {\em full} binary rooted trees
(meaning without non-branching vertices: every non-leaf vertex has
exactly two descendant vertices). The fact that the
trees are non-planar (without an assigned planar embedding)
corresponds to the commutativity of the magma operation, while
the non-associativity is responsible for the non-trivial tree structures.
Thus, here as in \cite{MCB}, we will also use the notation
$\fT_{\cS\cO_0}$ to denote the set of syntactic objects,
endowed with its magma structure. The magma operation, in this form is given by
$$ (T,T')\mapsto \fM(T,T') = \Tree[ $T$ $T'$ ] \, , $$
where one should keep in mind that these trees and non-planar, namely  
$$ \Tree[ $T$ $T'$ ]=  \fM(T,T')=\fM(T',T) =\Tree[ $T'$ $T$ ]  \, . $$ 

\smallskip

We will always assume that a binary rooted tree $T$ is oriented downward from the root, 
so each non-leaf vertex $v$ has two edges with source $v$ (the two edges just below). 
Thus, we will call successor and preceding vertices or edges, or above and below,
with respect to this orientation, with the root at the top. 

\smallskip

\subsection{Head functions and hypermagmas}

A head function on an abstract binary rooted tree is defined in the following way
(Definition 1.13.3 of \cite{MCB}).

\begin{defn}\label{headfuncDef1}{\rm
A {\em head function} on an abstract binary rooted tree $T$
is a function $h_T: V^o(T)\to L(T)$ from the set $V^o(T)$
of non-leaf vertices of $T$ to the set $L(T)$ of leaves of $T$,
with the property that if $T_v \subseteq T_w$ and
$h_T(w)\in L(T_v)\subseteq L(T_w)$, then $h_T(w)=h_T(v)$.}
\end{defn}

It is proved in Lemma~1.13.7 of \cite{MCB} that the
description of head functions given in Definition~\ref{headfuncDef1}
is equivalent to the description given in \cite{Chomsky-bare}. 
As shown in Lemmas~1.13.4 and 1.13.5 of \cite{MCB} there are exactly
$2^{n-1}$ possible head functions on an abstract binary rooted tree $T$
with $n$ leaves. They correspond to the $2^{n-1}$ different possible
planar embeddings of the tree.  This means that the characterization
given in Definition~\ref{headfuncDef1} has $2^{\# L(T)-1}$ possible
solutions. 

\smallskip

Among all the $2^{\# L(T)-1}$ possible abstract head function, 
one is interested in particular in 
the head function that corresponds to the actual syntactic head, which 
depends on the specific lexical items and syntactic features assigned
to the leaves of $T$. This, however, is not always well defined, as
the structure $T$ may contain exocentric constructions that are not
label-able, hence do not have a well determined syntactic head.
The existence of exocentric constructions is a manifestation of
the fact that the datum of a head function is not directly compatible
with the magma operation on syntactic objects. 
This incompatibility was observed in \cite{MCB}, where in order
to model the syntactic head, one introduces an assignment
of head functions $h: T \mapsto h_T$ that is in general only 
defined on a domain ${\rm Dom}(h) \subset \fT_{\cS\cO_0}$
that is {\em not} a sub-magma. We discuss here more in
detail the relation between head functions and the magma
structure. 

\smallskip

\begin{lem}\label{Hnomag}
Let $\fH_{\cS\cO_0}$ denote the set of pairs $(T, h_T)$ of
a syntactic object $T\in \fT_{\cS\cO_0}$ and a head function
$h_T$ on $T$ as in Definition~\ref{headfuncDef1}. The
magma operation $\fM$ on $\fT_{\cS\cO_0}$ does not
induce a compatible magma operation on $\fH_{\cS\cO_0}$,
but it induces a compatible commutative {\em hypermagma} 
structure defined by the hypermagma operation
\begin{equation}\label{Mhyper}
 \fM^\fH ((T,h_T), (T',h_{T'})) =( \fM(T,T'), \{ h_T, h_{T'} \} ) \, .
\end{equation}
\end{lem}

\proof
Given two pairs $(T, h_T)$ and $(T', h_{T'})$ 
of a syntactic object and a head function, there is no unique
way of assigning a head function $h_{\fM(T,T')}$ to $\fM(T,T')$ 
determined by $h_T$ and $h_{T'}$. Indeed, there are two
possible choices of how to extend $h_T$ and $h_{T'}$ to a
head function $h_{\fM(T,T')}$, by assigning to the new root
vertex $v$ of $\fM(T,T')$ either the value $h_{\fM(T,T')}(v)=h_T(v_T)$
or $h_{\fM(T,T')}(v)=h_{T'}(v_{T'})$, where $v_T$ and $v_{T'}$
are, respectively, the root vertex of $T$ and of $T'$. The data of
$(T, h_T)$ and $(T', h_{T'})$ do not suffice in general to determine one or
the other choice at the new root $v$. However, there is a 
hypermagma structure on the set $\fH_{\cS\cO_0}$, which
we denote by $\fM^\fH$, that is defined as in \eqref{Mhyper},
where the notation  $(\fM(T,T'), \{ h_T, h_{T'} \} )$ stands for
the set
$$ \{ ( \fM(T,T'), h_T ), ( \fM(T,T'), h_{T'} ) \} $$
where the pair $(\fM(T,T'), h_T)$ (respectively, $( \fM(T,T'), h_{T'} )$) 
consists of  $\fM(T,T")$, the syntactic object obtained by multiplication in the magma,
endowed with the head function $h_{\fM(T,T')}$
with $h_{\fM(T,T')}(v)=h_T(v_T)$ (respectively, $h_{\fM(T,T')}(v)=h_{T'}(v_{T'})$). 
The hypermagma is commutative, since $\fM(T,T')=\fM(T',T)$ and the set
$\{ h_T, h_{T'} \}$ is also invariant under exchanging $T$ and $T'$.
\endproof 

\smallskip

The relation between the magma $\fT_{\cS\cO_0}$ and the
hypermagma $\fH_{\cS\cO_0}$ is described as follows.

\begin{defn}\label{projhypermag} {\rm
Given two hypermagmas $(X,\star_X)$ and $(Y,\star_Y)$, a projection
$\pi: (Y,\star_Y) \to (X,\star_X)$ is a surjection $\pi: Y\to X$ that is a short morphism of hypermagmas. 
A section $\sigma$ of the projection $\pi$ is a coshort morphism of 
hypermagmas $\sigma: (X,\star_X) \to (Y,\star_Y)$, satisfying $\pi \circ \sigma ={\rm id}_X$. 
A partial section is a section $\sigma$ defined on a subset $D(\sigma)\subset X$, 
with the induced weak subhypermagma structure $(D(\sigma), \star_\sigma)$. 
}\end{defn}

Using the hypermagma formalism, we can refine our understanding of the
properties of the domain ${\rm Dom}(h) \subset \fT_{\cS\cO_0}$ of a
head function, discussed in \cite{MCB}.  

\begin{prop}\label{maghypmag} 
There is a short morphism of hypermagmas $\pi: \fH_{\cS\cO_0} \to \fT_{\cS\cO_0}$
that forgets the datum of the head function. The assignment of a head function $h: T \mapsto h_T$
defined on a domain ${\rm Dom}(h)$ that is a subset of $\fT_{\cS\cO_0}$ determines a
weak subhypermagma of $\fH_{\cS\cO_0}$ and a coshort hypermagma morphism
$\sigma_h : ({\rm Dom}(h), \fM_h) \to (\fH_{\cS\cO_0}, \fM^\fH)$ that is a partial section of $\pi$,
where $\fM_h$ is the induced hypermagma structure on ${\rm Dom}(h)$.
\end{prop} 

\proof Here we view the magma $(\fT_{\cS\cO_0}, \fM)$ as a hypermagma where the
hyperoperation is singleton-valued. Note that, even if $(\fT_{\cS\cO_0}, \fM)$ is
an actual magma, subsets of $\fT_{\cS\cO_0}$ will have an induced structure of
weak subhypermagma, where the hyperoperation is set values, because we have
to include the possibility that the emptyset is a value. In particular, whenever the chosen
subset  of $\fT_{\cS\cO_0}$ is not a submagma, the induced structure will include the
emptyset as a possible value of the hyperoperation. Only in the case of a submagma
the induced weak subhypermagma structure is itself a magma structure. Consider then
the case of a subset ${\rm Dom}(h) \subset \fT_{\cS\cO_0}$ that is the domain of
a head function $h: T \mapsto h_T$. We know from \cite{MCB}, \S 1.13.2, that 
${\rm Dom}(h)$ is not a submagma. Let $\fM_h$ denote the induced 
weak subhypermagma structure on ${\rm Dom}(h)$. This has the form (for $T,T'\in {\rm Dom}(h)$):
$$ \fM_h(T,T') = \left\{ \begin{array}{ll} \fM(T,T') & \fM(T,T') \in {\rm Dom}(h) \\ \emptyset &
\fM(T,T') \notin {\rm Dom}(h)\, . \end{array} \right. $$
This is clearly also compatibly describing ${\rm Dom}(h)$ as a weak subhypermagma of $\fH_{\cS\cO_0}$ with
\begin{equation}\label{hsubhyp}
 \fM^\fH((T,h_T), (T',h_T')) = \left\{ \begin{array}{ll} (\fM(T,T'), h_{\fM(T,T')}) & \fM(T,T') \in {\rm Dom}(h) \\ \emptyset &
\fM(T,T') \notin {\rm Dom}(h)\, . \end{array} \right. 
\end{equation}
Note that this agrees with the weak subhypermagma relation \eqref{starYweak}, since we know
that, if $\fM(T,T') \in {\rm Dom}(h)$ then by the properties of head functions we have either
$h_{\fM(T,T')}=h_T$ or $h_{\fM(T,T')}=h_{T'}$, in the sense discussed above, of being the unique
extension of $h_T$ and $h_{T'}$ to $\fM(T,T')$ with value at the root agreeing with either $h_T$ or
$h_{T'}$ at the respective roots of $T$ and $T'$, so that
$$ (\fM(T,T'), h_{\fM(T,T')}) = (\fM(T,T'), \{ h_T, h_{T'}  \}) \cap {\rm Dom}(h) \, . $$ 
This embedding $\sigma_h: {\rm Dom}(h) \hookrightarrow \fH_{\cS\cO_0}$ given by
$\sigma_h(T)=(T,h_T)$ is then a coshort morphism
and it clearly satisfies $\pi \circ \sigma_h ={\rm id}$. 
\endproof

\begin{rem}\label{hchoice}{\rm 
Thus, one can think of a choice of a head function $h: T \mapsto h_T$ as a choice of
a weak subhypermagma of $\fH_{\cS\cO_0}$ with the subhypermagma structure of
the form \eqref{hsubhyp}. }
\end{rem}

We will return to discuss the hypermagma structure in \S \ref{HypColorMergeSec}.

\section{Command relations}\label{CommandSec}

In this section we compare the c-command and m-command relations
used on syntax. We show that the c-command relation is compatible with
the magma structure of the set of syntactic objects, while the m-command
relation, which depends not only on the syntactic object but also on the
head function, is not compatible with the magma operation but becomes
compatible with the hypermagma structure introduced in the previous
section, which accounts for the properties of the head function.

\subsection{c-command and m-command}

Also, recall from \cite{MCB} that, for $T\in \fT_{\cS\cO_0}$ a syntactic object,
the set ${\rm Acc}(T)$ of accessible terms of $T$ consists of all the subtrees
$T_v$ of $T$ for $v$ a non-root vertex of $T$, where $T_v$  denotes the
subtree rooted at $v$, and containing everything that lies below $v$. 
We also write ${\rm Acc}'(T)={\rm Acc}(T)\cup \{ T \}$. 

\smallskip

\begin{defn}\label{cmcommdef}{\rm For a syntactic object $T\in \fT_{\cS\cO_0}$, 
two vertices, $v_1$ and $v_2$, of $T$ are
sisters if there is a vertex $v$ of $T$ above them (i.e., closer to the root) and
connected to both $v_1$ and $v_2$.
\begin{itemize}
\item A vertex $v$ of $T$ dominates another vertex $w$ if $v$ is
on the unique path from the root of $T$ to $w$.
\item A vertex $v$ in $T$ c-commands  another vertex $w$ if
neither dominates the other and the lowest vertex that
dominates $v$ also dominates $w$.  (Equivalently, $w\in T_{v'}$, the accessible term rooted at the sister vertex $v'$ of $v$.)
\item A vertex $v$ asymmetrically
c-commands a vertex $w$ if $v$ c-commands $w$, and $v$ and $w$ are
not sisters. (Equivalently, $T_w\in {\rm Acc}(T_{v'})$, the set of accessible terms of $T_{v'}$, with $v'$ the sister vertex of $v$.)
\item For a pair $(T,h_T)$ of a syntactic object $T$ and a head function $h_T$, 
a maximal projection is a vertex $v$ with the property that $T_v$ is not
strictly contained in any larger $T_w\in {\rm Acc}'(T)$ with $h_T(w)=h_T(v)$.
\item A vertex $v$ in $(T,h_T)$ m-commands another vertex $w$ if  
neither dominates the other and the maximal projection of $v$ dominates $w$.
\end{itemize}
The c-command relation only depends on $T$, while the m-command relation
depends on both $T$ and $h_T$, as the maximal projection is determined by
the head function.   }
\end{defn}

\smallskip

\begin{rem}\label{gammaell}{\rm 
It is shown in \cite{MCB}, Lemma~1.14.1, that a head function $h_T$ on a 
syntactic object $T\in \fT_{\cS\cO_0}$ determines a partition of the vertices of $T$
into a collection of paths (some of them trivially consisting of a single leaf), 
$\{ \gamma_\ell \}_{\ell \in L(T)}$ for $L(T)$ the set of leaves of $T$, where
$v\in \gamma_\ell$ iff $h_T(v)=\ell$.  The internal vertex $v_\ell$ where the path 
$\gamma_\ell$ (if nontrivial) terminates gives the maximal projection of the head $\ell$.
The resulting accessible term $T_{v_\ell}$ will also include the specifier position.
In Definition~\ref{PhaseOp}, we will refer to $T_{v_\ell}$ as the ``narrow phase" of $\ell$
and we will discuss the setting of projection and phases more precisely. }
\end{rem}

\smallskip

\subsection{c-command and the magma structure}

We show that the c-command relation on syntactic objects is compatible with the magma structure
of $\fT_{\cS\cO_0}$. 

\smallskip

Let ${\rm Rel}$ denote the category where objects are pairs $(S, R)$ of a set
and a relation $R \subset S\times S$, and with morphisms
$f: (S,R)\to (S',R')$ that are functions of sets $f: S \to S'$ that 
intertwine the relations, namely for all $(x,y)\in R$, we have
$(f(x),f(y))\in R'$. Let $\cF\cS$ denote the category of finite sets.

\begin{defn}\label{magmaRel}{\rm 
Suppose given a magma $(X,\star_X)$ and a map $\rho: X \to \cF\cS$ to finite sets
that assigns a set $\rho(x)=S_x$ to each element of the magma. Given a collection $\cR$
of relations $R_x\subseteq S_x\times S_x$, let $\cS_{X,\cR}=\{ (S_x, R_x) \in {\rm Rel} \}$,
and let $\tilde\rho: X \to {\rm Rel}$ be given by $\tilde\rho(x)=(S_x, R_x)$.
The collection $\cR=\{ R_x\subseteq S_x\times S_x\,|\, x\in X \}$ is {\em magmatic} 
if the operation $\star_X$ determines a magma structure 
$\star_{X,\rho} :  \cS_{X,\cR} \times \cS_{X,\cR} \to \cS_{X,\cR}$, with
$\tilde\rho(x)\star_{X,\rho} \tilde\rho(y)=\tilde\rho(x \star_X y)$.  In the case of a
hypermagma $(H,\star_H)$, and a map $\rho: H \to \cF\cS$ with $\rho(x)=S_x$,
a collection $\cR=\{ R_x\subseteq S_x\times S_x\,|\, x\in H \}$ is  {\em hypermagmatic} 
if $\star_H$ determines a hypermagma structure 
$\star_{H,\rho} :  \cS_{H,\cR} \times \cS_{H,\cR} \to \cP(\cS_{H,\cR})$ on
$\cS_{H,\cR}=\{ (S_x, R_x) \in {\rm Rel} \}$, with
$\tilde\rho(x)\star_{H,\rho} \tilde\rho(y)\supseteq \tilde\rho(x \star_H y)$,
where $\tilde\rho(x)=(S_x,R_x)$. It is {\em strictly hypermagmatic} if
$\tilde\rho(x)\star_{H,\rho} \tilde\rho(y) = \tilde\rho(x \star_H y)$. 
} \end{defn}

\smallskip

\begin{lem}\label{cmagma}
The c-command relation is magmatic with the respect to the magma $\fT_{\cS\cO_0}$ of
syntactic objects and the map $V: T \mapsto V(T)$, the set of vertices of $T$.
\end{lem}

\proof We need to show that the magma operation $\fM$ on syntactic objects induces a
magma operation $\fM_V$ on the set $\cS_{\fT_{\cS\cO_0},\cR_c}$, where $\cR_c$ is
the c-command relation $R_{c,T} \subset V(T)\times V(T)$ with 
$R_{c,T}=\{ (v,w) \,|\, v \text{ c-commands } w \}$. The set  $\cS_{\fT_{\cS\cO_0},\cR_c}$
consists of pairs $(V(T), R_{c,T})$ in ${\rm Rel}$. To see that the magma operation $\fM$ on
$\fT_{\cS\cO_0}$ induced a magma operation $\fM_V$ on $\cS_{\fT_{\cS\cO_0},\cR_c}$ we
need to show that we can define $\fM_V ( (V(T), R_{c,T}), (V(T'), R_{c,T'}) )$ 
in terms of $V(T)$, $V(T')$ and $R_{c,T}$, $R_{c,T'}$, in such a way that it satisfies
\begin{equation}\label{MVmag}
 \fM_V ( (V(T), R_{c,T}), (V(T'), R_{c,T'})) = (V(\fM(T,T')), R_{c,\fM(T,T')} )\, . 
\end{equation}  
This relation shows that the magma operation $\fM_V$ on $\cS_{\fT_{\cS\cO_0},\cR_c}$
should be of the form
$$ \fM_V ( (V(T), R_{c,T}), (V(T'), R_{c,T'})) =( \fM_V ( V(T),  V(T') ) , \fM_V ( R_{c,T}, R_{c,T'} ) ) \, , $$
with $\fM_V ( V(T),  V(T') ) =V(\fM(T,T'))$ and $\fM_V ( R_{c,T}, R_{c,T'} ) )=R_{c,\fM(T,T')}$.
To express $V(\fM(T,T'))$ in terms of $V(T)$ and $V(T')$ and $R_{c,\fM(T,T')}$
in terms of $R_{c,T}$ and $R_{c,T'}$, note that 
$$ V(\fM(T,T'))  = V(T) \sqcup V(T') \sqcup \{ v_{\fM(T,T')} \}\, , $$
with $v_{\fM(T,T')}$ the root vertex of $\fM(T,T')$, and 
$$ R_{c,\fM(T,T')} = R_{c,T} \sqcup R_{c,T'} \sqcup V(T)\times \{ v_{T'} \}
\sqcup \{ v_{T'} \}\times V(T) \sqcup V(T')\times \{ v_T \}
\sqcup \{ v_T \}\times V(T') \, . $$ 
This shows that the c-control relation $R_{c,\fM(T,T')}$ on $\fM(T,T')$ is uniquely determined as a function
$\fM_V ( R_{c,T}, R_{c,T'} )$ of $R_{c,T}$ and $R_{c,T'}$, which satisfies  \eqref{MVmag}. 
\endproof

\subsection{m-command and hypermagmas} 

Unlike the c-command relation, the m-command relation is not magmatic with respect to 
the magma $\fT_{\cS\cO_0}$. Indeed, first notice that, since the m-command relation requires
a head function, it can only be defined on a pair $(T,h_T)$. So we write
$R_{m,(T,h_T)} \subset V(T)\times V(T)$ for the relation 
$$ R_{m,(T,h_T)} =\{ (v,w) \,|\, v \text{ m-commands } w \} \, , $$
where $v$ m-commands $w$ with respect to the maximal projection induced by the head function $h_T$. 
So we can assume given a choice of a head function $h: T \mapsto h_T$ defined on a domain
${\rm Dom}(h)\subset \fT_{\cS\cO_0}$, with respect to which the m-command is computed. 
For $T,T'\in {\rm Dom}(h)$, we then have the m-command relations $R_{m,(T,h_T)}\subset V(T)\times V(T)$ 
and $R_{m,(T',h_{T'})}\subset V(T')\times V(T')$. The syntactic object $\fM(T,T')$ need not be in ${\rm Dom}(h)$.
When $\fM(T,T')\notin {\rm Dom}(h)$, the m-command relation is not defined on $\fM(T,T')$.
Thus, we see that m-command is not magmatic for the magma $\fT_{\cS\cO_0}$.

\begin{prop}\label{mcomhypmag}
The m-command relation is strictly hypermagmatic for the weak subhypermagma ${\rm Dom}(h)$
of $\fH_{\cS\cO_0}$. 
\end{prop} 

\proof Given a head function $h: T \mapsto h_T$  with domain ${\rm Dom}(h)$ and the hypermagma 
$({\rm Dom}(h), \fM_h)$ with $\fM_h$ induced by its embedding as weak subhypermagma of 
$\fH_{\cS\cO_0}$ as in Proposition~\ref{maghypmag}.  To see that m-command is strictly
hypermagmatic with respect to $({\rm Dom}(h), \fM_h)$, we take $\rho(T,h_T)=V(T)$ and
$\tilde\rho(T,h_T) = (V(T), R_{m, (T,h_T)} )$. 
The induced hypermagma structure $\fM_{V,h}$ on $\cS_{{\rm Dom}(h),\cR_m}=\{ (V(T), R_{m,(T,h_T)} \,|\, T\in {\rm Dom}(h) \}$
should satisfy
$$ \fM_{V,h}((V(T),R_{m,(T,h_T)}), (V(T'),R_{m,(T',h_{T'})}) = ( V(\fM(T,T')), R_{m,(\fM(T,T'),h_{\fM(T,T')) }} ) $$
if $\fM(T,T') \in {\rm Dom}(h)$ and $\fM_{V,h}((V(T),R_{m,(T,h_T)}), (V(T'),R_{m,(T',h_{T'})}) =
\emptyset$ if $\fM(T,T') \notin {\rm Dom}(h)$. This gives 
$$ \fM_{V,h}((V(T),R_{m,(T,h_T)}), (V(T'),R_{m,(T',h_{T'})}) =(\fM_{V,h}(V(T),V(T')), 
\fM_{V,h}(R_{m,(T,h_T)}, R_{m,(T',h_{T'})})\, , $$
with 
$$ \fM_{V,h}(V(T),V(T')) = V(\fM(T,T')) = V(T) \sqcup V(T') \sqcup \{ v_{\fM(T,T')} \}\, , $$
as in the c-command case, with $v_{\fM(T,T')}$ the root vertex of $\fM(T,T')$, and 
$\fM_{V,h}(R_{m,(T,h_T)}, R_{m,(T',h_{T'})}) =\emptyset$ if $\fM(T,T') \notin {\rm Dom}(h)$,
while for $\fM(T,T') \in {\rm Dom}(h)$, we have
$$ \fM_{V,h}(R_{m,(T,h_T)}, R_{m,(T',h_{T'})}) = R_{m,(\fM(T,T'),h_{\fM(T,T')) }}  $$
where depending on whether  $h_{\fM(T,T')} =h_T$ or $h_{\fM(T,T')} =h_{T'}$ (in the sense 
of extending either $h_T$ or $h_{T'}$ to the root of $\fM(T,T')$), we have, respectively 
$$ R_{m,(\fM(T,T'),h_{\fM(T,T')) }} = R_{m, (T,h_T)}\sqcup R_{m, (T', h_{T'})} \sqcup  \gamma_{h(v_T)} \times V(T') \, , $$
$$ R_{m,(\fM(T,T'),h_{\fM(T,T')) }} = R_{m, (T,h_T)}\sqcup R_{m, (T', h_{T'})} \sqcup  \gamma_{h(v_{T'})} \times V(T) \, , $$
where $v_T$ and $v_{T'}$ are the roots of $T$ and $T'$, respectively, and $\gamma_\ell$ are the paths
determined by the head function, as in Lemma~1.14.1 of \cite{MCB}, as recalled above. 
\endproof

\section{Complemented heads and coloring}

We now consider, in addition to the presence of a head function $h_T$ on the
syntactic object $T$, in the general sense recalled in Definition~\ref{headfuncDef1},
the finer property that this head function is
complemented, in the sense of Definition~1.14.2 of \cite{MCB}.

First observe that the assignment of a head function $h_T$ to a syntactic
object is equivalent to a bi-coloring of the edges of $T$ according to the
following rule. 

\begin{rem}\label{headbicolor}{\rm
Assigning a head function $h_T$ to a syntactic object $T$ is the same as assigning
a bi-coloring (say B/W) of the edges of $T$, with the rule that one of the edges
below each non-leaf vertex $v$ is colored B and the other is colored W (see 
Lemma~1.13.4 of \cite{MCB}), where one of the two colors (say B) marks the
direction of the head $h_T(v)$. In particular, the paths $\gamma_\ell$ mentioned
in Remark~\ref{gammaell} are the paths of B-colored edges in $T$, according to
this coloring.}
\end{rem}

We show here that assigning a complemented
head, and keeping track of the specifier-head-complement structure 
also has an interpretation in terms of coloring, with
appropriate coloring rules as discussed below.

\smallskip
\subsection{Complemented head, extended projection, and coloring}

We follow here a procedure analogous to the analysis of theta role
carried out in \cite{MarLar}. We first consider a coloring algorithm that
only accounts for the basic part of the structure, which in this case
consists of the three roles: specifier, head, and complement. We then
enrich this fundamental structure by allowing for additional modifier positions.
This is analogous to the coloring for theta roles of \cite{MarLar}, where we
first introduced what we called ``bare theta structures", and then
enlarged those to include non-theta positions. 

We start here by recalling the notion of ``complemented head"
introduced in Definition~1.14.2 of \cite{MCB}.

\begin{defn}\label{complhead}
A complemented head function $h_T$
is a function $h_T: V^o(T) \to L(T)\times ({\rm Acc}(T)\cup \{ 1 \})$ that assigns to
a non-leaf vertex $v$ the corresponding head $\ell=h_T(v)\in L(T)$, together with
an associated accessible term $Z_\ell \in {\rm Acc}(T)$ (or possibly empty $Z_\ell=1$,
the formal empty tree), that is the complement of the head $\ell$. When nonempty,
$Z_\ell \subseteq T_{s_{h_T(v)}}$, is ether equal to or contained in the 
accessible term rooted at the sister vertex $s_{h_T(v)}$ of the leaf $\ell=h_T(v)$. 
\end{defn}

Note that the last condition,
that $Z_\ell \subseteq T_{s_{h_T(v)}}$, will be relaxed, by allowing
$Z_\ell \subseteq T_v$ for some other $v\in \gamma_\ell$, when
we also introduce the possibility of additional modifier positions.

The complement $Z_\ell$ of the head $\ell$ is an accessible term of $T$,
so it is itself a syntactic object, and it is also endowed with a head function,
which is the restriction of $h_T$ to $Z_\ell$. This head function is complemented,
hence $Z_\ell$ itself recursively decomposes into substructures 
containing heads and their complements. 

We refine the structure of head and complement by also accounting for an
additional {\em specifier} position. 

To this purpose, we will need here to distinguish between {\em maximal projection}
and {\em extended projection}. In our formulation, the maximal projection of a head
$\ell$ is the endpoint $v_\ell$ of the path $\gamma_\ell$ of Remark~\ref{gammaell}.
When the head function $h_T$ is the actual syntactic head, this corresponds
to the usual notion of maximal projection: the highest level (highest vertex $v=v_\ell$)
of the phrase structure that maintains the same (lexical) head $\ell$. On the other hand, 
in the case of the syntactic head, the extended projection involves the maximal
projection $v_\ell$ of a {\em lexical head} and, above that, a certain number of 
{\em functional projections}, namely projections of {\em functional heads} such as
tense features ($\mathtt{T}$) or inflection ({\tt INFL} or $\mathtt{I}$) or complementizer ($\mathtt{C}$),
in the extended projection of $\mathtt{VP}$s or  determiner ($\mathtt{D}$) in the extended projection of $\mathtt{NP}$s.
The Extended Projection Principle states that a viable sentence has a subject 
located in {\em specifier position} of an extended projection. 
We will focus especially on the $\mathtt{C}$, {\tt INFL}, and $\mathtt{v}^*$ functional heads 
involved in the extended projection. We will also discuss cases involving heads
like $\mathtt{v}$, in the extended projection, that are not phase-heads, hence behave differently
with respect to IM movement. (In the setting of \cite{ChomskyGK}, $\mathtt{TP}$ is part of $v$
not of {\tt INFL}.)
The structure of {\em phases} and the distinction between the interior and the
edge of a phase (see the discussion in Section 1.14 of \cite{MCB}) and the rules
about movement by Internal Merge depend on
the structure of the extended projection and its functional heads, as we discuss further below.

\subsubsection{Tree structures and extended projection} 
We will use two different ways of representing the non-planar binary rooted trees
that are the syntactic objects, a simplified form where the leaves are lexical heads,
and a full form where the leaves include the functional heads that occur in the
extended projections. In the first case, one can remember the role of the functional 
projections by encoding them in the coloring. Principles of structural economy 
suggest that a simplified structure, where the extended projection may be
accounted for via some kind of coloring rules, may be preferable for both computational
and theoretical purposes. However, we will focus here on modeling the
full form of the syntactic objects where the extended projection is explicitly
included in the resulting binary rooted tree, so that we can analyze more
precisely how the usual formulation of the structure of phases can be
described in terms of coloring rules and colored operads bud generating systems. 

\smallskip

To see a simple example, consider the sentence {\em John saw Mary}. The
simplified form of the binary tree realizing this sentence as a syntactic object
is just of the form
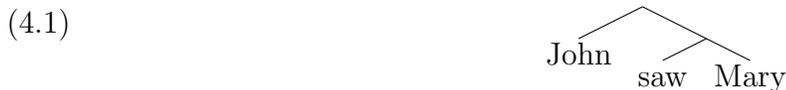
\begin{equation}\label{shzsimple}
 \Tree[ John [ saw Mary ] ] 
\end{equation} 
 where we identify ``saw" as the head, ``Mary" as the complement of the
head, and ``John" as the specifier position. In the full form, a syntactic
object would instead looks like
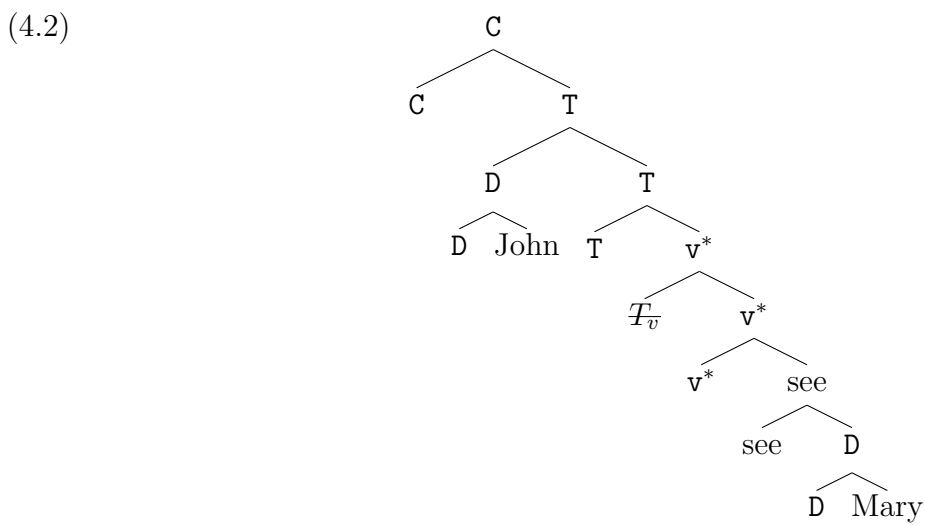
\begin{equation}\label{shznotsimple}
 \Tree[ .$\mathtt{C}$ $\mathtt{C}$ [ .$\mathtt{T}$ [ .$\mathtt{D}$ $\mathtt{D}$ John ] [ .$\mathtt{T}$ $\mathtt{T}$ [ .$\mathtt{v}^*$ \text{\sout{$T_v$}} [ .$\mathtt{v}^*$ $\mathtt{v}^*$ [ .see see [ .$\mathtt{D}$ $\mathtt{D}$ Mary ] ] ] ] ] ] ] 
\end{equation}  
where \sout{$T_v$} is the trace of the accessible term
$$ T_v = \Tree[ .$\mathtt{D}$ $\mathtt{D}$ John ] $$
moved by IM, and $\mathtt{C}$, $\mathtt{T}$, $\mathtt{D}$, $\mathtt{v}^*$ are the heads giving rise to functional projections. Note how
with this formulation the accessible term $T_v$ is moved to the edge of the phase by Internal Merge,
after the functional projections are added on top of the lexical projection. The head structure
is represented here by marking the internal vertices along each path $\gamma_\ell$ by the
head $\ell$. 

The way this syntactic object is depicted here is as obtained in the Merge derivation generated 
using Sandiway Fong's Minimalist Machine\footnote{\url{https://sandiway.arizona.edu/mpp/mm.html}.}, 
which can be used for many more examples.

\subsubsection{local coloring moves} 
Before describing the general form of the coloring that account for
the bare structure of specifier, head, complement, we can see it
applied to this example. In the simple form \eqref{shzsimple} we just have
$$ \Tree [.$\fh$ $\fs$ [.$\fh$ $\fh$ $\fz$ ] ]  $$
where the three colors $\{ \fs, \fh, \fz \}$ mark the specifier $\fs$, head $\fh$,
and complement $\fz$ positions. We can also write it in the form
$$ 
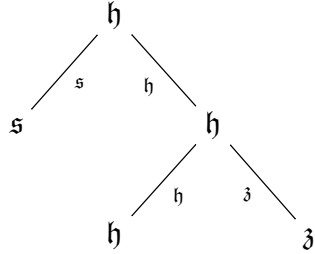
\xymatrix{  &  \fh \ar@{-}[dl]^{\fs} \ar@{-}[dr]_{\fh}  & & \\ 
\fs  & & \fh \ar@{-}[dl]^{\fh} \ar@{-}[dr]_{\fz} & \\ 
& \fh & & \fz 
} $$
In this case the labels at the edges are redundant, as they simply
follow the convention that coloring
a certain vertex is equivalent to coloring the unique edge of the tree
above that vertex, so the vertex and edge colorings are equivalent. 
The internal edge, in this case, is also colored $\fh$, and so is
the root $v_\ell$, as in this simplified form 
they lie on the path $\gamma_\ell$ of the head $\ell=\text{saw}$. 
This example shows a rephrasing in terms of coloring of the form of projections
and phases as described in Sections 1.13 and 1.14 of \cite{MCB}. 

We can also reinterpret this coloring of the simple form \eqref{shzsimple} 
in terms of generators of the form 
\begin{equation}\label{simTgens}
T^\fc_{\fh,\fs} = \Tree[ .$\fc$ {$(\fh, \fs^\uparrow)$} $\fs^\downarrow$ ] \ \ \ \text{ and } \ \ \  T^{\fh}_{\fh,\fz} := \Tree[ .{$(\fh, \fs^\uparrow)$} {$(\fh, \fz^\uparrow,\fs^\uparrow)$} $\fz^\downarrow$ ] 
\end{equation}
as the composition that plugs in the $(\fh,\fs^\uparrow)$-marked output at the root of the second generator
to the $(\fh,\fs^\uparrow)$-marked input at one of the leaves of the first generator. The color $\fc$ at the
root of the first generator can be either another $\fs^\downarrow$ or $\fz^\downarrow$ and it depends on what role
the whole structure obtained in this way will play if incorporated into a larger structure,  
either in a specifier position or as the complement of another head in a higher phase, which contributes 
another $(\fh, \fz^\uparrow, \fs^\uparrow)$. Here one reads the label $\fh$ at a vertex as a coloring
of the edge immediately above it: so these $\fh$-colored edges are exactly those that form the paths
$\gamma_\ell$ of the head function. The label $(\fz^\uparrow,\fs^\uparrow)$ works in the same way as
the labels $\underline{\theta}^\uparrow$ in the case of theta theory in \cite{MarLar}: $(\fh, \fz^\uparrow)$ means
the head is complemented, with a non-empty complement, 
and the further $\fs^\uparrow$ means that it will also carry a specifier. In the generator
$T^{\fh}_{\fh,\fz}$ the head binds with its complement, which receives the label $\fz^\downarrow$ marking it as
the complement position, while the output carries the remaining position that the head will bind to, the
specifier $\fs$. The second generator $T^\fc_{\fh,\fs}$ then receives the $\fs^\uparrow$ through the operad 
composition from the output of the previous one, and
discharges it onto the specifier position that gets marked as $\fs^\downarrow$. The output position then
remains free to assume a new role, either as a specifier or a complement to another head. This vertex is then
the $v_\ell$, the maximal projection of the head $\ell$ carrying the  $(\fh, \fz^\uparrow,\fs^\uparrow)$ and
the endpoint of the path $\gamma_\ell$ which in this case has two edges. 
The generator $T^\fc_{\fh,\fs}$ also accounts for cases where the head $\fh$ has 
an empty complement so only $(\fh, \fs^\uparrow)$ is injected with the specifier position
delivered to the other leaf marked by $\fs^\downarrow$, and an $\fc\in \{ \fz, \fs \}$ as
output to combine with the head of a larger structure. 

\smallskip

This composition of output to input with matching colors is the same type of
colored operad structure used in \cite{MarLar} to model theta role assignments. 

\smallskip

When we consider the second form \eqref{shznotsimple}, however, the structure of the $\gamma_\ell$ paths
is different. Again we want to keep track of the structure of complemented head as we follow the functional projections
and encode the structure into local coloring rules. 

We are going to interpret this case also as a coloring obtained via
a sequence of operad insertions of a certain set of generators.
Again the mathematical setting relevant here is that of colored operads
and bud generating systems already used in \cite{MarLar},
which we will recall more in detail below. 

\smallskip

Consider colored structures of the form
\begin{equation}\label{T1gen}
 T_{\fh^\omega\fz\fs,\fz}^{\fh^\omega\fs} :=  \Tree[ .{$(\fh^\omega,\fs^\uparrow)$} [ {$(\fh^\omega,\fz^\uparrow,\fs^\uparrow)$} $\fz^\downarrow$ ] ] 
\end{equation}
\begin{equation}\label{T2gen} 
T_{\fh^\omega\fz,\fs\fz}^{\fh^\omega\fs} := \Tree[ .{$(\fh^\omega, \fs^\uparrow)$} [ {$(\fh^\omega, \fz^\uparrow)$} {$(\fs^\uparrow, \fz^\downarrow)$} ] ]
\end{equation} 
\begin{equation}\label{T3gen} 
T_{\fh^\omega\fz\fs,\fz}^{\fs\fz} := \Tree[ .{$(\fs^\uparrow, \fz^\downarrow)$} [ {$(\fh^\omega,\fz^\uparrow, \fs^\uparrow)$} $\fz^\downarrow$ ]  ]  
\end{equation}
\begin{equation}\label{T4gen} 
T_{\fh^\omega\fz,\fz}^\fc := \Tree[ .$\fc^\downarrow$  [ {$(\fh^\omega, \fz^\uparrow)$} $\fz^\downarrow$ ] ] \ \ \ \text{ with } \fc\in \{ \fs, \fz \} 
\end{equation}
\begin{equation}\label{T5gen} 
 T_{\fh^\omega\fs,\fs}^\fc :=  \Tree[ .$\fc^\downarrow$ [  {$(\fh^\omega, \fs^\uparrow)$} $\fs^\downarrow$ ] ] 
  \ \ \ \text{ with } \fc\in \{ \fs, \fz \} 
\end{equation} 
where we allow a certain set of colors $\{ \fh^\omega \}_{\omega \in \Omega_\fh}$ 
that distinguishes the lexical heads ($\omega\in \Omega_{\fh,{\rm lex}}$, see Definition~\ref{OmegahDef}) and all 
different possible functional heads $\omega \in \Omega_{\fh,f}=\{ \mathtt{v}^*,\tt{T, C, D}, \ldots \}$.
Since the trees are not planar, the order of the two lower indices does not matter so we equivalently write
$T^{\fc_1}_{\fc_2,\fc_3}=T^{\fc_1}_{\fc_3,\fc_2}$.
In all these two structures, we read the labelled half-edge at the top as
the output and the two half-edges at the bottom as the inputs.

The way to read the labels on one of these generators $T^{\fc_1}_{\fc_2,\fc_3}$ is by interpreting
the label $(\fh^\omega, \underline{\fc}^\uparrow)$ as a head of type $\omega$ that injects colors
$\underline{\fc}$ (either $\fz$ or $\fs$ or a pair $(\fs,\fz)$ meaning that this head will have to bind
with a received of the role of complement $\fz$ or specifier $\fs$: the arrows $\downarrow$, $\uparrow$
distinguish whether a leaf (or a whole structure when it occurs at the root vertex) receives the
corresponding role, marked as $\fc^\downarrow$, or whether a head is assigning it color marking $\fc^\uparrow$. When a
color $\fc^\downarrow$ occurs at an output it means the entire structure below that vertex
receives that role. (The case of theta roles assignments in \cite{MarLar} is
analogous.) 

In \eqref{T1gen}, \eqref{T2gen}, \eqref{T3gen}, \eqref{T4gen}, \eqref{T5gen}, to simplify notation, we only keep track
of the type of head $\fh^\omega$ in the $\fh$ labeling but not explicitly in the associated $\fz$ and $\fs$. To be more
precise, as we will need to in \S \ref{extprojsec} where we 
select an appropriate subset of generators for the Extended Projection Principle (EPP) 
condition, we also write the other labels as $\fz^\uparrow_\omega$,
$\fz^\downarrow_\omega$, $\fs^\uparrow_\omega$, $\fs^\downarrow_\omega$, to keep track of the fact that they are
associated to a $\fh^\omega$-labelled head. Thus, for example, we write, more precisely 
\begin{equation}\label{Tomega1gen}
T_{\fh^\omega\fz\fs,\fz}^{\fs\fz} := \Tree[ .{$(\fs_\omega^\uparrow, \fz_{\omega'}^\downarrow)$} [ {$(\fh^\omega,\fz_\omega^\uparrow, \fs_\omega^\uparrow)$} $\fz_\omega^\downarrow$ ]  ]  
\end{equation}
\begin{equation}\label{Tomega2gen}
 T_{\fh^{\omega'}\fz,\fs\fz}^{\fh^{\omega'}\fs} := \Tree[ .{$(\fh^{\omega'}, \fs_{\omega}^\uparrow)$} [ {$(\fh^{\omega'}, \fz_{\omega'}^\uparrow)$} {$(\fs_\omega^\uparrow, \fz_{\omega'}^\downarrow)$} ] ] 
\end{equation}
Here one interprets a label $\fz_{\omega'}^\downarrow$ at the output as indicating that the whole
structure below that vertex forms the complement of a head of type $\fh^{\omega'}$ in a larger structure,
while a label $\fs_\omega^\uparrow$ in a leaf input indicates that that input can receive an $\fs_\omega$ role
coming from a lower head of type $\fh^\omega$. 

As will be discussed more in detail in \S \ref{bareHSZsec} and \S \ref{bareHSZextsec}, the 
composition rule between these generators corresponds to gluing an output to an input when
they carry matching labels. 

\smallskip

Then we can see that the example \eqref{shznotsimple} can be realized as a chain of
colored operad compositions
$$ 
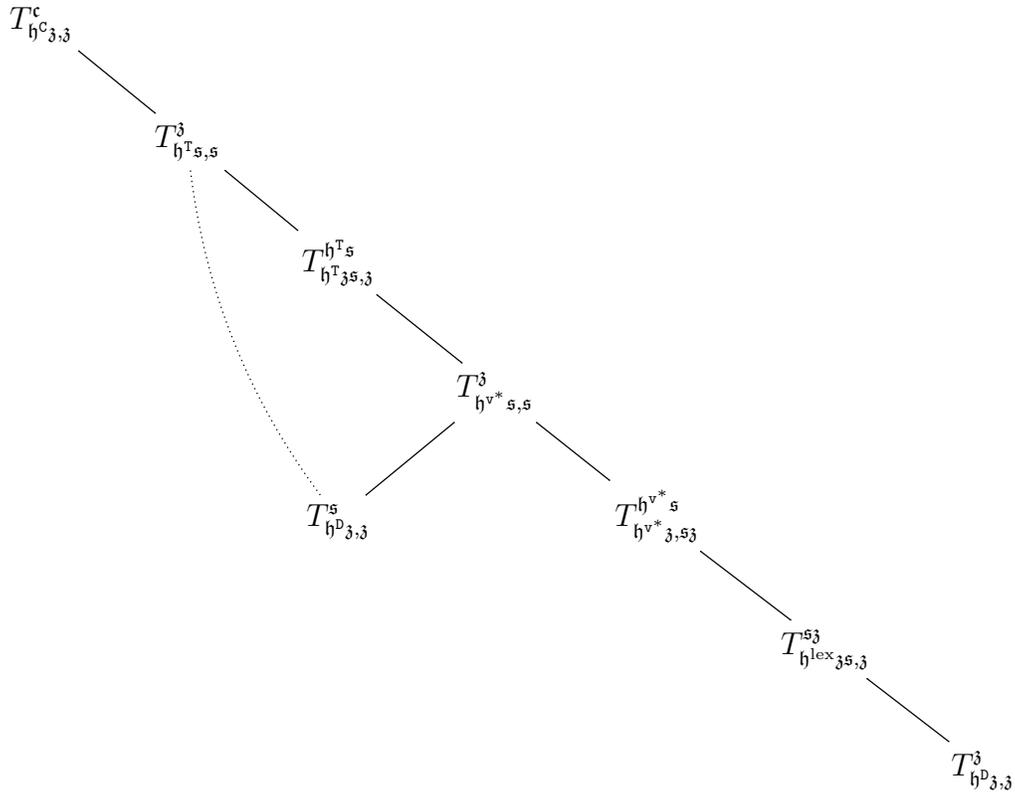
\xymatrix{ 
T^\fc_{\fh^{\mathtt{C}}\fz,\fz} \ar@{-}[dr] & &  &  &  & &  \\
& T^\fz_{\fh^{\mathtt{T}}\fs,\fs} \ar@{-}[dr] \ar@{.}@/_1pc/[dddr]& &  &  &  &   \\
& & T^{\fh^{\mathtt{T}}\fs}_{\fh^{\mathtt{T}}\fz\fs,\fz} \ar@{-}[dr] & & & &   \\
& & & T^\fz_{\fh^{\mathtt{v}^*}\fs,\fs} \ar@{-}[dl] \ar@{-}[dr] & & &   \\
& & T^\fs_{\fh^{\mathtt{D}}\fz,\fz}  & & T^{\fh^{\mathtt{v}^*}\fs}_{\fh^{\mathtt{v}^*}\fz,\fs\fz} \ar@{-}[dr] &  \\
& & & & &   T^{\fs\fz}_{\fh^{\text{lex}}\fz\fs,\fz} \ar@{-}[dr]   &    \\
& & & & & & T^\fz_{\fh^{\mathtt{D}}\fz,\fz}  
} $$

where each line represent an insertion of output of the lower
term to the input of the term above it with matching color 
(colored operad composition). The dotted line represents the movement by Internal Merge
of the term $T^\fs_{\fh^{\mathtt{D}}\fz,\fz}$ from input of $T^\fz_{\fh^{\mathtt{v}^*}\fs,\fs}$ to input of $T^\fz_{\fh^{\mathtt{T}}\fs,\fs}$.
We will describe how this type of movement happens in \S \ref{MergeColSec} and \S \ref{PhaseMoveSec}.

\smallskip

\subsection{Colored operads and bud generating systems}\label{colopSec}

We refer the reader to Section~2.1 of \cite{MarLar} and Section~3.8 of \cite{MCB} for
a quick summary of colored operads. Operads were originally introduced in \cite{May}
and the notion of colored operad is discussed in great detail in \cite{DYau}.
We just recall here that an operad in the category
of sets consists of a collection of sets $\fO=\{ \fO(n) \}_{n\geq 1}$ whose
elements are operations with $n$ inputs and one output. The sets $\fO(n)$ have
composition rules that consist exactly of plugging the output of one operation
into the input of another, 
\begin{equation}\label{insertions}
 \circ_i: \fO(n)\otimes \fO(m) \to \fO(n+m-1). 
\end{equation}
where the index $i$ ranges over the set of $m$ inputs of operations $T\in \fO(m)$
and the result $T\circ_i T'$ of combining the output of $T'$ to the $i$-th input of $T$
has $n+m-1$ remaining free inputs. The associativity of these compositions are expressed
in the form 
$$
 (X\circ_j Y)\circ_i Z = \left\{ \begin{matrix} (X\circ_i Z)\circ_{j+c-1} Y & 1\leq i < j \\
X\circ_j (Y\circ_{i-j+1} Z)& j\leq i < b+j \\
(X\circ_{i-b+1} Z)\circ_j Y & j+b \leq i \leq a+b-1.  \end{matrix} \right. 
$$
The composition operation 
\begin{equation}\label{operadcomp}
 \gamma: \fO(n) \times \fO(k_1)\times \cdots \times \fO(k_n) \to \fO(k_1+\cdots + k_n) 
\end{equation} 
where all the $n$ inputs of $T\in \fO(n)$ are paired to outputs of operations $T_j \in \fO(k_j)$
is obtained as multiple compositions $\circ_i$,
$$
 \gamma(X; Y_1,\ldots, Y_n)=(\cdots (X\circ_n Y_n)\circ_{n-1} Y_{n-1}) \cdots \circ_1 Y_1). 
$$

An algebra $A$ over the operad $\fO$ is a set with the property that the operations of $\fO$
can take inputs in $A$ are return outputs in $A$, namely there is an action
\begin{equation}\label{algoper}
\gamma_A: \fO(n) \times A^n \to A  
\end{equation}
satisfying compatibility with composition \eqref{operadcomp} in the operad, 
\begin{equation}\label{gammaAgamma}
\begin{array}{c}
 \gamma_A(\gamma(T;T_1,\ldots, T_n); a_{1,1}, \ldots, a_{1,k_1}, \ldots, a_{n,1}, \ldots, a_{n,k_n}) = \\
 \gamma_A(T; \gamma_A(T_1;  a_{1,1}, \ldots, a_{1,k_1}),\ldots, \gamma_A(T_n; a_{n,1}, \ldots, a_{n,k_n})) \, . 
 \end{array}
 \end{equation}
 
 A colored operad is a collection of sets 
 $$ \fO=\{ \fO(c,\underline{c})\,|\, \underline{c}=(c_1,\ldots, c_n), \,\, n\geq 1, \,\, c,c_i \in \Omega \} $$
 with $\Omega$ a (usually finite) set of colors. The composition operations are then of the form
 \begin{equation}\label{colopergamma}
\begin{array}{rl}
\gamma: & \fO(c, (c_1,\ldots, c_n)) \times \fO(c_1, (c_{1,1}, \ldots, c_{1,k_1})) \times \cdots \times
\fO(c_n, (c_{n,1}, \ldots, c_{n,k_n}))  \\ & \to \fO(c, (c_{1,1}, \ldots, c_{1,k_1}, \ldots, c_{n,1}, \ldots, c_{n,k_n}))  \end{array}
\end{equation}
subject to the same associativity property, 
 and similarly for the $\circ_i$ compositions. An algebra over a colored operad is a collection of sets $A=\{ A_c \}_{c\in \Omega}$
 with an action
 $$ \gamma_A: \fO(c, (c_1,\ldots,c_n)) \times A_{c_1}\times \cdots \times A_{c_n} \to A_c $$
 satisfying again compatibility with the composition $\gamma$ as in the non-colored case. 
 
 In \cite{Giraudo} colored operads and algebras over colored operads are constructed in terms of
 {\em bud generating systems}. This is the formalism we used in \cite{MarLar} to formalize theta
 theory within the mathematical formulation of Merge and SMT. We recall the setting here briefly.
 
 Given an ordinary operad $\fO$ and a finite set $\Omega$, the bud-operad
$\bB_\Omega(\fO)$ is the colored operad with $\Omega$ as the set of colors
and with all the possible color assigments
\begin{equation}\label{BudO}
\bB_\Omega(\fO)(n) := \Omega \times \fO(n) \times \Omega^n \, .
\end{equation}
The local coloring rules, namely the set of generators of the colored
operad and algebra to be constructed are chosen among the operations in $\bB_\Omega(\fO)(n)$.

A bud generating system $\bB=(\fO,\Omega,\cR, \cI,\cT)$ consists of a (non-colored) operad $\fO$,  a finite set of 
colors $\Omega$, containing a set $\cI$ of initial colors and a set $\cT$ of terminal colors, and a choice of 
a finite subset $\cR  \subset \bB_\Omega(\fO)$ of operations in the bud-operad \eqref{BudO}, which gives
the set of generators (local coloring rules). 

The colored operad $\fO_{\Omega, \bB}$ generated by the bud system $\bB=(\fO,\Omega,\cR, \cI,\cT)$ is 
\begin{equation}\label{opBO}
\fO_{\Omega,\bB}(n):= \{ x=(c,T,\underline{c})\in 
\bB_\Omega(\fO)(n) \,|\, {\bf 1}_c \to_{\bB} x, \, \, c\in \Omega \smallsetminus \cT, \, \, 
\underline{c}\in (\Omega \smallsetminus \cT)^n \} \, ,
\end{equation}
where we write ${\bf 1}_c \to_{\bB} x$ to mean that $x$ is obtained, starting from ${\bf 1}_c$
via a sequence of operad compositions $\circ_{i_k} r_k$ of generators $r_1,\ldots, r_N\in \cR$. 

The language $\cL(\bB)$ generated by the bud system $\bB=(\fO,\Omega,\cR, \cI,\cT)$ is an
algebra over the colored operad $\fO_{\Omega, \bB}$ given by 
\begin{equation}\label{LBlang}
 \cL(\bB)=\{ x=(c,T,\underline{c})\in \bB_\Omega(\fO)\,|\, {\bf 1}_c \to_{\bB} x, \, \, c\in \cI, \, \, \underline{c}\in \cT^n \} \, . 
\end{equation} 

\smallskip
\subsection{The bare colored operad of narrow complemented heads}\label{bareHSZsec}

As a first step, we construct a colored operad that accounts for the
specifier, head, and complement structure in the cases of syntactic objects
represented in the simple form, such as in the example of \eqref{shzsimple}
where all the leaves have lexical items associated to them. 

In this case, we take the set of colors to be of the form
$$ \Omega_b =\Omega_{b,f} \sqcup \Omega_{b, {\rm lex}} $$
$$ \Omega_{b,f} =\{ (\fh,\fz^\uparrow, \fs^\uparrow), (\fh,\fs^\uparrow), \fz^\downarrow, \fs^\downarrow \} $$
$$ \Omega_{b, {\rm lex}}  =  \cS\cO_0 \times \Omega_{b,f} \, . $$
We take $\cT_b=\Omega_{b, {\rm lex}} $ and $\cI_b= \Omega_{b,f}$. 
As the local coloring rules we choose the three elements $T^\fz_{\fh,\fs}$, $T^\fs_{\fh,\fs}$, and $T^\fh_{\fh,\fz}$
of \eqref{simTgens},
\begin{equation}\label{Rb}
\cR_b =\cR_{b,f}\sqcup \cR_{b,\text{lex}} \ \ \ \text{ with } \ \ \ \cR_{b,f}=\{ T^\fz_{\fh,\fs}, T^\fs_{\fh,\fs}, T^\fh_{\fh,\fz} \} \, , 
\end{equation} 
and $\cR_{b,\text{lex}}$ consisting of the same generators, where at least one of the leaves marked 
by elements $(\fh, \fz^\uparrow, \fs^\uparrow)$,
$\fz^\downarrow$, $\fs^\downarrow$ of $\Omega_{b,f}$ is replaced by a leaf with a corresponding marking 
$(\alpha_1, \fh, \fz^\uparrow, \fs^\uparrow)$,
$(\alpha_2, \fz^\downarrow)$, $(\alpha_3, \fs^\downarrow)$ in $\Omega_{b, {\rm lex}}$, where 
the $\alpha_i$ are lexical items. 

\begin{defn}\label{barePhinarrow}
The bare colored operad of narrow complemented heads is the operad $\fO_{\Omega_b, \bB_b}$, obtained
as in \eqref{opBO}, 
generated by the bud generating system $\bB_b=(\cM_h, \Omega_b, \cR_b, \cI_b, \cT_b)$
described above, with the Merge operad $\cM_h$ with head, as described in Section~2.2 of
\cite{MarLar} and in Section~3.8 of \cite{MCB}. The language of bare complemented heads
is the language $\cL(\bB_b)$ obtained as in \eqref{LBlang}.
\end{defn}

\smallskip
\subsection{The bare colored operad of extended complemented heads}\label{bareHSZextsec}

\begin{defn}\label{OmegahDef}
Let $\Omega_\fh$ denote the set of functional categories ({\tt C}, {\tt INFL}, {\tt v}, {\tt v}$^*$, etc.) and lexical categories ({\tt V}, {\tt N}, etc.) of heads, with these two subsets respectively denoted by $\Omega_{\fh,f}$ and $\Omega_{\fh,{\rm lex}}$,
$$ \Omega_\fh = \Omega_{\fh,f} \sqcup \Omega_{\fh,{\rm lex}} \, . $$
\end{defn}

Now we consider the full structure including both the functional and the lexical heads $\fh^\omega$ with $\omega\in \Omega_\fh$ 
and the extended
projection. Consider colors of the form 
$$ \Omega_{b,ext} = \Omega_{b,ext,f} \sqcup \Omega_{b,ext,{\rm lex}} $$
$$ \Omega_{b,ext,f}= \tilde\Omega |_{\omega\neq {\rm lex}} \ \ \ \text{ and } \ \ \ 
 \Omega_{b,ext,{\rm lex}}= \cS\cO_0 \times \tilde\Omega |_{\omega={\rm lex}}  \, , $$
 where
 \begin{equation}\label{Ombextf}
 \tilde \Omega := \{ (\fh^\omega,\fz^\uparrow, \fs^\uparrow), (\fh^\omega,\fz^\uparrow), (\fh^\omega,\fs^\uparrow), \fz^\downarrow, \fs^\downarrow, (\fs^\uparrow,\fz^\downarrow) \, |\, \omega \in \Omega_\fh \} \, ,
\end{equation} 
with $\cT_{b,ext}=\Omega_{b,ext,{\rm lex}}$ and $\cI_{b,ext}
=\Omega_{b,ext}\smallsetminus \cT_{b,ext}=\Omega_{b,ext,f}$. 

As the local coloring rules we take $\cR_{b,ext}=\cR_{b, ext, f} \sqcup \cR_{b, ext, \text{lex}} $ with
\begin{equation}\label{Rbext}
  \cR_{b,ext,f} =\{ 
T_{\fh^\omega\fz_\omega,\fz_\omega}^\fc, T^{\fh^\omega\fs_\omega}_{\fh^\omega\fz_\omega\fs_\omega,\fz_\omega},
T_{\fh^\omega\fz_\omega\fs_\omega,\fz_\omega}^{\fs_\omega\fz_{\omega'}}, T_{\fh^{\omega'}\fz_{\omega'},\fs_\omega\fz_{\omega'}}^{\fh^{\omega'}\fs_\omega}, 
T_{\fh^\omega\fs_\omega,\fs_\omega}^\fc, T_{\fh^{\omega'}\fs_\omega,\fs_\omega}^\fc \} \, , 
\end{equation} 
with the generators
\begin{equation}\label{omegaT1T2gen} 
T_{\fh^\omega\fz_\omega,\fz_\omega}^\fc := 
\Tree[ .$\fc$  [ {$(\fh^\omega, \fz_\omega^\uparrow)$} $\fz_\omega^\downarrow$ ] ]   \ \ \ \ \  \ \ \ \ 
T^{\fh^\omega\fs_\omega}_{\fh^\omega\fz_\omega\fs_\omega,\fz_\omega}= \Tree[ .{$(\fh^\omega, \fs_\omega^\uparrow)$} [ {$(\fh^\omega, \fz_\omega^\uparrow, \fs_\omega^\uparrow)$}  $\fz_\omega^\downarrow$ ] ] 
\end{equation} 
\begin{equation}\label{omegaT3T4gen}
T_{\fh^\omega\fz_\omega\fs_\omega,\fz_\omega}^{\fs_\omega\fz_{\omega'}} := \Tree[ .{$(\fs_\omega^\uparrow, \fz_{\omega'}^\downarrow)$} [ {$(\fh^\omega,\fz_\omega^\uparrow, \fs_\omega^\uparrow)$} $\fz_\omega^\downarrow$ ]  ]     \ \ \ \  \ \ \ \ 
 T_{\fh^{\omega'}\fz_{\omega'},\fs_\omega\fz_{\omega'}}^{\fh^{\omega'}\fs_\omega} := \Tree[ .{$(\fh^{\omega'}, \fs_{\omega}^\uparrow)$} [ {$(\fh^{\omega'}, \fz_{\omega'}^\uparrow)$} {$(\fs_\omega^\uparrow, \fz_{\omega'}^\downarrow)$} ] ]  \ \ \ \text{ with } \omega'\neq \omega
 \end{equation}
\begin{equation}\label{omegaT5T6gen}
 T_{\fh^\omega\fs_\omega,\fs_\omega}^\fc :=  \Tree[ .$\fc$ [  {$(\fh^\omega, \fs_\omega^\uparrow)$} $\fs_\omega^\downarrow$ ] ]  
  \ \ \ \  \ \ \ \ 
T_{\fh^{\omega'}\fs_\omega,\fs_\omega}^\fc := 
 \Tree[ .$\fc$ [  {$(\fh^{\omega'}, \fs_\omega^\uparrow)$} $\fs_\omega^\downarrow$ ] ]  \ \ \ \text{ with } \omega'\neq \omega
\end{equation}
and with $\cR_{b, ext, \text{lex}}$ containing the same generators with $\omega=\text{lex}$ and with the label
of at least one of the two leaves in $\Omega_{b,ext,{\rm lex}}$, namely with one or both leaf-labels 
$(\fh^{\text{lex}}, \fz^\uparrow, \fs^\uparrow)$, $(\fh^{\text{lex}}, \fz^\uparrow)$, $(\fh^{\text{lex}}, \fs^\uparrow)$, $\fs^\downarrow$, $\fz^\downarrow$ replaced by a corresponding
$(\alpha_1, \fh^{\text{lex}}, \fz^\uparrow, \fs^\uparrow)$, $(\alpha_2, \fh^{\text{lex}}, \fz^\uparrow)$, $(\alpha_3, \fh^{\text{lex}}, \fs^\uparrow)$, $(\alpha_4, \fs^\downarrow)$, $(\alpha_5, \fz^\downarrow)$ where the $\alpha_i$ are lexical items.

\begin{defn}\label{barePhi}
The bare colored operad of complemented heads with extended projection is the operad $\fO_{\Omega_{b,ext}, \bB_{b,ext}}$
generated by the bud generating system $\bB_{b,ext}=(\cM_h, \Omega_{b,ext}, \cR_{b,ext}, \cI_{b,ext}, \cT_{b,ext})$
described above. The language of bare complemented heads with extended projection is the language $\cL(\bB_{b,ext})$
obtained as in \eqref{LBlang}.
\end{defn}

\smallskip
\subsection{Cartographic Models}\label{cartoSec}

In the choice of generators of the form \eqref{Tomega1gen} and \eqref{Tomega2gen},
$$ T_{\fh^\omega\fz\fs,\fz}^{\fs\fz} := \Tree[ .{$(\fs_\omega^\uparrow, \fz_{\omega'}^\downarrow)$} [ {$(\fh^\omega,\fz_\omega^\uparrow, \fs_\omega^\uparrow)$} $\fz_\omega^\downarrow$ ]  ]   \ \ \ \text{ and } \ \ \ 
 T_{\fh^{\omega'}\fz,\fs\fz}^{\fh^{\omega'}\fs} := \Tree[ .{$(\fh^{\omega'}, \fs_{\omega}^\uparrow)$} [ {$(\fh^{\omega'}, \fz_{\omega'}^\uparrow)$} {$(\fs_\omega^\uparrow, \fz_{\omega'}^\downarrow)$} ] ]  \, , $$
the pairs $(\omega,\omega')$ occurring in the color label $(\fs_\omega^\uparrow, \fz_{\omega'}^\downarrow)$
specify the possible hierarchies of heads in well formed syntactic objects and in particular the hierarchy
of functional heads in the Extended Projection. 

So far we have left the set of pairs $(\omega,\omega')$ unconstrained. We will discuss in the following
sections empirical constraints that one can impose, that restrict the set of generators \eqref{Tomega1gen} 
and \eqref{Tomega2gen}, that limit these possibilities according to the usual structure of the Extended 
Projection in verbal phrases. 

One can, more generally, allow for possible models of the constraints on the pairs $(\omega,\omega')$
that can occur in the generators \eqref{Tomega1gen}  and \eqref{Tomega2gen}. Linguistically these
are the cartographic models, in the sense of \cite{CinRi}.

\begin{defn}\label{cartodef}
Let $\Omega_\fh$ denote the set of head categories as in Definition~\ref{OmegahDef}, with the
subset $\Omega_{\fh,f}$ of functional categories ({\tt C}, {\tt INFL}, {\tt v}, {\tt v}$^*$, etc.) and
the subset $\Omega_{\fh,{\rm lex}}$ of lexical categories ({\tt V}, {\tt N}, etc.) of heads. 
A cartographic model is a choice of a subset $\cK \subset \Omega_\fh \times \Omega_\fh$ with
the properties that the relation $\omega' \succeq_\cK \omega$ iff $(\omega,\omega')\in \cK$ is a partial order.
Generators of the form \eqref{Tomega1gen} and \eqref{Tomega2gen} where the coloring 
$(\fs_\omega^\uparrow, \fz_{\omega'}^\downarrow)$ satisfy $\omega' \succeq_\cK \omega$ are
a $\cK$-cartographic coloring. 
\end{defn}

\begin{defn}\label{RKcarto}
Given a cartographic model $\cK$ as above, we can restrict the set of generators $\cR_{b,ext}$ in
\eqref{Rbext} to a subset $\cR_{\cK,b,ext}\subset \cR_{b,ext}$
with the constraint that the generators of the form \eqref{Tomega1gen} and \eqref{Tomega2gen}
belong to $\cR_{\cK,b,ext}$ iff $\omega' \succeq_\cK \omega$, namely they give a $\cK$-cartographic coloring.
\end{defn}

\begin{defn}\label{KbarePhi}
Given a cartographic model $\cK$, in the sense of Definition~\ref{cartodef}, the 
$\cK$-cartographic bare colored operad of complemented heads with extended 
projection is the operad $\fO_{\Omega_{b,ext}, \bB_{\cK,b,ext}}$
generated by the bud generating system $\bB_{\cK,b,ext}=(\cM_h, \Omega_{b,ext}, \cR_{\cK,b,ext}, \cI_{b,ext}, \cT_{b,ext})$.
\end{defn}

In the following we will make some specific assumptions about the 
cartographic model $\cK$ that reflect the structure of the Extended Projection of {\tt VP}s. 
In general, one can consider the operad $\fO_{\Omega_{b,ext}, \bB_{b,ext}}$ as
in Definition~\ref{barePhi}, without any choice of a cartographic model, and
implement specific cartographic models in a successive filter at the interfaces (syntax-semantics
and Externalization), or incorporate some cartographic assumptions directly in the
construction of the colored operad of phases.

\smallskip
\subsection{Extended Projection Principle} \label{extprojsec}

The Extended Projection Principle (see \cite{Chomsky82}, \cite{ChomskyPP}, \cite{ChomskyPPext})
states as a linguistic hypothesis that clauses must necessarily contain a noun phrase or determiner 
phrase in the ``subject position",
namely in the specifier position of an inflectional phrase {\tt INFL}, or of a verb phrase, 
when subject does not raise to $\mathtt{TP}$/$\mathtt{IP}$, as in NullSubject languages, where one can have 
$\{ {\tt INFL}, \{ {\tt EA}, {\tt v}^*{\tt P} \}\}$. 
Note that, in the setting of \cite{ChomskyGK}, $\mathtt{T}$ is not in {\tt INFL} 
but is part of light verb $v$ head: this accounts for cases
like {\em ``John arrives every day at noon and met Bill yesterday"}, where each conjunct has its own, different, tense, 
but there is just a single {\tt INFL}. 

\smallskip

More precisely, one can split this into two properties:
\begin{enumerate}
\item the functional heads occur as extended projection of the lexical heads (above the part of the structure involving
the lexical head and its maximal projection, as in the $\mathtt{CP}$-{\tt INFLP}-$\mathtt{v}^*${\tt P}/{\tt vP} 
extended projection of {\tt VP});
\item the specifier position of the {\tt INFL} head (or possibly $\mathtt{v}^*$ head, as in NullSubject languages) is occupied 
by a noun phrase or determiner phrase.
\end{enumerate}
For a discussion of movement by Internal Merge and the structure of the extended projection, see \S \ref{PhaseMoveSec}
and \S \ref{ThetaPhaseMoveSec} below.

\begin{lem}\label{bareEPP}
The language $\cL(\bB_b)$ of narrow bare complemented heads satisfies a weaker form of Extended Projection Principle (EPP).
Namely every phrase (meant as the maximal projection of a head) must contain either a lexical item or a phrase in the
specifier position. 
\end{lem}

\proof  Given the form \eqref{simTgens} of the generators $T^\fz_{\fh,\fs}$, $T^\fs_{\fh,\fs}$, and $T^\fh_{\fh,\fz}$ in $\cR_b$,
every head $\ell$ is marked by a label $(\alpha, \fh, \fz^\uparrow, \fs^\uparrow)$ and has to occur in combination with a
complement, in a generator of type $T^\fh_{\fh,\fz}$. The output of this generator is labelled $(\fh, \fs^\uparrow)$ and
must therefore combine with an input with the same marking, hence with the $(\fh, \fs^\uparrow)$-marked leaf of the 
generator of the form $T^\fc_{\fh,\fs}$, where $\fc$ is either $\fz^\downarrow$ or $\fs^\downarrow$. This generator 
has the other leaf marked $\fs^\downarrow$, which means exactly that the phrase (terminating at the $v_\ell$ vertex)
must contain either a lexical item or a phrase at the determiner position.
\endproof

In the bare colored operad of complemented heads with extended projection and the associated language
$\cL(\bB_{b,ext})$ it is possible to make the Extended Projection Principle more precise. We write here the
generators in the form as in \eqref{Tomega1gen} and \eqref{Tomega2gen}, keeping tracks of the different
functional heads. 

\smallskip

\begin{defn}\label{EPPgens}
Consider the subset $\cR_{b,ext, {\rm EPP}}\subset \cR_{b,ext}$ consisting of the same generators in  as in \eqref{T1gen}, \eqref{T2gen}, \eqref{T3gen}, \eqref{T4gen}, \eqref{T5gen} in $\cR_{b, ext, f}$ and corresponding generators in $\cR_{b, ext, \text{lex}}$, but with the following restrictions on the generators:
\begin{enumerate}
\item the generator $T_{\fh^{\omega'}\fs_\omega,\fs_\omega}^\fc$ 
only occurs (in both $\cR_{b, ext, f}$ and $\cR_{b, ext, \text{lex}}$)
when $\omega'\in \{ \mathtt{v}^*, \mathtt{INFL}, \mathtt{C} \}$ (verb phrase, inflectional phrase, complementizer phrase, respectively), for either $\omega=\omega'$ or $\omega\neq \omega'$
\item the generators in both sets 
$\cR_{b, ext, f}$ and $\cR_{b, ext, \text{lex}}$ with output marked by $\fs_{\omega'}^\downarrow$ carry a head label 
$\fh^\omega$ with $\omega$ marking a determiner or noun phrase; 
\item Let $\Omega_\fh$ be the set of heads. 
The generators $T_{\fh^\omega\fz_\omega\fs_\omega,\fz_\omega}^{\fs_\omega\fz_{\omega'}}, T_{\fh^{\omega'}\fz_{\omega'},\fs_\omega\fz_{\omega'}}^{\fh^{\omega'}\fs_\omega}$
belong to $\cR_{b,ext, {\rm EPP}}$ iff the pair $(\omega,\omega')$ in the label $(\fs_\omega^\uparrow, \fz_{\omega'}^\downarrow)$
has $\omega' \neq {\rm lex} \in \Omega_\fh$ (functional heads occur above lexical head in the extended projection).   When
both $\omega,\omega'\neq {\rm lex}$, the pair $(\omega,\omega')$ follows a partial order structure among the
functional head that determines the order in which they occur in the extended projection (e.g., $\mathtt{C}> \mathtt{INFL}> \mathtt{v}^*$).
\end{enumerate}
\end{defn}

The partial ordering in the Extended Projection that constraints the choice of pairs $(\omega,\omega')$ 
can be seen in terms of choices of cartographic models, as in Definition~\ref{cartodef}
and Definition~\ref{KbarePhi}.

\begin{prop}\label{bareEPPext}
Let $\bB_{b,ext, {\rm EPP}}$ denote the bud generating
system with the same colors as $\bB_{b,ext}$ and the restricted set of generators $\cR_{b,ext, {\rm EPP}}$, as in
Definition~\ref{EPPgens}. 
The resulting colored operad $\fO_{\Omega_{b,ext}, \bB_{b,ext, {\rm EPP}}}$ and the associated language $\cL(\bB_{b,ext, {\rm EPP}})$ satisfy the Extended Projection Principle.
\end{prop}

\proof The third condition constrains the composition rules in the operad $\fO_{\Omega_{b,ext}, \bB_{b,ext}}$ 
(namely, the composition rules between generators in $\cR_{b,ext}$) implying that the functional heads must
occur above a lexical head, a necessary part of the EPP principle. The first and second constraints
imply that the discharging of a $\fs^\downarrow_\omega$ can only occur in the Spec position
of a head of type $\omega' \in \{ \mathtt{v}^*, \mathtt{INFL}, \mathtt{C} \}$. These happen at the top of the extended projection, 
since the generators involving a leaf labelled $\fs^\downarrow_\omega$ have a single $\fc^\downarrow$
at the output, which means that there is no further upward pointing $\fs^\uparrow$ from the head that requires
the phase to be completed. Thus, the EPP principle is satisfied in $\cL(\bB_{b,ext, {\rm EPP}})$ and
$\fO_{\Omega_{b,ext}, \bB_{b,ext, {\rm EPP}}}$.
\endproof

\begin{rem}\label{fheadcompl}{\rm
The restrictions of Definition~\ref{EPPgens} on the generators of $\cR_{b,ext}$ give the minimal restriction
(largest subset of $\cR_{b,ext}$) that guarantees that EPP is satisfied. On top of this condition, one can further
restrict the set of generators of $\cR_{b,ext}$ by including other constraints, for example on the type of complement
that a certain head can take. Namely given $\fh^\omega$ with either $\omega={\rm lex}$ or with $\omega$ a 
functional category, we can further distinguish the associated $\fz_\omega$ and $\fs_\omega$ into types 
depending on $\omega$, For example, for lexical heads $\mathtt{A,N,P,V}$ the complements would be $\fz_A, \fz_N \in \{ \mathtt{CP}, \mathtt{PP} \}$,
$\fz_P\in \{ \mathtt{IP, DP, PP} \}$, $\fz_V\in \{ \mathtt{CP, DP, PP} \}$. In the case of $\omega \in \Omega_\fh$ a functional category,
if $\omega=\mathtt{C}$ then $\fz_C=\mathtt{IP}$, if $\omega=\mathtt{I}$ then $\fz_I=\mathtt{VP}$, if $\omega=\mathtt{D}$ then $\fz_D=\mathtt{NP}$, etc. 
These rules can be implemented as further restrictions on the choice of the generators that form $\cR_{b,ext, {\rm EPP}}$.
We can consider these restrictions on the generators as being part of the choice of a cartographic model in the
sense of Definition~\ref{cartodef} and Definition~\ref{KbarePhi}. 
}\end{rem}

Note further that some of the terminology for functional categories of heads is somewhat different
in the recent literature on Minimalism, but the same selection of generators can be performed with
the updated notation.

\smallskip
\subsection{The colored operad of complemented heads with modifiers} \label{fullHSZsec}

In the colored operads constructed in \S \ref{bareHSZsec} and \S \ref{bareHSZextsec}, 
we only consider positions that are heads, complements, specifiers. This approach is similar to the
construction done in \cite{MarLar} of the colored operad of bare theta structures, where only 
theta positions are included. One can then refine this construction by including positions that
do not participate in a mandatory set of relations: in the case of head and phases, these non-mandatory
relations are the {\em modifiers}, while in the case of theta theory they correspond to non-theta positions,
as in \cite{MarLar}.

\smallskip

In the setting considered here, one can include modifier positions, which modify the complement, or
the head, or the specifier. We introduce an additional marking $\fm$ for these modifiers. Again we
can consider a simplified form of the syntactic objects, as in \S \ref{bareHSZsec} or the more complete
form as in \S \ref{bareHSZextsec} with the Extended Projection, and introduce possible additional modifier positions. 

\begin{defn}\label{modEPPlangN}
Consider the set of colors $\Omega=\Omega_f\sqcup \Omega_{\rm lex}$ with
$\Omega_f=\Omega_{b,f} \sqcup \{ \fm \}$ and $\Omega_{\rm lex}=\Omega_{b,{\rm lex}}\sqcup  \cS\cO_0\times \{ \fm \}$, 
and the set of generators 
$$ \cR =  \cR_f \sqcup \cR_{\rm lex} \ \  \ \text{ with } \ \    \cR_f=\cR_{b,f} \sqcup \{ T^\fc_{\fc,\fm}, \, \, \fc\in \Omega \}  $$
with
\begin{equation}\label{mTgens}
T^\fc_{\fc,\fm} = \Tree[ .$\fc$ [ $\fc$ $\fm$ ] ] 
\end{equation}
and with $\cR_{\rm lex}$ with the same generators as $\cR_f$ but where at least one of the leaves
occurs together with a lexical item, $(\alpha,\fc)$ or $(\alpha,\fm)$. 
The colored operad $\fO_{\Omega, \bB}$ of complemented heads with modifiers is generated
by the bud system $\bB=(\cM_h,\Omega, \cR, \cI=\Omega_f, \cT=\Omega_{\rm lex})$, with 
associated language $\cL(\bB)$.
\end{defn}

\smallskip
\subsection{The colored operad of extended complemented heads with modifiers} \label{fullHSZextsec}

As in \S \ref{bareHSZextsec} we passed from the
bare colored operad of narrow complemented heads of \S \ref{bareHSZsec}
to the bare colored operad of extended complemented heads, which includes the structure
of the Extended Projection, here we pass from the 
colored operad of complemented heads with modifiers of \S \ref{fullHSZsec},
to a version that accounts for the Extended Projection and for the
effect of movement.

\begin{defn}\label{modEPPlang}
Consider the set of colors $$ \Omega_{ext}=\Omega_{b,ext} \sqcup \{ \fm \} \sqcup \cS\cO_0\times \{ \fm \} \,\,\, \sqcup \{ (1, \fm) \}\, , $$ with $\Omega_{ext,f}=\Omega_{b,ext,f}\sqcup\{ \fm \}$ and 
$$ \Omega_{ext,{\rm lex}}=\Omega_{b,ext,{\rm lex}}\,\, \sqcup \,\, \cS\cO_0\times \{ \fm \}\,\,\, \sqcup \{ (1, \fm) \} $$ and the 
subset $\cR_{b,ext, {\rm EPP}}\subset \cR_{b,ext}$ as in Definition~\ref{EPPgens}.
Take
$$ \cR_{ext}= \cR_{ext,f} \sqcup \cR_{ext,\text{lex}} \ \ \  \text{ with } \cR_{ext,f}=\cR_{b,ext,{\rm EPP},f} \sqcup
\{ T^\fc_{\fc,\fm}, \, \fc\in \Omega \} $$
with $T^\fc_{\fc,\fm}$  as in \eqref{mTgens} with $\fh=\fh^\omega$ for some
$\omega\in \Omega_\fh$, 
\begin{equation}\label{extmTgens}
T^\fc_{\fc,\fm} = \Tree[ .$\fc$ [ $\fc$ $\fm$ ] ] 
\end{equation}
representing modifiers (including modifiers of complement, specifier, and head). We also include in the
generators $T_{\fh^\omega\fz,\fz}^\fc$ and $T_{\fh^\omega\fs,\fs}^\fc$ in $\cR_{b,ext,{\rm EPP},f}$
the possibility that $\fc=\fm$ at the root. The set $\cR_{ext,\text{lex}}$ contains generators of the same form
as those in $\cR_{ext,f}$,  but where at least one of the leaves labels is accompanied by a lexical item 
$(\alpha,\fc)$, for $\fc$ occurring at the leaves of the generators of $\cR_{ext,f}$. 
Additionally, in $\cR_{ext, \text{lex}}$ also contains generators that involve the color $(1,\fm)$, and are of the form
\begin{equation}\label{gens1m}
T^\fs_{\fc,(1,\fm)}:= \Tree[ .$\fs^\downarrow_{\omega'}$ [ $\fc^\downarrow_\omega$ $(1,\fm)$ ] ] \ \ \ \text{ with }  \fc\in \{ \fs, \fz \} \, .  
\end{equation}
The colored operad $\fO_{\Omega_{ext}, \bB_{ext, {\rm EPP}}}$ of extended complemented heads with modifiers is generated by $$\bB_{ext, {\rm EPP}}=(\cM_h,\Omega_{ext}, \cR_{ext}, \cI_{ext}=\Omega_{ext,f}, \cT_{ext}=\Omega_{ext,\text{lex}}), $$
with the associated language $\cL(\bB_{ext, {\rm EPP}})$. 
\end{defn}

Note that we are also formally including here a new color $(1,\fm)$ where $1$ is the
unit of the magma of syntactic objects, namely the formal empty tree. The reason for
including this additional color is analogous to the case of $(1,\theta_0)$ in \cite{MarLar},
and provides the rules for IM movement. In particular, in \eqref{gens1m} 
the color $\fc^\downarrow_\omega$ will determine the position that IM moves from
and $\fs^\downarrow_{\omega'}$ the position it moves to. 

\begin{rem}\label{nomovemh}{\rm
Here taking $\fc\in \{ \fs, \fz \}$ in the generator \eqref{gens1m}
implies not moving from a modifier $\fm$ position or from a head $\fh^\omega$ position.
Both of these restrictions could be lifted, by including additional generators of the form
\begin{equation}\label{genT1mm}
T^\fs_{\fm,(1,\fm)}:= \Tree[ .$\fs^\downarrow_{\omega'}$ [ $\fm$ $(1,\fm)$ ] ] 
\end{equation}
or of the form
\begin{equation}\label{genT1mh}
T^\fs_{\fc,(1,\fm)}:= \Tree[ .$\fs^\downarrow_{\omega'}$ [ $\fc$ $(1,\fm)$ ] ]  \ \ \ \text{ with } \ \ 
\fc\in \{   (\fh^\omega,\fz_\omega^\uparrow, \fs_\omega^\uparrow), (\fh^\omega,\fz_\omega^\uparrow), (\fh^{\omega'},\fs_\omega^\uparrow)   \} \, .
\end{equation}
Introducing the possibilities \eqref{genT1mh} creates a possible problem with the labeling algorithm, as we will
discuss more in detail in \S \ref{2ObjMoveSec}.
}\end{rem}

The formulation in terms of movement is discussed in \S \ref{PhaseMoveSec} below
and following sections.

The resulting colored operad $\fO_{\Omega_{ext}, \bB_{ext, {\rm EPP}}}$  and language $\cL(\bB_{ext, {\rm EPP}})$
still satisfy the EPP, since the part handling the specifier-head-complement structure comes from  
the generators in $\cR_{ext,f}=\cR_{b,ext,{\rm EPP}}$ that build structures satisfying EPP. The additional presence of
the $\fm$ modifier positions do not alter this fact, as they do not alter the specifier-head-complement structure. 

As in the case of the bare colored operad of extended complemented heads, we can restrict the
generators according to a cartographic model, as in Definition~\ref{cartodef} and Definition~\ref{KbarePhi}.

\smallskip
\subsection{Phases} \label{PhasesSec}

The formulation of the specifier-head-complement structure (including modifiers) in terms of
a coloring algorithm and a colored operad bud generating system then provides a way of
describing phases in terms of this same colored operad and its associated language.

In the simplified form of the trees, which corresponds to the colored operad of narrow complemented head,
the phases are simply determined by the paths $\gamma_\ell$ and the maximal projection vertices
$v_\ell$ of $\gamma_\ell$ (when the path is nontrivial). We refer to this structure as {\em narrow phases}.

On the other hand, in the full form that determines the colored operad of extended complemented
heads, where the functional heads are accounted for, and the extended projection, the structure of
phases is more subtle, because not all the heads that project determine phases, and this
complicates the relation between the paths $\gamma_\ell$ of the head function
and the structure of phases, see \cite{ChomskyPP}, \cite{ChomskyPPext}, \cite{Citko}.  For example, a
lexical V head $\ell$ now does not complete its phase at the $v_\ell$ vertex, but at the $v_{\ell'}$
of the functional head $\ell'=\mathtt{v}^*$, and while $\mathtt{C}$ and $\mathtt{v}^*$ 
heads determine phases, {\tt INFL}, $\mathtt{v}$ heads
are not phase heads. A result of the fact that not all functional heads are phase heads is that a
Spec-of-{\tt INFL} or Spec-of-Root position, for example, is in the interior rather than at the edge of 
the phase. This has consequences on movement via Internal Merge. 

\smallskip

\begin{defn}\label{OmegahphiDef}
 Let $\Omega_\fh$ denote, as in Definition~\ref{OmegahDef}, the set of possible heads categories, with $\Omega_{\fh,f} \subset \Omega_\fh$
 the set of functional heads $\omega\notin \Omega_{\fh, {\rm lex}}$. One can further identify a subset $\Omega_{\fh,\phi} \subset \Omega_{\fh,f}$
 of functional heads that are phase heads. 
 \end{defn}

\begin{defn}\label{PhaseOp}
Given a syntactic object $T\in \fT_{\cS\cO_0}$ with a complemented head function $h_T$, the
phases of $T$ can be described in the following way. 

\begin{itemize}
\item In the simple representation $T\in \cL(\bB)$ 
as in Definition~\ref{modEPPlangN}, a {\em narrow phase} of $T$ is an accessible term $T_v\subset T$ with the properties:
\begin{enumerate}
\item the leaf $\ell=h_T(v)$ is labelled by $(\alpha, \fh, \fz^\uparrow, \fs^\uparrow)$ or $(\alpha, \fh,  
\fs^\uparrow)$, for some lexical $\alpha\in \cS\cO_0$;
\item the root $v=v_\ell$ is the maximal projection of this head and is a bud generator of the form
$T^\fc_{\fh,\fs}$ as in \eqref{simTgens}.
\end{enumerate}
\item In the full form with the Extended Projection, $T \in \cL(\bB_{ext,{\rm EPP}})$ as in Definition~\ref{modEPPlang},
a {\em phase} of $T$ is an accessible term $T_v\subset T$ with the properties:
\begin{enumerate}
\item $T_v$ contains an accessible term $T_w$ with $w=v_\ell$ for $\ell=h_T(w)$ a leaf labelled by
a lexical head $(\alpha, \fh^\omega, \fz_\omega^\uparrow, \fs_\omega^\uparrow)$ or $(\alpha, \fh^\omega,
\fs_\omega^\uparrow)$, for lexical $\alpha\in \cS\cO_0$ and $\omega\in \Omega_{\fh,{\rm lex}}$;
\item all heads $\ell'$ in $T_v/^d T_w$ with $v_{\ell'}$ on the path from $v$ to $w$ are in $\Omega_{\fh,f}$;
\item the head $\ell' =h_T(v)$ is in $\Omega_{\fh,\phi}$, as in Definition~\ref{OmegahphiDef};
\item the root $v$ of $T_v$ is a bud generator of the form $T^\fc_{\fh\fs,\fs}$ or $T_{\fh^\omega\fz,\fz}^\fc$ as in \eqref{T4gen}
and \eqref{T5gen}.
\end{enumerate}
\item in both cases the {\em edge of the phase} (Spec-of-Phase) is the accessible term 
$T_{v_1}$ with $v_1$ the vertex immediately below $v_\ell$
and marked by $\fs$. The {\em interior of the phase} is the remaining $T_{v_2}=T/^d T_{v_1}$, with $v_2$ the sister
vertex of $v_1$ below $v_\ell$.
\end{itemize}
\end{defn}

There is a subtlety that needs to be taken into account in the assignment of the set $\Omega_{\fh,\phi} \subset \Omega_{\fh,f}$
 of functional heads that are phase heads. The case of $\omega = \mathtt{INFL}$ is particular, in the sense that, depending on its
 interaction with a head $\omega' \succ \omega$, $\omega'=C$, it may inherit a phase-head role or not.
 
 This can be seen, for instance, by comparing the two sentences:
 \begin{equation}\label{exINFL1}
  \text{What did you say } [ C [_1  \text{\sout{what}} [ {\tt INFL} [ \text{\sout{what}} [ {\tt v}^* [ \text{ caused it } ] ] ] ] ] ]  
 \end{equation}
 \begin{equation}\label{exINFL2}
 \text{What did you say } [_1 \text{\sout{what}} [ \text{ that } [_2 \text{\sout{what}} [ {\tt INFL}  [ \text{\sout{what}} [ {\tt v}^* [ \text{ caused it } ] ] ] ] ] ]  ]  \, .
 \end{equation}  
In the first case, {\tt INFL} inherits phase features from C, which labels the position at $[_1$, and ``what" 
can then move from the edge-of-the-phase position,
while in the second case it is ``that" which has to be phasal, while {\tt INFL} is not: the node at $[_2$ 
in the Spec-of-{\tt INFL} position remains just a trace at the point $[_1$ where the phase is completed
and so it is invisible to the labeling algorithm. A consequence is ECP (empty category principle) violation:
labeling operating at the phase level (here $\mathtt{CP}$ level) fails to label $[_2$ by criterial agreement 
$\langle \varphi, \varphi\rangle$ since its Spec is invisible. Labeling requirement thus unify EPP and 
ECP (that-trace filter) for non-null subject languages. 
This behavior of {\tt INFL} with respect to phases is discussed in \cite{ChomskyPPext}.

To account for these two different roles of the {\tt INFL} functional head, we can assume that the set $\Omega_\fh$
contains both an element {\tt INFL} and an element {\tt INFL}$_\varphi$, with the second (which pertains to the cases where
{\tt INFL} inherits the role of phase-head) in  $\Omega_{\fh,\phi}$ and the first ({\tt INFL} that is not a phase-head) in
$\Omega_{\fh,f}\smallsetminus \Omega_{\fh,\phi}$.

\section{Colorings, Hypermagmas, and Colored Merge} \label{HypColorMergeSec} 

As observed also in the case of the colored operad of theta role
assignments, the language $\cL(\bB)$ generated by the bud generating system
is not a submagma of the magma $(\cS\cO,\fM)$ of syntactic objects. In other
words the subset $\cL(\bB)\subset \cS\cO$ of those syntactic objects that are
well colored according to the local coloring rules $\cR$ of $\bB$ is not stable
under the magma operation $\fM$.

This can be seen easily in the cases of both $\cL(\bB)$ of Definition~\ref{modEPPlangN}
and $\cL(\bB_{ext,{\rm EPP}})$ of Definition~\ref{modEPPlang}.

\begin{lem}\label{nomagL}
The languages $\cL(\bB)$ of Definition~\ref{modEPPlangN}
and $\cL(\bB_{ext,{\rm EPP}})$ of Definition~\ref{modEPPlang}
are not stable under the magma operation $\fM$.
\end{lem}

\proof It suffices to produce examples of pairs of elements $T_1,T_2\in \cL(\bB)$
and $T'_1, T'_2\in \cL(\bB_{ext,{\rm EPP}})$ such that $\fM(T_1,T_2)\notin \cL(\bB)$
and $\fM(T'_1, T'_2)\notin \cL(\bB_{ext,{\rm EPP}})$. This can be seen already at the
level of generators. For example, take two generators of the form 
$T_1=T^\fz_{\fh,\fs}$ and $T_2=T^\fs_{\fh,\fs}$ as in \eqref{simTgens}. Then
$\fM(T^\fz_{\fh,\fs},T^\fs_{\fh,\fs})$ cannot be in $\cL(\bB)$ because the
root vertex has the two inputs colored by $\fz^\downarrow$ and $\fs^\downarrow$
but there is no generator in the bud system $\bB$ that has the two inputs labelled 
in this way.  The case of $\cL(\bB_{ext,{\rm EPP}})$ is similar: it suffices to take, for
example, $T'_1=T^\fs_{\fh^\omega\fz,\fz}$ and
$T'_2=T^\fz_{\fh^\omega\fs,\fs}$.
Then by the same argument $\fM(T^\fs_{\fh^\omega\fz,\fz}, T^\fz_{\fh^\omega\fs,\fs})$
cannot be in $\cL(\bB_{ext,{\rm EPP}})$ because no generator of the bud system $\bB_{ext,{\rm EPP}}$
has inputs labelled by $\fs^\downarrow$ and $\fz^\downarrow$.
\endproof

This is a typical situation that we have already encountered for the case of head functions,
where ${\rm Dom}(h)$ is compatible with operad compositions (giving rise to the underlying
operad $\cM_h$ in all the bud generating systems considered here and in \cite{MarLar}),
but not with the magma structure. We also encountered the same behavior with respect to
theta role assignments, and here for the structure of phases. This means that, while the
magma structure is the crucial mechanism of structure formation (alongside movement
that requires the additional coproduct structure of the Hopf algebra of workspaces
as discussed in Chapter~1 of \cite{MCB}), the {\em filtering} that takes care of eliminating
ill formed structures produced by the free symmetric Merge follows a different type
of algebraic structure, described by operads and coloring, which does not preserve the
magma operation.

\subsection{Hypermagmas and coloring algorithms}\label{HypColorSec}

As we discussed in \S \ref{HyperHSec} above, this lack of compatibility with the
magma operation can be expressed in terms of a hypermagma structure, as we did for
the head function in \S \ref{HyperHSec}. 

It is in fact a natural idea that compositional coloring problems (meaning
coloring problems where the local coloring rules can be assembled into
large structures via certain specific composition operations obeying a
specified algebraic structure) may be formulated in terms of hyperstructures,
where the multivalued result of the composition operation keeps track of
possible choices of how to combine the local coloring moves forming
the elements being composed. Well-colorings would then be described as 
coshort hypermagma morphisms that are sections of the forgetful projection
that forgets the coloring (as ${\rm Dom}(h)$ in the case of the head function in 
Proposition~\ref{maghypmag}). Surprisingly, it seems that formulations
of coloring algorithms in terms of hyperstructures is rare (but see for instance \cite{Golmo}). 

For simplicity of notation, we analyze explicitly the case of the languages $\cL(\bB)$ of Definition~\ref{modEPPlangN}.
The case of  $\cL(\bB_{ext,{\rm EPP}})$ of Definition~\ref{modEPPlang} can be treated in exactly the same way.
Let $\bB_\Omega$ be the bud-operad with arbitrary color assignments in the color set $\Omega$ at the
leaves and the root of trees $T\in \cM_h$, as in \eqref{BudO}. We write here $(T,c_T)$ for an element in 
$\bB_\Omega$, with $c_T$ denoting an arbitrary assignment
$c_T: V(T) \to \Omega$ of colors, with $V(T)$ the set of vertices. 

\begin{defn}\label{hypBOmega}
The hypermagma $\fH_{\bB_\Omega}$ is obtained by setting
$$ \fM_{\bB_\Omega}((T,c_T),(T', c'_{T'}))=(\fM(T,T'), \{ c_{\fM(T,T')} \} ) $$
where the set of all $c_{\fM(T,T')}$ consists of all possible functions $c_{\fM(T,T')}: V(\fM(T,T'))\to \Omega$
that agree with $c_T$ and $c'_{T'}$ on the respective subsets $V(T)$ and $V(T')$ of vertices of $V(\fM(T,T'))$,
with an additional arbitrary assignment $c(v_{\fM(T,T')})\in \Omega$ at the root vertex of $\fM(T,T')$.
\end{defn}

\begin{prop}\label{LBhyp}
Given $(T,c_T),(T', c'_{T'})\in \cL(\bB)\subset \bB_\Omega$, let
$$ \cC_{\fM(T,T')}=\{ c_{\fM(T,T')} \,|\, \Tree[ .{$c(v_{\fM(T,T')})$} [ {$c_T(v_T)$} {$c_{T'}(v_{T'})$} ]] \in \cR \}\, , $$
where $v_T$, $v_{T'}$ are the roots of $T$, $T'$. 
The inclusion $\cL(\bB) \subset \bB_\Omega$ induces a weak subhypermagma structure on $\cL(\bB)$ with
\begin{equation}\label{weakhypM}
 \fM_{\cL(\bB)} ((T,c_T),(T', c'_{T'}))=\left\{ \begin{array}{ll} (\fM(T,T'), \{ c_{\fM(T,T')} \in \cC_{\fM(T,T')} \})   & \cC_{\fM(T,T')}\neq \emptyset \\
\emptyset & \text{otherwise}
\end{array} \right.  
\end{equation}
\end{prop}

One can give an analogous hypermagma formulation for the language of the
colored operad of theta roles assignments described in \cite{MarLar}, following the same construction. 

\subsection{Colored Merge} \label{MergeColSec}

As discussed in \cite{MarLar}, one can think of the coloring algorithm 
that selects the well-colored syntactic objects $\cL(\bB)$ and $\cL(\bB_{ext,{\rm EPP}})$,
among all the possible structures in $\fT_{\cS\cO_0}$ freely constructed by
Merge, as a filter in I-language that eliminates the ill formed structures:
the filter described here being the one that checks for the specifier-head-complement
structure and phases, and the EPP condition. Another similar filter, functioning
according to the same algebraic mechanism, is the one described in \cite{MarLar}
that checks for the proper assignment of theta roles and the theta criterion. 

This view as filters that select among the freely formed structures produced
by the free symmetric Merge may give the impression that there should be
a temporal relation between these two mechanism: {\em first} all structures
are freely formed, {\em then} they are filtered. However, this is not necessarily
the case. What the algebraic formalism shows is that one can {\em decouple}
these two computational mechanisms, which makes it possible to seperately
analyze them in a more precise and efficient way. However, in actual language
production, these operations of structure formation and filtering do not need
to occur in the time-ordering suggested. In fact, it was shown in \cite{MarLar}
that this description in terms of filtering is {\em equivalent} to a description in
which structures are filtered while they are formed, at each stage of the
Merge action on workspaces that leads to structure formation. While the
decoupled description of structure formation and filtering is best for 
separately analyzing the algebraic structures involved and their properties,
this ``combined" view is preferable for algorithmic implementations, as
it avoids the combinatorial explosion inherent in the free structure formation.
As shown in \cite{MarLar}, this combined view can be formulated in terms
of a ``colored Merge", that accounts for theta role assignments during
structure formation. We show here the analogous construction for the
case of head, complement, and phases discussed above. It is important
to keep in mind, though, that these two formulations are equivalent, so
describing the combined process as a colored Merge does not mean
that the additional structures of phases and theta roles impose constraints
on the free structure formation of the free symmetric Merge, which is
unchanged. It just means that the filtering by coloring can be done at each
stage of structure formation. The crucial property that makes this possible
is a peculiar characteristic of the colored operads of both phases and
theta theory: {\em all their bud system generators consist of a single vertex
with two leaves inputs and one output}. This is by no means typical, as
usually colored operad generators can involve larger structures
with multiple vertices. 

\begin{rem}\label{2ways} {\rm 
The fact that in both of these two different filters
(phases and theta roles) all the colored operad generators have a single 
vertex {\em implies} that a colored Merge reformulation is possible, 
hence that the description as filters after structure formation and the 
description as filters during structure formation are equivalent. }
\end{rem}

We discuss here also how the formulation in terms of colored Merge
is related to the hypermagma description of \S \ref{HypColorSec}. 

\smallskip

In the formulation of the Merge action on workspaces in Chapter~1 of \cite{MCB}, one
considers the vector space $\cV(\fF_{\cS\cO_0})$ spanned by the set $\fF_{\cS\cO_0}$
of forests whose connected components are syntactic objects in $\fT_{\cS\cO_0}$.
A product operation $\sqcup$ on $\cV(\fF_{\cS\cO_0})$ is given by the disjoint union of forests,
extended by linearity to the vector space. 
Two forms of coproduct are considered, both of the form
 \begin{equation}\label{Deltacut}
   \Delta(T)=\sum_{\underline{v}} F_{\underline{v}}  \otimes T/ F_{\underline{v}} = \sum_C \pi_C(T) \otimes \rho_C(T)\, , 
  \end{equation} 
where an admissible cut $C$ is a cut of a number of edges in the tree with the property that no two of them are
  on the same path from the root to a leaf, and extended to forests by multiplicativity $\Delta(F)=\sqcup_a \Delta(T_a)$
  for $F=\sqcup_a T_a$. The difference between the two versions of \eqref{Deltacut} is in the form of the
  quotient term $T/ F_{\underline{v}}$, respectively indicated by $T/^c F_{\underline{v}}$ (each component $T_{v_i}$
  of $F_{\underline{v}}$ is contracted to its root vertex) and $T/^d F_{\underline{v}}$ (the components $T_{v_i}$
  are removed and the largest full binary tree determined by what is left is obtained via some edge contractions
  that eliminate non-branching vertices). These two forms of the coproduct have different algebraic properties
  as discussed in Chapter~1 of \cite{MCB}, and linguistically they correspond to keeping a trace of movement
  (as needed at the CI interface)  or deleting it (as in Externalization). We will not explicitly distinguish between
  the two forms of the coproduct, except when addressing specific properties of each.  The form of the Merge action on
  workspaces is then given by a Hopf algebra Markov chain (in the sense of \cite{Diac}) of the form
\begin{equation}\label{MergeHMC}
 \cK = \sum_{S,S'} \fM_{S,S'} = \sum_{S,S'} \sqcup \circ (\cB\otimes {\rm id}) \circ \delta_{S,S'}\circ  \Delta \, ,
\end{equation}
where the sum is over pairs of syntactic objects, $\delta_{S,S'}$ is the indicator function supported
on the terms of the coproduct with $S\sqcup S'$ in the left-hand-side, and $\cB$ is the grafting operator
that grafts a forest to a common root. The sum is apparently an infinite sum, but it becomes a finite
sum when applied to any workspace, as $\delta_{S,S'}$ is equal to one on only finitely many terms and
zero otherwise.  

\smallskip

\begin{defn}\label{defXiPhi}
To simplify notation, we write $\bB_\Phi$ and $\cR_\Phi$ for the bud generating system $\bB_{ext,{\rm EPP}}$ of
Definition~\ref{modEPPlang} and for its local coloring rules $\cR_{ext}$, with 
$\cL(\bB_\Phi)$ indicating the resulting language $\cL(\bB_{ext,{\rm EPP}})$. (The subscript $\Phi$ is meant to
indicate that this is the coloring filter for the structure of phases.) We define $\Xi_\Phi$ as the set of triples
$(c, \{ c', c''\})$ such that there is a generator in $\cR_{ext}$ with output labelled by $c$ and inputs labelled by $c'$ and $c''$,
\begin{equation}\label{XiPhi1}
\Xi_\Phi =\{ (c, \{ c', c''\}) \,|\, \Tree[ .$c$ [ $c'$ $c''$ ] ]  \,\, \in \cR_\Phi \} \,. 
\end{equation}
Thus, $\Xi_\Phi$ contains the following triples
\begin{equation}\label{XiPhi}
\begin{array}{ll}
( \fc, \{ \fc, \fm \})\, , & 
( (\fs_\omega^\uparrow, \fz_{\omega'}^\downarrow), \{ ( \fh^\omega, \fz^\uparrow_\omega, \fs^\uparrow_\omega), \fz_\omega^\downarrow \} )\, , \\
( (\fh^{\omega'}, \fs_\omega^\uparrow) , \{ (\fh^{\omega'}, \fz_{\omega'}^\uparrow) ), (\fs_\omega^\uparrow, \fz_{\omega'}^\downarrow) \} ) \, , & ((  \fh^\omega, \fs^\uparrow_\omega), \{ ( \fh^\omega, \fz^\uparrow_\omega, \fs^\uparrow_\omega), \fz^\downarrow_\omega \} )\, , \\
( \fz^\downarrow_{\omega'}, \{ (  \fh^{\omega}, \fz^\uparrow_\omega), \fz^\downarrow_\omega \} ) &
( \fs^\downarrow_{\omega'}, \{ (  \fh^{\omega}, \fz^\uparrow_\omega), \fz^\downarrow_\omega \} ) \\
( \fz^\downarrow_{\omega''}, \{ (  \fh^{\omega'}, \fs^\uparrow_\omega), \fs^\downarrow_\omega \} ) &
( \fs^\downarrow_{\omega''}, \{ (  \fh^{\omega'}, \fs^\uparrow_\omega), \fs^\downarrow_\omega \} ) \\
( \fm, \{ (  \fh^{\omega}, \fz^\uparrow_\omega), \fz^\downarrow_\omega \} ) & (\fm , \{ (  \fh^{\omega'}, \fs^\uparrow_\omega), \fs^\downarrow_\omega \} ) 
\end{array}
\end{equation}
subject to the constraints of Definition~\ref{EPPgens}, and where, in the last cases,
in the label $(\fh^{\omega'}, \fs^\uparrow_\omega)$ the $\omega$ and $\omega'$ may be either the
same or different, with the constraint of Definition~\ref{EPPgens} if different. The set $\Xi_\Phi$ also contains the
same triples with one or both of the inputs carrying an item $\alpha\in \cS\cO_0$ with $\omega\in \Omega_{\fh,{\rm lex}}$,
and also the additional triples of the form
$( \fs^\downarrow_{\omega'}, \{  \fc^\downarrow_\omega, (1,\fm)  \})$ with $\fc\in \{ \fs, \fz \}$ (with
the possibility of additional $( \fs^\downarrow_{\omega'}, \{  \fc, (1,\fm)  \})$ as in \eqref{genT1mm}, \eqref{genT1mh} 
as discussed in Remark~\ref{nomovemh}).
\end{defn}

As in \cite{MarLar}, we define the colored grafting operator $\cB^c$ as the grafting that assigns a color $c$ to
the root,
$$ \cB^c(F): =\Tree[.$c$ $T_1$ $T_2$ $\cdots$ $T_N$ ] $$
which, in the case of two components $F=T \sqcup T'$, can also be written as
$$ \cB^c(F)=\Tree[.$c$ $T$ $T'$ ] =: \fM^c(T,T') \, . $$

We obtain the following colored Merge operator exactly as in the case of theta coloring in \cite{MarLar}.

 \begin{prop}\label{ColorMerge}
 Let $\cV(\fF(\cL(\bB_\Phi)))$ be the vector space spanned by the set $\fF(\cL(\bB_\Phi))$ of 
 forests whose components are in $\cL(\bB_\Phi)$. The colored Merge is given by
 \begin{equation}\label{colMerge}
 \fM^c_{S,S'}=\sqcup \circ (\cB^c \otimes {\rm id}) \circ \delta^c_{\{ c_S, c_{S'} \}} \delta_{S,S'} \circ \Delta \, ,
 \end{equation}
 where 
 \begin{equation}\label{deltaccc}
 \delta^c_{\{ c_S, c_{S'} \}}=\left\{ \begin{array}{ll} 1 & (c, \{ c_s, c_{S'} \})\in \Xi_\Phi \\
 0 & (c, \{ c_s, c_{S'} \})\notin \Xi_\Phi
 \end{array}\right.
 \end{equation}
 with $c_S$ and $c_{S'}$ the colors at the root vertices of $S, S'\in \cL(\bB_\Phi))$. 
 It is a linear map on $\cV(\fF(\cL(\bB_\Phi)))$. 
  \end{prop}
  
 By construction of \eqref{colMerge}, the result of structure building by the operators
 $\{ \fM^c_{S,S'} \,|\, c\in \Omega_\Phi, \,\, S,S'\in \cL(\bB_\Phi) \}$, starting from pairs
 $S,S' \in \Omega_{\Phi,{\rm lex}}$ (lexical items with labels in $\Omega_{\Phi,f}$)
 gives syntactic objects in $\cL(\bB_\Phi$ and all those can be obtained in this way,
 {\em because} all the generators in $\cR_\Phi$ have a single vertex so they can
 all occur in $\fM^c_{S,S'}$. This ensures that all of $\cL(\bB_\Phi))$ can also be
 constructed by the colored Merge, not only by the colored operad, hence giving an
 equivalent reformulation as discussed above. 
 
 One can also see the relation between the colored Merge formulation and
 the hypermagma formulation described above, by considering the colored
 version of \eqref{MergeHMC}, of the form
 \begin{equation}\label{colMergeHMC}
 \cK_\Phi = \sum_{S,S' \in \cL(\bB_\Phi)} \sum_{c\in \Omega_{\Phi,f}} \fM^c_{S,S'}  \, .
\end{equation}

\begin{prop}\label{colMhyp}
For a set $A$, with $\cP(A)=2^A$ its power set, and $\cV(A)$ the vector space spanned by $A$,
we write $\Sigma: \cP(A) \to \cV(A)$ with $\Sigma(B):=\sum_{x\in B} x \in \cV(A)$. 
The colored Merge in the form \eqref{colMergeHMC} can be equivalently written as
 \begin{equation}\label{colMergeHMC2}
  \cK_\Phi = \sqcup \circ \Sigma \circ (\fM_{\bB_\Phi}\otimes {\rm id}) \circ \Pi_{(2)} \circ \Delta \, ,
 \end{equation}
 where $\fM_{\cL(\bB_\Phi)}$ is the hypermagma operation of \eqref{weakhypM}, the projection $\Pi_{(2)}$
 is the linear projection onto the subspace of $\cV(\fF(\cL(\bB_\Phi)))\otimes \cV(\fF(\cL(\bB_\Phi)))$
 spanned by elements of the form $(T \sqcup T') \otimes F$ with two components in the left channel
 of the coproduct.
\end{prop}

\proof The equivalence of \eqref{colMergeHMC} and \eqref{colMergeHMC2} follows from
the fact that the set $ \cC_{\fM(T,T')} $ of \eqref{weakhypM} can be identified with
$$ \cC_{\fM(T,T')} =\{ (\fM(T,T'), c) \,|\, (c(v_{\fM(T,T')}), \{ c_T(v_T), c'_{T'}(v_{T'}) \}) \in \Xi_\Phi \} $$
hence the summation
$$ \sum_{T,T'\in \cL(\bB_\Phi)} \sum_{c \in \Omega_{\Phi,f}} \delta^c_{c_T(v_T), c'_{T'}(v_{T'})} \fM^c(T,T') $$
is the same as the summation 
$$ \sum_{T,T'\in \cL(\bB_\Phi)} \Sigma(\cC_{\fM(T,T')}) $$
where the sum over $T,T'\in \cL(\bB_\Phi)$ is a finite sum when restricted to $T,T'$ in the
accessible terms of a given forest $F\in \fF(\cL(\bB_\Phi))$.
\endproof

An important part of the Merge model is the structure of Internal Merge as a composition
$\fM_{T_v,T/T_v}\circ \fM_{T_v,1}$ of two Merge operations, where in the first one applied,
one formally takes $S=T_v$, an accessible term of one of the components of the workspace,
and $S'=1$ the unit of the magma of syntactic objects, the formal ``empty tree". The first operation
$\fM_{T_v,1}$ results in extracting the accessible term $T_v$ and depositing it, alongside the
remainder term $T/T_v$ in the resulting workspace, as usually described in the linguistics
literature, where they are then merged together by the second part $\fM_{T_v,T/T_v}$ of the operation.

As we discussed in the case of theta theory in \cite{MarLar} this particular form of
Internal Merge has important consequences for the rules governing movement on
colored structure: for example, in the case of theta theory, it is responsible for the
dichotomy (also called duality) in semantics between Internal and External Merge. 
We analyze in \S \ref{PhaseMoveSec} how this formulation of Internal Merge 
relates to rules about movement and the structure of phases. 

This discussion in \S \ref{PhaseMoveSec}, and the following \S \ref{PhaseThetaSec} 
addresses an issue formulated in the last section of \cite{MarLar} about movement
in theta theory and its compatibility with phases.

\section{Phases Coloring and Movement} \label{PhaseMoveSec}

In Minimalism there are strong constraints on movement by Internal Merge and the structure of phases.
In the SMT formulation, these constraints are not built into the free structure building operation of Merge,
hence they need to be realized at the level of the filtering that accounts for well formed phase structure.
In view of the formulation in terms of colored Merge discussed above, this means that constraints on
movement by Internal Merge that arise from phases have to be implemented in the formulation of
Internal Merge $\fM_{T_v,T/T_v}\circ \fM_{T_v,1}$, 
where both $\fM_{T_v,1}$ and $\fM_{T_v,T/T_v}$
have to be realized by the colored Merge of Proposition~\ref{ColorMerge}.

In order to investigate how this works, we need to fit into the 
colored Merge of Proposition~\ref{ColorMerge} the additional case of a Merge
$\fM_{T_v,1}$, where one of the objects is the empty tree, the unit of the magma.
This case is not directly accounted for by \eqref{colMergeHMC2}, since it is not
in the range of the projector $\Pi_{(2)}$. As in \cite{MCB}, one needs to add 
the case of the map $\sum_S \fM_{S,1} = \sqcup \circ \Pi_{(1)} \circ \Delta$,
where $\Pi_{(1)}$ consists of the projection onto the terms of the coproduct
with a single component in the left channel. In the colored case, 
one then replaces $\sum_S \fM_{S,1}$ with 
$$\sum_{S\in \cL(\bB_\Phi)} \sum_c \delta^c_{c',c''} \fM^c_{S,1} \, , $$
where $\delta^c_{c',c''}$ represents the local coloring rule for merging
with a colored $(1,c'')$ when the coloring of $S$ has $c_S(v_S)=c'$. 

Since $1$ is the unit of the magma, $\fM(T,1)=T$, so merging with $1$
does not alter the structure $T$. Thus, the unit $1$ does not participate
in building the specifier-head-complement structure, which means that
the only reasonable coloring option is $(1,\fm)$. This argument is the
same as in the theta theory case of \cite{MarLar}, where for the same structural reason,
the unit $1$ does not participate in the assignment of theta roles, hence
it can only carry the coloring $(1,\theta_0)$ of a non-theta position.

There is, however, a difference in behavior between the case of the
coloring that accounts for phases and the coloring for theta roles. 

\begin{rem}\label{Mphasesrem}{\rm 
 The basic rules that are known to govern compatibility of movement 
 by Internal Merge with respect to phases can be summarized in the
 following way. 
 \begin{enumerate}
 \item  Internal Merge (IM) can move from the interior of the 
 phase to the edge of the phase, Spec-of-Phase. 
 \item IM can also move within the interior or the phase as IM({\tt EA}, {\tt INFLP})
 and IM(XP, RootP), to Spec-of-{\tt INFL} and Spec-of-Root positions (simple
clauses, raising, exceptional case-marking (ECM))
 \item IM does not move from the interior of lower phases once the phases are
 completed (for example, movement to Spec-of-Root cannot be further accessed
 from higher phases because it is not in Spec-of-Phase position). 
\end{enumerate} }
\end{rem}

These restrictions on the action of Internal Merge have to be
implemented via the coloring rules of $\cL(\bB_\Phi)$ and the
colored Merge formulation discussed in \S \ref{MergeColSec}.

The key to obtain this is the rules for the colored version of the Merge
operation $\fM_{T,1}$ involved in Internal Merge. 

\begin{prop}\label{MT1}
The additional color in $\Omega_\Phi$ of the form $(1,\fm)$ and the corresponding
generators in $\cR_\Phi$ of the form \eqref{gens1m} involving this color are the
part of the colored operad that accounts in $\cL(\bB_\Phi)$ for the results of movement  by Internal Merge.
In order to ensure that this type of movement satisfies the rules listed above, 
it is necessary to restrict the pairs $(\omega,\omega')$ that occur in \eqref{gens1m} 
accordingly. Given a lexical head $\ell=(\alpha,{\rm lex})$ consider the set $\Omega(\ell)$ 
of functional heads $\ell'$ that can occur above $\ell$, in the extended projection of $\ell$. 
We take $\Omega(\ell)$ to be partially ordered by the order in which they can occur.
The generators \eqref{gens1m} that give IM
movement satisfying the conditions listed above are of the form
\begin{equation}\label{gens1mell}
T^\fs_{\fc,(1,\fm)} = \Tree[ .$\fs^\downarrow_{{\rm RootP}}$ [ $\fc_{(\alpha,{\rm lex},V)}$ $(1,\fm)$ ] ] \, ,
\end{equation}
which moves to Spec-of-RootP at $v_\ell$ and
\begin{equation}\label{gens2mell}
T^\fs_{\fc,(1,\fm)} = \Tree[ .$\fs^\downarrow_{\omega'}$ [ $\fc_{\omega}$ $(1,\fm)$ ] ] \, ,
\end{equation}
where $\omega,\omega' \in \Omega(\ell)$ with $\omega' > \omega$ and either
\begin{itemize}
\item both $\omega$ and $\omega'$ are phase-heads
\item $\omega'=\mathtt{INFL}$ or $\omega'={\rm Root}$
\end{itemize}
\end{prop}

This follows directly by seeing that repeated use of generators \eqref{gens1mell} and \eqref{gens2mell}
as stated give rise to movement as described in the cases listed above. 

The partial ordering on $\Omega(\ell)$ reflects ordering such as
$\mathtt{C}>\mathtt{INFL}>\mathtt{v}^*$ or $\mathtt{C}>\mathtt{INFL}>\mathtt{v}$. This partial
ordering can be seen as part of the choice of a cartographic model,
in the sense of Definition~\ref{cartodef} and Definition~\ref{KbarePhi}.

\begin{rem}\label{PIC} {\rm
In particular, the Phase Impenetrability Condition (PIC) follows from
the set of generators of $\cL(\bB_\Phi)$ if the generators \eqref{gens1m} 
are restricted to satisfy the conditions of Proposition~\ref{MT1}. }
\end{rem}

\subsection{Labeling}

In \cite{ChomskyPP} and \cite{ChomskyPPext} a labeling algorithm is discussed.
Section~1.15 of \cite{MCB} discussed this labeling algorithm in the framework
of the mathematical formulation of Merge. This algorithm assigns labeling to the internal 
vertices of syntactic objects $T\in {\rm Dom}(h) \subset \fT_{\cS\cO_0}$, by labeling
vertices on a path $\gamma_\ell$ by $\ell$. The algorithm also assigns a labeling
to {\em some} syntactic objects $T=\fM(T_1,T_2)$, where $T_1,T_2\in {\rm Dom}(h)$ but
$T$ is only in the hypermagma $\fH_{\cS\cO_0}$ and does not, a priori, 
have a unique assignment of a head function fully determined by those of $T_1$ and $T_2$ (exocentric).
Cases of this form that can still be labelled by the labeling algorithm (hence for which a head
assignment giving $T\in {\rm Dom}(h)$ becomes possible) are those for which
raising of either $T_1$ or $T_2$ gives 
a structure that can be seen to be in ${\rm Dom}(h)$, namely that has a uniquely assigned head, and this in turn
induces a choice of head on $T$ itself which resolves the ambiguity in the hypermagma $\fH_{\cS\cO_0}$.

The simplest case of this (mentioned in Section~1.15 of \cite{MCB}) would be if 
either $\fM(T_1,T/^c T_1)$ or $\fM(T_2, T/^c T_2)$ is in ${\rm Dom}(h)$,
which necessarily means $h_{\fM(T_1,T/^c T_1)}=h_{T_2}$ (meaning that it 
assigns to the root vertex of $\fM(T_1,T/^c T_1)$ the same value that $h_{T_2}$
assigns to the root vertex of $T_2$), respectively, 
$h_{\fM(T_2, T/^c T_2)}=h_{T_1}$. This is usually expressed in terms of ECP (empty category principle)
but it follows here directly from the definition of head function.  In this case one assigns
labeling to the root vertex $v$ of $\fM(T_1,T_2)$ as   $h_{\fM(T_1,T/^c T_1)}(v)$ (respectively, 
$h_{\fM(T_2, T/^c T_2)}(v)$. See Definitions~1.15.1 and 1.15.2 of \cite{MCB}.

One should also allow the possibility that raising yields an object in ${\rm Dom}(h)$
in a more general form, for instance with some $T'\in {\rm Dom}(h)$ and the
syntactic object $\tilde T=\fM(T_1, \fM(T',T/^c T_1)) \in {\rm Dom}(h)$. This implies that
$\fM(T',T/^c T_1)$ has a head function $h_{\fM(T',T/^c T_1)}$ induced by $h_{\tilde T}$,
which agrees either with $h_{T'}$ or with $h_{\fM(T',T/^c T_1)}=h_{T_2}$ (in the same
sense as above). This again means that the root vertex of $\fM(T',T/^c T_1)$, which
is the same as the root vertex of $T$, is assigned a label, hence $T$ gets a labeling too.
The argument is the same as for the simpler case of $\fM(T_1,T/^c T_1)$ discussed
above: the only reason for considering this more general case is to account for the restrictions
on movement imposed by the coloring rules (namely movement to the edge of phase position,
and the other cases discussed above).

We show here that the labeling of \cite{ChomskyPP} and \cite{ChomskyPPext}  
is also directly implemented by the coloring in terms of the generators of the colored operad 
of phases. The main point here is that, while just knowing the head functions of $T_1$ and
$T_2$ may not suffice to resolve the ambiguity of $T=\fM(T_1,T_2) \in \fH_{\cS\cO_0}$ to
a unique head assignment, taking into account the form of the generators of $\cR_\Phi$,
the local coloring rules applied to the structure obtained after raising resolve the ambiguity
and select a unique possibility for the head structure and the coloring of $T=\fM(T_1,T_2)$. 

\begin{prop}\label{labeling}
For all syntactic objects $T\in \cL(\bB_\Phi)$, the decomposition into
generators in $\cR_\Phi$ uniquely determines a labeling of the non-leaf
vertices of $T$ and this labeling matches the one prescribed by the
labeling algorithm of \cite{ChomskyPP} and \cite{ChomskyPPext}.
\end{prop}

\proof
For any $T\in \cL(\bB_\Phi)$, the labeling assigned by the labeling algorithm 
of  \cite{ChomskyPP} and \cite{ChomskyPPext} can be recovered from the
coloring of the vertices by the colored operad insertions realization of $T$.

We make the following labeling assignment at the vertex of these generators:
\begin{equation}\label{T2genLab} 
T_{\fh^\omega\fz,\fz}^\fc := \Tree[ .$\fc$  [ {$(\fh^\omega, \fz^\uparrow)$} $\fz^\downarrow$ ] ]  \mapsto \fh^{\omega} \, , \ \ \ \ \ 
 T_{\fh^\omega\fs,\fs}^\fc :=  \Tree[ .$\fc$ [  {$(\fh^\omega, \fs^\uparrow)$} $\fs^\downarrow$ ] ]  \mapsto 
\fh^{\omega} \, , 
\end{equation} 
\begin{equation}\label{Tomega1genLab}
T_{\fh^\omega\fz\fs,\fz}^{\fs\fz} = \Tree[ .{$(\fs_\omega^\uparrow, \fz_{\omega'}^\downarrow)$} [ {$(\fh^\omega,\fz_\omega^\uparrow, \fs_\omega^\uparrow)$} $\fz_\omega^\downarrow$ ]  ]   \mapsto \fh^\omega \, , \ \ \ \  
 T_{\fh^{\omega'}\fz,\fs\fz}^{\fh^{\omega'}\fs} = \Tree[ .{$(\fh^{\omega'}, \fs_{\omega}^\uparrow)$} [ {$(\fh^{\omega'}, \fz_{\omega'}^\uparrow)$} {$(\fs_\omega^\uparrow, \fz_{\omega'}^\downarrow)$} ] ]  \mapsto \fh^{\omega'} 
 \end{equation}
\begin{equation}\label{Tomega2genLab}
T^{\fh^\omega\fs}_{\fh^\omega\fz\fs,\fz}= \Tree[ .{$(\fh^\omega, \fs_\omega^\uparrow)$} [ {$(\fh^\omega, \fz_\omega^\uparrow, \fs_\omega^\uparrow)$}  $\fz_\omega^\downarrow$ ] ] 
 \mapsto \fh^\omega \, ,  \ \ \ \ 
T^\fh_{\fh,\fm} = \Tree[ .{$(\fh^\omega, \fz^\uparrow, \fs^\uparrow)$} [ {$(\fh^\omega, \fz^\uparrow, \fs^\uparrow)$} $\fm$ ] ] \mapsto \fh^\omega \, ,
\end{equation}
We also set
\begin{equation}\label{Tomega3genLab}
T^\fh_{\fh,\fm} = \Tree[ .{$\fc$} [ {$\fc$} $\fm$ ] ] \mapsto \fh^\omega \, ,  \ \ \text{ whenever the color $\fc=(\fh^\omega,\fc')$ 
contains a head label $\fh^\omega$} 
\end{equation}
This assignment of labeling to the vertex, for the generators in \eqref{T2genLab}  and \eqref{Tomega1genLab}, 
is obvious because they are on the path $\gamma_\ell$ for the head $\ell$ colored by $\fh^\omega$, so
we have to assign $\fh^\omega$ (or equivalently $\ell$) to the vertex, and this agrees with what
prescribed by the labeling algorithm of \cite{ChomskyPP} and \cite{ChomskyPPext} and 
Section~1.15 of \cite{MCB}. Then we need to assign a labeling to the vertex in the remaining generators 
\begin{equation}\label{gens1mLab}
T^\fz_{\fz,\fm} = \Tree[ .$\fz_\omega^\downarrow$ [ $\fz_\omega^\downarrow$ $\fm$ ] ]  \, ,  \ \ \ \ 
T^\fs_{\fs,\fm} = \Tree[ .$\fs_\omega^\downarrow$ [ $\fs_\omega^\downarrow$ $\fm$ ] ]   \ \ \ \text{ and } \ \ \ 
T^\fs_{\fs,(1,\fm)}:= \Tree[ .$\fs^\downarrow_{\omega'}$ [ $\fc^\downarrow_\omega$ $(1,\fm)$ ] ]  \, .  
\end{equation}
This labeling is assigned to be the same $\fh^\omega$ label carried by the generator 
of the form \eqref{T2genLab} whose root output is matched in $T$, via  
the operad composition, with the non-$\fm$-labelled leaf of the generator of the form \eqref{gens1mLab}.
In the case of the first two generators in \eqref{gens1mLab} 
this labeling assignment matched the prescription of the labeling algorithm of 
of \cite{ChomskyPP} and \cite{ChomskyPPext} and  Section~1.15 of \cite{MCB} because 
these vertices are on the $\gamma_\ell$ path of the head $\ell$ colored by $\fh^\omega$.

The last case of \eqref{gens1mLab} is what also accounts for the cases of the
labeling algorithm of of \cite{ChomskyPP} and \cite{ChomskyPPext} where the
labeling is assigned via raising. To see that, suppose we have a syntactic object
$T\in \fT_{\cS\cO_0}$ with $T=\fM(T_1,T_2)$, such that $T_1,T_2\in {\rm Dom}(h)$ but 
for which one only has $\fM(T_1,T_2) \in \fH_{\cS\cO_0}$. This is the typical situation of exocentric structures
like $\{ XP, YP \}$, which in terms of the coloring and generators in $\cR_\Phi$,
correspond to the top nodes of $T_1$ and $T_2$ being both of the form \eqref{T2genLab},
where $\fc_{\omega'}^\downarrow$ is either a $\fz^\downarrow$ or a $\fs^\downarrow$, but
there is no generator with leaves input labels $\{ \fz_\omega^\downarrow, \fs_{\omega'}^\downarrow \}$
or $\{ \fz_\omega^\downarrow, \fz_{\omega'}^\downarrow \}$ or $\{ \fs_\omega^\downarrow, \fs_{\omega'}^\downarrow \}$,
hence $\fM(T_1,T_2)$ cannot be colored by $\cR_\Phi$ local coloring moves, and is not in $\cL(\bB_\Phi)$.

However, let's consider a situation where $T=\fM(T_1,T_2)$ is not in ${\rm Dom}(h)$ but a structure
obtained by raising $T_1$ is. We assume here that both $T_1,T_2\in \cL(\bB_\Theta)$,
so they both have a decompositions into generators in $\cL(\bB_\Theta)$. The fact that
$T=\fM(T_1,T_2)$ is not in ${\rm Dom}(h)$ means that we cannot assign a good coloring to the input 
and outputs of the root vertex of $T=\fM(T_1,T_2)$. 

Because of the form of the generators in $\cR_\Phi$, the smallest
structure involving raising of $T_1$ that can be colorable would be of the form $\fM(T_1, \fM(\alpha, T/^c T_1))$
where $\alpha$ is a (functional) head.  This means that the root vertex of $\fM(T_1, \fM(\alpha, T/^c T_1))$
must correspond to a generator of the form 
\begin{equation}\label{IMlabel1}
\Tree[ .$\fc$ [ $\fs_\omega^\downarrow$ {$(\fh^{\omega'}, \fs_{\omega}^\uparrow)$} ] ] 
\end{equation}
where $\omega'\succeq \omega$, with $\fc$ either a $\fz^\downarrow$ or an $\fs^\downarrow$ of an $\fm$. 
The output at the root of $T_1$ is labelled by some other $\tilde\fc^\downarrow$ 
(which is either a $\fz$ or a $\fs$) and is attached to a generator of the form
\begin{equation}\label{IMlabel2}
 \Tree[ .$\fs^\downarrow_\omega$ [ $\tilde\fc^\downarrow$ $(1,\fm)$ ] ]  
\end{equation} 
whose root matched the $\fs^\downarrow_\omega$-labelled input of the generator \eqref{IMlabel1}.
We can assign to the vertex of \eqref{IMlabel2} the same label as the vertex that appears below the 
$\tilde\fc^\downarrow$-labelled leaf of \eqref{IMlabel1}.
The coloring of inputs and output of the generator \eqref{IMlabel1} 
in turn implies that the root vertex of $\fM(\alpha, T/^c T_1)$ must be colored by 
a generator that has output $(\fh^{\omega'}, \fs_\omega^\uparrow)$, which has to be of the form
\begin{equation}\label{IMlabel3}
\Tree[ .{$(\fh^{\omega'}, \fs_{\omega'}^\uparrow)$} [ {$(\fh^{\omega'},\fz_{\omega'}^\uparrow, \fs_{\omega'}^\uparrow)$} {$\fz_{\omega'}^\downarrow$} ] ] \ \ \  \text{ or } \ \ \  \Tree[ .{$(\fh^{\omega'}, \fs_{\omega}^\uparrow)$} [ {$(\fh^{\omega'}, \fz_{\omega'}^\uparrow)$} {$(\fs_\omega^\uparrow, \fz_{\omega'}^\downarrow)$} ] ]
\end{equation}
with $\fh^{\omega'}$ the functional head $\alpha$, depending on whether $\omega$ and $\omega'$ agree or not. In turn this implies that the root vertex of $T/^c T_1$
needs to be labelled by a generator of the form 
\begin{equation}\label{IMlabel4}
\Tree[ .$\fz_{\omega'}^\downarrow$ [ $\fs_{\tilde\omega}^\downarrow$ {$(\fh_{\tilde\omega'}, \fs_{\tilde\omega}^\uparrow)$} ] ] 
\ \ \  \text{ or } \ \ \  \Tree[ .{$(\fs_\omega^\uparrow, \fz_{\omega'}^\downarrow)$} [ $\fz_\omega^\downarrow$ {$(\fh^\omega,\fz_\omega^\uparrow, \fs_\omega^\uparrow)$}  ]  ] 
\end{equation}
respectively, according to the two cases of \eqref{IMlabel3}. In the first case $\fh_{\tilde\omega'}$ is the head of $T_2$
and it labels the root vertex of $T$, so this would give a consistent coloring and labeling to $T=\fM(T_1,T_2)$ itself,
hence providing a resolution of the head choice for $T\in \fH_{\cS\cO_0}$, consistent with the heads of $T_1$ and $T_2$,
this choice means that $T$ can in fact fit in the section ${\rm Dom}(h)$ of the hypermagma.
The object $T$ is then also in $\cL(\bB_\Phi)$, with 
$\tilde\fc^\downarrow =\fs_{\tilde\omega}^\downarrow$ in \eqref{IMlabel2}. In the second case, similarly, the
root vertex of $T$ gets labelled by $\fh^{\omega'}$ and this gives a consistent coloring and labeling
to $T=\fM(T_1,T_2)$ itself which is then in ${\rm Dom}(h)$ and in $\cL(\bB_\Phi)$, with 
$\tilde\fc^\downarrow =\fz_\omega^\downarrow$ in \eqref{IMlabel2}. This shows that, in fact, the only cases
where $\fM(T_1, \fM(\alpha, T/^c T_1)) \in \cL(\bB_\Phi)$ occur when already $T=\fM(T_1,T_2) \in \cL(\bB_\Phi)$,
since we have shown that, in such cases, assigning a head function and a good coloring to $T=\fM(T_1,T_2)$  
can be achieved directly in terms of the head function on $T_1$ and $T_2$, 
without having to take into account also the additional information on the possible colorings of $\fM(T_1, \fM(\alpha, T/^c T_1))$. 
\endproof

\smallskip

\begin{rem}\label{ECPrem}{\rm 
One can then make an observation about the ECP (empty category principle). The trace of
movement should be invisible to labeling. On the other hand, the coloring of the generators
in $\cR_\Phi$ does assign a color $\tilde\fc^\downarrow$ to the trace of movement, 
which can be either a $\fs^\downarrow$ or a $\fz^\downarrow$
according to \eqref{IMlabel4} (with the additional possibilities as in \eqref{genT1mm}, \eqref{genT1mh},
as in Remark~\ref{nomovemh}). However, in all cases (except the possible additional generator
\eqref{genT1mh}) this color associated to the position from which IM movement is performed
does not participate into the determination
of labeling, since in both cases in \eqref{IMlabel4} (as well as in \eqref{genT1mm}) 
the labeling of the node is determined 
only by the $\fh$ color at the other leaf input. Thus, the fact that movement changes the
coloring at the root of the accessible term being moved, from $\tilde\fc^\downarrow$ 
to $\fs_\omega^\downarrow$, by effect of the generator \eqref{IMlabel2}, does not affect
in any way the labeling algorithm. (The case of the possible generator \eqref{genT1mh} is
discussed separately in \S \ref{2ObjMoveSec}.) }
\end{rem}

\section{Phases coloring and Theta coloring: transductions of colored operads} \label{PhaseThetaSec}

 A well formed parsable syntactic object has both a consistent system of phases
 and a correct assignment of theta roles. Thus, both filtering have to apply to the
 same structure. This means that two different colored operads need to be
 compared and viable structures should be those that meet both coloring
 requirements, according to both sets of generators simultaneously. Equivalently,
 in terms of the formulation via a colored Merge, the well-formed objects should
 have a derivation according to both types of colored Merge, for phases and
 for theta roles. This in principle may be interpreted in different ways: for
 example, it is a priori different to require that 
 a single colored Merge simultaneously implements both filters
 or that there is a derivation according to the colored Merge of phases 
 and a derivation according to the colored Merge for theta roles, but these two 
 derivations need not be the same. In terms of colored operads, one can require
 that a singled colored operad with single-vertex bud generators implements 
 both filters for phases and theta roles simultaneously, or that the two
 colored operads that select for these two properties are in an appropriate 
 sense both obtainable as specializations from a single colored operad. 
 
 Ultimately, a good model of the relation between these two colored
operads and languages, $\cL(\bB_\Phi)$ and $\cL(\bB_\Theta)$, 
for the bud systems $\bB_\Phi$ and $\bB_\Theta$ of phases and theta roles, 
should rely on empirical evidence on when phases and theta roles in Merge
derivations do or do not not move in coordinated ways, 
for example, in cases like the more complex theta role assignments of languages 
such as Mandarin, and their relation to the phase structure. We will present here
a proposal for how to combine the filters for theta roles and for phases. 

\smallskip

The key idea here is that we want viable structures to ``pass both filters" for
theta roles and for phases. This would mean that such structures have to
be in the intersection $\cL(\bB_\Phi)\cap \cL(\bB_\Theta)$. However, to make
sense of this intersection, both $\cL(\bB_\Phi)$ and $\cL(\bB_\Theta)$ have to
be realized as subsets of a common larger set. However, we constructed
$\cL(\bB_\Phi)$ and $\cL(\bB_\Theta)$ as different colored operads,
each with its own bud generating system, hence there is no common
ambient set where they would intersect. Thus, we need a more careful
formulation of what it means that a syntactic object ``passes both filters". It turns
out that the right way to express the idea that such syntactic objects are in the
intersection $\cL(\bB_\Phi)\cap \cL(\bB_\Theta)$ is to ``pull back" 
$\cL(\bB_\Phi)$ and $\cL(\bB_\Theta)$ to a correspondence and intersect them there.
This idea is frequently used, and is for example the key to the notion of
transducers and transduction in the theory of formal languages. We will
discuss in this section how to formulate it in the case of colored operads and
their bud generating systems.

\smallskip

It is useful to start by reviewing how one compares different colored operads,
especially those obtained via bid generating systems. There is the usual notion
of morphisms of colored operads. 

\begin{defn}\label{opmorph}
A morphism $\varphi: \fO \to \fO'$ of colored operads consists of a map $\varphi_c: \Omega \to \Omega'$ between
the sets of colors and maps $\varphi_n: \fO(a, (a_1, \ldots, a_n)) \to \fO(\varphi_c(a), (\varphi_c(a_1), \ldots \varphi_c(a_n)))$
between the operad sets, that are compatible with the operad composition operations, 
through the appropriate commutative diagrams. 
In particular, a morphism of ordinary operads (single-colored) $\varphi: \cO \to \cO'$ is a collection of maps
$\varphi_n: \fO(n) \to \fO'(n)$ compatible with the operad composition operations. 
\end{defn}

On the basis of this, for example, one can expect that there may be a colored operad $\tilde\fO$
that maps to both $\fO_{\Omega_\Phi,\bB_\Phi}$ and $\fO_{\Omega_\Theta, \bB_\Theta}$ 
$$ \xymatrix{ & \tilde\fO_{\tilde\Omega_{\Phi,\Theta},\tilde\bB_{\Phi,\Theta}} \ar[dl]^{\varphi} \ar[dr]_{\varphi'} & \\
 \fO_{\Omega_\Phi,\bB_\Phi} & & \fO_{\Omega_\Theta, \bB_\Theta} } $$ 
 where the structures that simultaneously satisfy both filtering constraints for phases and
 theta roles would be identified with
 $\varphi^{-1}(\cL(\bB_\Phi)) \cap \varphi'^{-1}(\cL(\bB_\Theta))$.

A somewhat more flexible model comes from generalizing 
morphisms of colored operads to correspondences between colored operads
with bud generating systems. These generalize the usual notion of transduction in formal languages. 

\begin{defn}\label{optransduce}
Suppose given two colored operads $\fO_{\Omega,\bB}$ and $\fO'_{\Omega', \bB'}$ 
with bud generating systems $\bB=(\fO,\Omega,\cR,\cI,\cT)$ and $\bB'=(\fO',\Omega',\cR',\cI',\cT')$.
Let $\hat\Omega$ be a set with maps $\pi: \hat\Omega \to \Omega$ and $\pi': \hat\Omega \to \Omega'$, 
and let $\tilde\fO$ be an ordinary operad with
operad homomorphisms $\varphi: \tilde\fO \to \fO$ and $\varphi': \tilde\fO \to \fO'$. 
Consider a bud generating system $\tilde\bB=(\tilde\fO, \tilde\Omega, \tilde\cR, \tilde\cI, \tilde\cT)$, with
$\tilde\cI =\pi^{-1}(\cI)\cap \pi'^{-1}(\cI')$ and $\tilde\cT = \pi^{-1}(\cT)\cap \pi'^{-1}(\cT')$ and $\tilde\Omega=\tilde\cI \cup \tilde\cT \subset \hat\Omega$, with a given finite set of generators $\tilde\cR\subset \tilde\bB_{\tilde\Omega}(\tilde\fO)$, with
$\tilde\bB_{\tilde\Omega}(\tilde\fO)(n)=\tilde\Omega \times \tilde\fO(n) \times \tilde\Omega^n$, as in \eqref{BudO}.
Let $\tilde\fO_{\tilde\Omega,\tilde\bB} \subset \tilde\bB_{\tilde\Omega}(\tilde\fO)$ be the colored operad
generated by the bud system $\tilde\bB$, and $\cL(\tilde\bB)$ the associated language.
The maps $\pi, \pi'$ restricted to $\tilde\Omega$ extend the operad morphisms 
$\varphi: \tilde\fO \to \fO$ and $\varphi': \tilde\fO \to \fO'$ to morphisms of the colored operads 
$$ \varphi: \tilde\bB_{\tilde\Omega}(\tilde\fO) \to \bB_{\Omega}(\fO) \ \ \ \text{ and } \ \ \  
\varphi': \tilde\bB_{\tilde\Omega}(\tilde\fO) \to \bB'_{\Omega'}(\fO') \, . $$
A {\em transduction of colored operads} $\bT: \fO_{\Omega,\bB} \to \fO'_{\Omega', \bB'}$ 
consists of data $(\tilde\bB, \varphi, \varphi')$ such that
\begin{equation}\label{transductionoperads}
 \varphi'(\varphi^{-1}(\fO_{\Omega,\bB}) \cap \tilde\fO_{\tilde\Omega,\tilde\bB}) \subset \fO'_{\Omega', \bB'} \, . 
\end{equation} 
This induced a {\em transduction of languages} $\bT:  \cL(\bB) \to \cL(\bB')$ if \eqref{transductionoperads}
restricts to
$$ \varphi'(\varphi^{-1}(\cL(\bB)) \cap \cL(\tilde\bB)) \subset \cL(\bB') \, . $$
\end{defn}

With this more general model of relations between colored operads with bud systems, we can regard
structures that simultaneously satisfy both filtering for phases and theta roles as 
$$ \varphi^{-1}(\cL(\bB_\Phi)) \cap \varphi'^{-1}(\cL(\bB_\Theta)) \cap \cL(\tilde\bB_{\Phi,\Theta})\, , $$
where $\cL(\tilde\bB_{\Phi,\Theta})$ is the language of the transduction that encodes compatibility
restrictions between phases and theta roles, namely further restrictions that are not directly captured
just by the intersection $\varphi^{-1}(\cL(\bB_\Phi)) \cap \varphi'^{-1}(\cL(\bB_\Theta))$.

\section{Theta and Phase Coloring and Movement} \label{ThetaPhaseMoveSec}

In both cases of Definition~\ref{opmorph} and Definition~\ref{optransduce},  the relation 
between the two colored operads does constrain the structures that are accepted by both, 
and the kind of movement that is achievable via the colored Merge. For example, not all the
cases of movement discussed in \cite{MarLar} that satisfies the coloring of $\cL(\bB_\Theta)$
would also satisfy the coloring of $\cL(\bB_\Phi)$, since some involved movement within
the interior of the phase, to positions other than those marked by $\fs^\downarrow$ in the phase-coloring,
for positions within the phase as in Remark~\ref{Mphasesrem}. 
 
In this section we analyze some cases  that illustrate more explicitly the constraints on
compatibility between the $\cL(\bB_\Phi)$ and $\cL(\bB_\Theta)$ filtering systems.

\subsection{Theta roles and movement}
As an example of interaction between the coloring rules for phases and for
theta roles, consider the case of passive. Let's take the example of the 
sentence ``{\em  The fox was chased}", for which the Merge derivation is
analyzed in \S 6.1.2 of \cite{ChomskyElements}. We follow the generators
of both $\cL(\bB_\Theta)$ and $\cL(\bB_\Phi)$, or equivalently both colored
Merge operations.

We can start with forming the object $\{ \text{ chase } , \text{ the fox } \}$.
We have, in $\cL(\bB_\Phi)$ a generator of $\cR_\Phi$
$$ \Tree[ .$\fz^\downarrow_{{\rm lex}, V}$ [ {$(\text{the}, \fh^{{\rm lex},D}, \fz^\uparrow_{{\rm lex},D})$} {$(\text{fox}, \fz^\downarrow_{{\rm lex},D})$}  ] ] $$
 that composes with a generator
 \begin{equation}\label{gen2fox}
 \Tree[ .{$(\fh^{{\rm lex},V}, \fs^\uparrow_{{\rm lex},V})$} [ {$(\text{chase}, \fh^{{\rm lex},V}, \fz^\uparrow_{{\rm lex},V}, \fs^\uparrow_{{\rm lex},V})$}  $\fz^\downarrow_{{\rm lex},V}$ ] ] 
\end{equation} 
This second generator is the interesting one to compare with the theta role assignments, as in 
$\cL(\bB_\Theta)$ it would correspond to a generator of $\cR_\Theta$ of the form
 \begin{equation}\label{gen2foxTheta}
 \Tree[ .{$\theta_E^\uparrow$} [ {$(\text{chase}, (\theta_E^\uparrow, \theta_I^\uparrow))$} $\theta_I^\downarrow$ ] ] 
\end{equation} 
The Merge derivation described in \S 6.1.2 of \cite{ChomskyElements} then moves the {\tt IA} ``{\em the fox}" by IM.
This produces, in the two coloring bud systems, respective generators
 \begin{equation}\label{gen3foxIM}
\Tree[ .$\fs^\downarrow_{{\rm lex},V}$ [ $\fz^\downarrow_{{\rm lex},V}$ {$(1,\fm)$} ] ] \ \ \ \ \text{ and } \ \ \ \
\Tree[ .$\theta_0$ [ $\theta_I^\downarrow$ {$(1,\theta_0)$} ] ] 
\end{equation}
The output of these is then composed, respectively, with one of the inputs of the generators
 \begin{equation}\label{gen3foxv*}
 \Tree[ .{$\fz^\downarrow_{v}$} [ $\fs^\downarrow_{{\rm lex},V}$ {$(\fh^{{\rm lex},V}, \fs^\uparrow_{{\rm lex},V})$} ] ]  \ \ \ \ \text{ and } \ \ \ \  \Tree[ .$\theta_E^\uparrow$  [ $\theta_0$ $\theta_E^\uparrow$ ] ] 
 \end{equation}
 where the other input is respectively matched to the outputs of \eqref{gen2fox} and \eqref{gen2foxTheta}.
 These are then matched, at the next step, with generators
  \begin{equation}\label{gen4foxv}
  \Tree[ .{$(\fh^{v}, \fs^\uparrow_{v})$} [ {$(\fh^{v}, \fz^\uparrow_{v}, \fs^\uparrow_{v})$} {$\fz^\downarrow_{v}$} ] ] \ \ \ \text{ and } \ \ \  \Tree[ .$\theta_E^\uparrow$  [ $\theta_0$ $\theta_E^\uparrow$ ] ] 
  \end{equation}
  Note that, with this formalism, at this point the $\theta_E$ role has not been discharged yet. Note also that, 
  while $\mathtt{v}^*$ is a phase-head (transitive case), $\mathtt{v}$ is a head
that is not a phase-head, so IM here does not have to move {\tt IA} to Spec positions for $\mathtt{v}$, as it can move directly to the edge of the phase that contains the $v$ head. In the
colored operad of phases, this generator is then followed by
   \begin{equation}\label{gen5foxv}
  \Tree[ .$\fz^\downarrow_I$  [ {$({\rm was}, \fs^\downarrow_v)$} {$(\fh^v , \fs^\uparrow_v)$} ] ] \,\, .
  \end{equation}
The next step is the merging in of an {\tt INFL} ($\mathtt{I}$) head.
 At the level of the colored operad of phases, we can associate to this a generator of the form
\begin{equation}\label{genIfox}
  \Tree[ .{$(\fh^I, \fs^\uparrow_I)$} [ {$(\fh^I, \fz^\uparrow_I, \fs^\uparrow_I)$} $\fz^\downarrow_I$ ] ]  
\end{equation}  
At the next step we can proceed by moving {\tt IA} at the spec of {\tt INFL} position within the phase, using generators 
 $$  \Tree[ .{$\fs^\downarrow_I$} [ {$\fs^\downarrow_{{\rm lex}, V}$} $(1,\fm)$ ] ] \ \ \ \ \text{ and } \ \ \  \Tree[ .$\theta_0$  [ $\theta_0$ $(1,\theta_0)$ ] ] \, . $$
 In the colored operad of phases this is finally capped with a generator of the form
 $$ \Tree[ .$\fc^\downarrow$ [ $\fs^\downarrow_I$ {$(\fh^I,\fs^\uparrow_I)$} ] ] \, .  $$

However, from the side of theta theory, we need to discharge the $\theta_E^\uparrow$, and the
IM movement of the {\tt IA} to Spec-of-{\tt INFL} position will only move to a $\theta_0$-marked position that
cannot be the receiver. Thus, the only possibility for matching the theta roles in this case is that
the $\theta_E^\downarrow$ is discharged into the $\fs^\downarrow_I$ position, by a generator
\begin{equation}\label{thetawas}
 \Tree[ .$\theta_0$ [ {$({\rm was}, \theta_E^\downarrow)$} $\theta_E^\uparrow$ ] ]  
\end{equation}
which is paired to the generator  \eqref{gen5foxv}.
In other words that the ``{\em was}"
of the ``{\em was chased}" acts as receiver of the $\theta_E$ assignment of ``{\em chase}" when
in the passive form. After the $\theta_E$ is discharged there are no more theta roles carried, so
the remaining generators \eqref{gen5foxv}, \eqref{genIfox} of the colored operad of phases are
paired to the generator
$$ \Tree[ .$\theta_0$  [ $\theta_0$ $(1,\theta_0)$ ] ]  $$
of the colored operad of theta roles. 

Note that adding an optional {\em ``by someone"} to the phrase would not change where the
$\theta_E$ can be discharged, as that would receive an $\fm$ marker
in $\cL(\bB_\Phi)$ and a $\theta_0$ marker in $\cL(\bB_\Theta)$, since it is not a mandatory part
of the structure. 

Note that this handling of the $\theta_E$ assignment also provides a reason, in terms of compatibility 
of coloring algorithms, for why $\mathtt{v}$ (unlike $\mathtt{v}^*$) should not be a phase head, otherwise its specifier position 
would be a receiver of IM movement, hence necessarily a $\theta_0$-colored position and would not 
possibly act as discharger for the $\theta_E^\uparrow$. The specifier position for {\tt INFL} is also not
Spec-of-phase, since the {\tt INFL} head is not a phase head, but it is one of the positions in the interior
of the phase that IM can move to, in which case they carry a $\theta_0$ position, as illustrated here. 
This proposal of how to handle the external theta role assignment in passive forms,
which is here forced by the coloring rules, was previously proposed in \cite{BaJoRo}. It differs from
the proposal of disentangling $\theta_E$ from the injected $\underline{\theta}^\uparrow$ and
assigning $\theta_E^\uparrow$ at the $\fh^{\mathtt{v}^*}$ position as in \cite{ChomskyGK}, \cite{ChomskyGE2}, 
\cite{Chomsky23}, \cite{ChomskyMiracleCreed}.  A discussion of an interpretation of $\theta_E$ as
associated to the $\mathtt{v}^*$ head in this colored operad is included in \cite{MarLar}.

\subsection{Exceptional case-marking}
As another example, to see how to accommodate exceptional case-marking (ECM) in our formulation,
consider, for example, the two sentences ``{\em John expected her to win}" and ``{\em John persuaded her to win}".
In the first case, one can argue that IM raises {\em her} to a position in the phase
interior of $\mathtt{v}^*$. In so doing, it moves it to a non-theta position, in ECM, $\theta_0$-labelled.
This is not at the edge of the phase, but it has a ``subject-of predication" interpretation.
On the other hand if it were inserted in that same position through an EM operation,
then it would have to receive a theta role assignment from ``expect" but it does not: in
the ECM construction the raised pronoun, although it is marked as object case morphology,
it semantically received its theta role from the lower verb ``win" not from ``expect". 
In the second sentence, instead, it is indeed the recipient of a theta role assignment 
from ``persuade", hence it is obtained via an EM operation. Both cases are captured 
by the different local coloring move involved, namely two different generators of the 
colored operad of theta roles. Similar examples are discussed
in the context of raising vs.~control in \cite{ChomskyGK} and
in \cite{ChomskyGE}.

One can see in examples like this the combined rules of movement for 
the $\Phi$ and the $\Theta$ colored Merge operations, with the rule for
IM movement and phases marking the position with an $\fz_{\mathtt{v}^*}^\downarrow$
and the IM realized by a generator
$$ \Tree[ .$\fs^\downarrow_{\omega'}$ [ $\fs^\downarrow_{\omega}$ $(1,\fm)$ ] ] $$
that moves to a $\fs^\downarrow_{\omega}$ position located in the interior of the $\mathtt{v}^*$
phase, hence below a $\fz_{\mathtt{v}^*}^\downarrow$ colored operad insertion. 
The fact that, in the Merge colored for phases, this movement is to a position inside 
the $\fs_{\mathtt{v}^*}^\downarrow$-marked interior of the
$\mathtt{v}^*$ phase is consistent with the object case morphology. On the other hand, at the level 
of the Merge colored for theta role assignments, this movement by IM is implemented 
by a generator of the form
$$ \Tree[ .$\theta_0$ [ $\theta_E^\downarrow$ $(1,\theta_0)$ ] ] $$
which marks {\em her} with a $\theta_0$ position, while the $\theta_E^\downarrow$
marker remains at the trace of movement, hence as a theta role assigned by {\em win},
not by {\em expect}. 

\subsection{Double object constructions and theta hierarchies} \label{2ObjSec}

We now analyze again some double object constructions discussed in \cite{MarLar},
where the coloring by $\bB_\Theta$ for theta role assignment would support a 
IM movement interpretation of certain configurations of theta roles. In the analysis
given in \cite{MarLar}, we only considered the compatibility of movement with the
bud generating system of the colored operad of theta theory. However, as we
have been discussing in this paper, movement is also filtered by the colored
operad of phases. Thus, we can analyze again the possible syntactic objects that
account for double object constructions, and check whether movement is needed
and whether it is compatible with the bud generating systems of both operads,
the colored operad of phases $\bB_\Phi$ and the colored operad of theta roles
$\bB_\Theta$. This will lead to favoring certain possibilities. It will also serve as
another good test case for how to formulate the simultaneous effect of both filters.

\subsubsection{Double object and coloring by phases}\label{2ObjPhiSec}
Consider a double object sentence like ``{\em Mary gave a book to John}". From the
point of view of phases, we have a structure of the form 
$$ \{ \mathtt{v}^* \{ \mathtt{NP}, \{ {\rm Root}, \mathtt{XP} \} \} \} \, . $$
In general, such structures can occur in two different ways: 
$\mathtt{NP}$ can be put into its position by EM (in the case of Control) or by IM (in the case of ECM discussed
above). These possibilities clearly differ when it comes to the structure of theta roles. This
fact shows that we do not expect the implementation of both filters, for phases and theta roles,
to be realized as a matching of conditions via a morphism of colored operads in the sense of 
Definition~\ref{opmorph}, but it requires the more general form of correspondences as in
Definition~\ref{optransduce}. We will return to discuss this in \S \ref{ComboPhiThetaSec}.

\smallskip

More specifically in the example of ``{\em Mary gave a book to John}", we can
discuss all the possible candidates for syntactic objects representing this double
object construction, with their full Extended Projection, and check what their coloring
looks like, according to the bud generating systems for phases and for theta roles. 
We will see that the interaction between the coloring by phases and by theta roles
selects for some of them.

\smallskip

\begin{enumerate}
\item Consider first a candidate form of a syntactic object of the form 
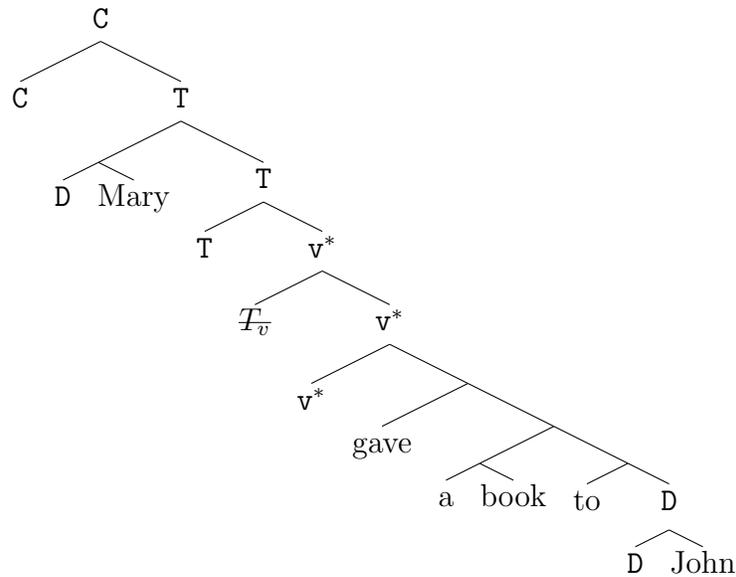
\begin{equation}\label{2ObjForm1}
 \Tree[ .$\mathtt{C}$  $\mathtt{C}$ [ .$\mathtt{T}$ [ $\mathtt{D}$ Mary ] [ .$\mathtt{T}$ $\mathtt{T}$ [ .$\mathtt{v}^*$ \text{\sout{$T_v$}} [ .$\mathtt{v}^*$ $\mathtt{v}^*$  [ gave [ [ a book ]  [ to [ .$\mathtt{D}$ $\mathtt{D}$ John ] ]]]]]]] ]
\end{equation}
with the IM movement of the term 
$$ T_v = \Tree[ .$\mathtt{D}$ $\mathtt{D}$ Mary ] \, . $$
\end{enumerate}

In terms of coloring according to $\bB_\Phi$ for the structure of phases, the part
$$ \Tree[ gave [ [ a book ]  [ to [ .$\mathtt{D}$ $\mathtt{D}$ John ] ] ] ] $$
can be colored by operad compositions of generators 
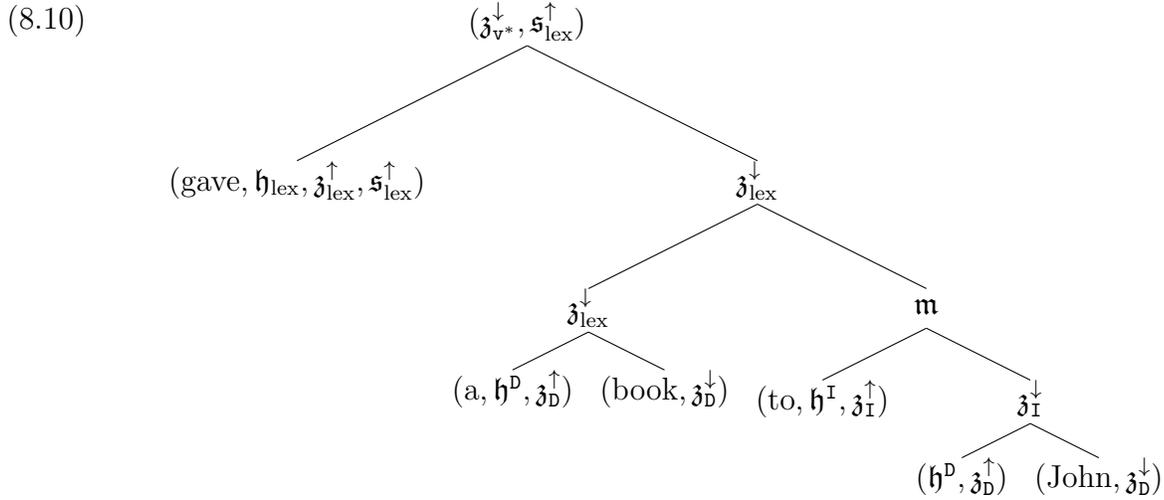
\begin{equation}\label{2Ob1Phi}
  \Tree[ .{$(\fz_{\mathtt{v}^*}^\downarrow, \fs_{{\rm lex}}^\uparrow)$}    {$(\text{gave},\fh_{{\rm lex}}, \fz_{{\rm lex}}^\uparrow, \fs_{{\rm lex}}^\uparrow)$} 
[ .$\fz_{{\rm lex}}^\downarrow$ [ .$\fz_{{\rm lex}}^\downarrow$ {$(\text{a},\fh^{\mathtt{D}},\fz_{\mathtt{D}}^\uparrow)$} {$(\text{book},\fz_{\mathtt{D}}^\downarrow)$} ] [ .$\fm$  {$(\text{to}, \fh^{\mathtt{I}}, \fz^\uparrow_{\mathtt{I}})$} [.$\fz^\downarrow_{\mathtt{I}}$ {$(\fh^{\mathtt{D}}, \fz_{\mathtt{D}}^\uparrow)$}  
{$(\text{John}, \fz_{\mathtt{D}}^\downarrow)$} ] ] ] ] 
\end{equation}
Then, in the rest of the Extended Projection we have further generators 
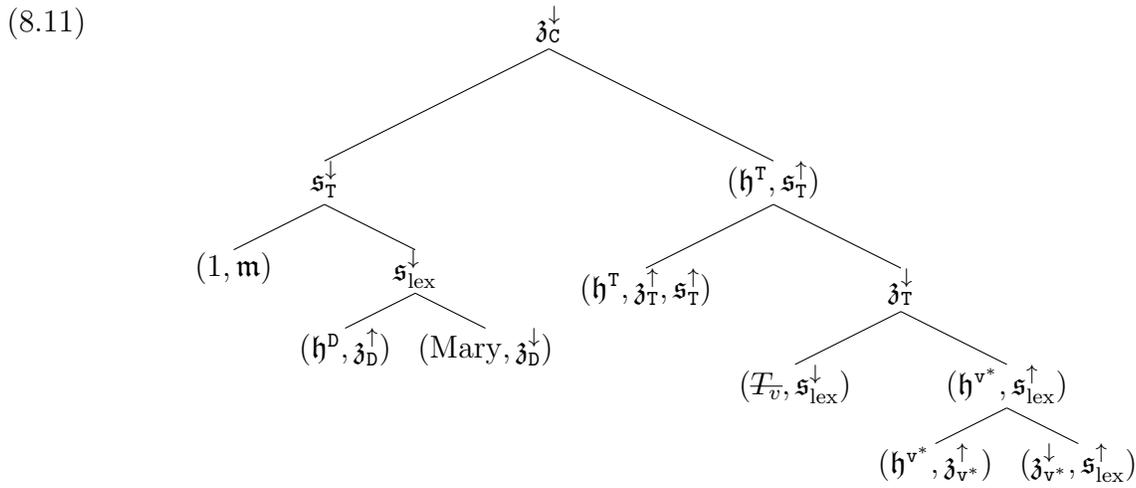
\begin{equation}\label{2Ob2Phi}
 \Tree[ .$\fz_{\mathtt{C}}^\downarrow$ [ .$\fs_{\mathtt{T}}^\downarrow$ $(1,\fm)$ [ .$\fs_{{\rm lex}}^\downarrow$ 
{$(\fh^{\mathtt{D}}, \fz_{\mathtt{D}}^\uparrow)$} {$(\text{Mary}, \fz_{\mathtt{D}}^\downarrow)$} ] ] [ .{$(\fh^{\mathtt{T}}, \fs_{\mathtt{T}}^\uparrow)$} {$(\fh^{\mathtt{T}}, \fz_{\mathtt{T}}^\uparrow, \fs_{\mathtt{T}}^\uparrow)$} [ .$\fz_{\mathtt{T}}^\downarrow$ {$(\text{\sout{$T_v$}},\fs_{\rm lex}^\downarrow)$} [ .{$(\fh^{\mathtt{v}^*},\fs_{\rm lex}^\uparrow)$} {$(\fh^{\mathtt{v}^*},\fz_{\mathtt{v}^*}^\uparrow)$} 
{$(\fz_{\mathtt{v}^*}^\downarrow, \fs_{\rm lex}^\uparrow)$} ] ] ] ] 
\end{equation}
with the last leaf input of \eqref{2Ob2Phi} matched with the root output of \eqref{2Ob1Phi}. 

\begin{enumerate}
  \setcounter{enumi}{1}
\item We then consider another possible candidate syntactic object of the form
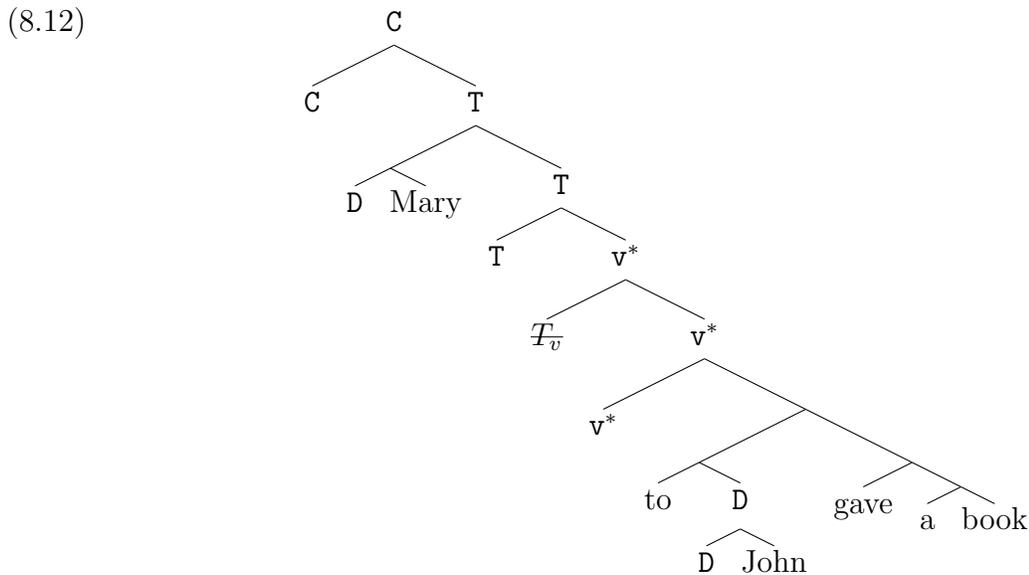
\begin{equation}\label{2ObjForm2}
 \Tree[ .$\mathtt{C}$  $\mathtt{C}$ [ .$\mathtt{T}$ [ $\mathtt{D}$ Mary ] [ .$\mathtt{T}$ $\mathtt{T}$ [ .$\mathtt{v}^*$ \text{\sout{$T_v$}} [ .$\mathtt{v}^*$ $\mathtt{v}^*$   [ [ to [ .$\mathtt{D}$ $\mathtt{D}$ John ] ] [ gave  [ a book ]  ] ] ] ] ] ] ]
\end{equation}
\end{enumerate}

In this case the upper part of the coloring in \eqref{2Ob2Phi} is unchanged but in the lower part, instead of \eqref{2Ob1Phi}
we now have
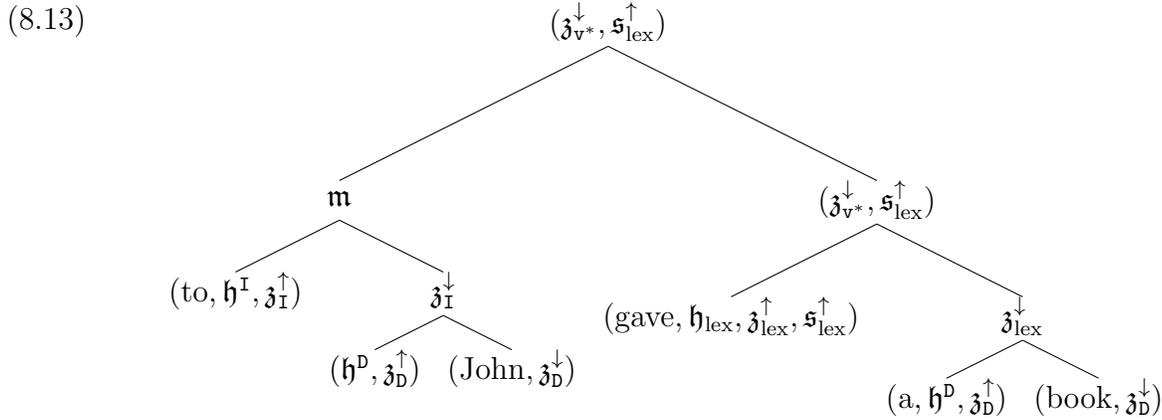
\begin{equation}\label{2Ob3Phi}
  \Tree[ .{$(\fz_{\mathtt{v}^*}^\downarrow, \fs_{{\rm lex}}^\uparrow)$} 
   [ .$\fm$  {$(\text{to}, \fh^{\mathtt{I}}, \fz^\uparrow_{\mathtt{I}})$} [.$\fz^\downarrow_{\mathtt{I}}$ {$(\fh^{\mathtt{D}}, \fz_{\mathtt{D}}^\uparrow)$}  {$(\text{John}, \fz_{\mathtt{D}}^\downarrow)$} ]  ] 
    [ .{$(\fz_{\mathtt{v}^*}^\downarrow, \fs_{{\rm lex}}^\uparrow)$}  
 {$(\text{gave},\fh_{{\rm lex}}, \fz_{{\rm lex}}^\uparrow, \fs_{{\rm lex}}^\uparrow)$}  [ .$\fz_{{\rm lex}}^\downarrow$ {$(\text{a},\fh^{\mathtt{D}},\fz_{\mathtt{D}}^\uparrow)$} {$(\text{book},\fz_{\mathtt{D}}^\downarrow)$} ] ] ] 
\end{equation}

Since all the trees are non-planar, both structures \eqref{2ObjForm1} and \eqref{2ObjForm2}
can be planarized in Externalization to produce the same string ``{\em Mary gave a book to John}".
Note that \eqref{2ObjForm1} can also be planarized in the form ``{\em Mary gave (to) John a book}"
as well as two verb-last forms that would be filtered out as Externalizations in English. It does
not, however, planarize as ``{\em Mary to John gave a book}" (as in the Mandarin ``{\em T\={a} g\u{e}ile t\={a} y\={\i} b\u{e}n sh\={u}}").
On the other hand, \eqref{2ObjForm2} can be planarized as ``{\em Mary to John gave a book}" as in Mandarin
but it cannot be planarized in the form ``{\em Mary gave (to) John a book}".

\begin{enumerate}
  \setcounter{enumi}{2}
\item There is a further possibility that is compatible with coloring with respect to the generators in $\cR_\Phi$, namely
a syntactic object of the form 
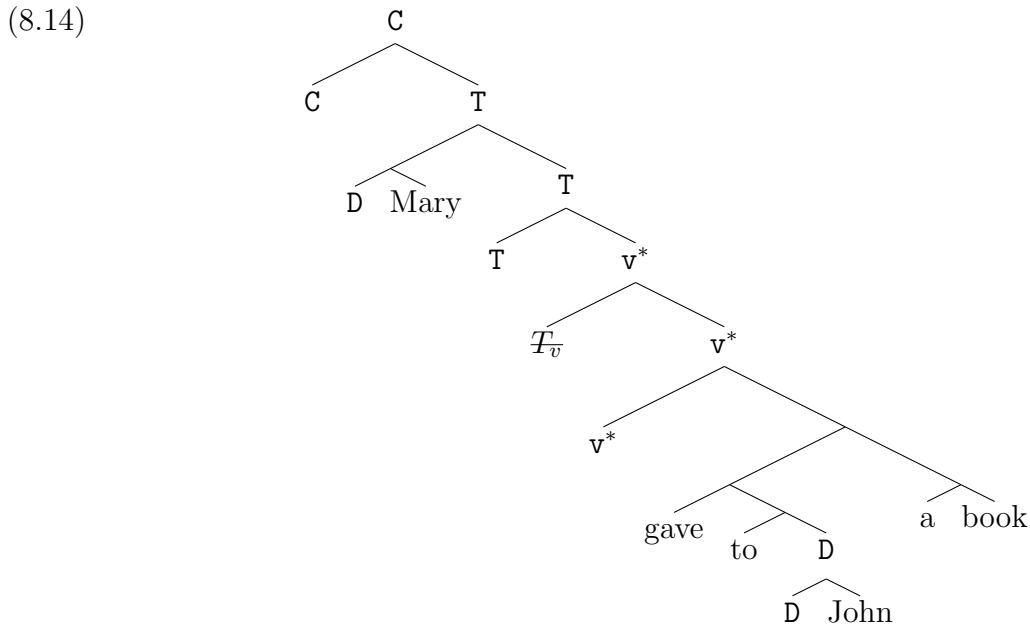
\begin{equation}\label{2ObjForm3}
 \Tree[ .$\mathtt{C}$  $\mathtt{C}$ [ .$\mathtt{T}$ [ $\mathtt{D}$ Mary ] [ .$\mathtt{T}$ $\mathtt{T}$ [ .$\mathtt{v}^*$ \text{\sout{$T_v$}} [ .$\mathtt{v}^*$ $\mathtt{v}^*$ [  [ gave [ to [ .$\mathtt{D}$ $\mathtt{D}$ John ]  ] ]  [ a book ]  ] ]  ] ] ] ]
\end{equation}
\end{enumerate}

Here again the coloring of \eqref{2Ob2Phi} is unchanged  but instead of \eqref{2Ob1Phi} or \eqref{2Ob3Phi} we have
\begin{equation}\label{2Ob4Phi}
  \Tree[ .{$(\fz_{\mathtt{v}^*}^\downarrow, \fs_{{\rm lex}}^\uparrow)$}  [ .{$(\fh_{{\rm lex}}, \fz_{{\rm lex}}^\uparrow, \fs_{{\rm lex}}^\uparrow)$} 
 {$(\text{gave},\fh_{{\rm lex}}, \fz_{{\rm lex}}^\uparrow, \fs_{{\rm lex}}^\uparrow)$} 
 [ .$\fm$  {$(\text{to}, \fh^{\mathtt{I}}, \fz^\uparrow_{\mathtt{I}})$} [.$\fz^\downarrow_{\mathtt{I}}$ {$(\fh^{\mathtt{D}}, \fz_{\mathtt{D}}^\uparrow)$}  {$(\text{John}, \fz_{\mathtt{D}}^\downarrow)$} ]  ]  ]
 [ .$\fz_{{\rm lex}}^\downarrow$ {$(\text{a},\fh^{\mathtt{D}},\fz_{\mathtt{D}}^\uparrow)$} {$(\text{book},\fz_{\mathtt{D}}^\downarrow)$} ] ] ]  \, .
\end{equation}
This can be planarized as `{\em Mary gave (to) John a book}" or as ``{\em Mary to John gave a book}" as in Mandarin,
but not as ``{\em Mary gave a book to John}". (Note that other planarizations are also possible: for example ``{\em Mary a book gave to John}", which will be filtered out in Externalization in both English and Mandarin. We focus here on these three
planarizations, as these provide enough cases realized in actual languages, to allow us to distinguish between the
different candidate syntactic objects.)

Note that the coloring by $\bB_\Phi$ does not rule out any of these three tree structures 
\eqref{2ObjForm1}, \eqref{2ObjForm2}, and \eqref{2ObjForm3}, as all of them can be colored according 
to the rules of $\cR_\Phi$ and give objects in $\cL(\bB_\Phi)$, so this coloring by phases 
alone does not resolve the ambiguity. We will see in \S \ref{2ObjThetaSec}  what happens
when one also considers the coloring by theta roles, with the renerators in $\cR_\Theta$.

\subsubsection{Movement in double object constructions} \label{2ObjMoveSec}
In  the cases  \eqref{2ObjForm2}, and \eqref{2ObjForm3} 
consistently colored as objects in $\cL(\bB_\phi)$, the remaining
planarization  ``{\em Mary gave (to) John a book}" for 
 \eqref{2ObjForm2}, and ``{\em Mary gave a book to John}" for \eqref{2ObjForm3} 
 could in principle be achieved via movement
 by IM of {\em gave} to spec-of-$\mathtt{v}^*$ position, respectively resulting in
 $\{ \text{gave-}v*, \{\{ \text{\sout{gave}}, \text{a-book} \}, \text{to-John} \}\}$ and 
 $\{ \text{gave-}v*, \{\{ \text{\sout{gave}}, \text{to-John} \}, \text{a-book}  \}\}$.
 This movement description was used in \cite{MarLar} to formulate the theta
 roles coloring for the double object constructions. 
 However, when we consider the full Extended Projection and the coloring by phases, 
 for this example we see that the spec-of-$\mathtt{v}^*$ position is used
 in the movement of the $T_v$ accessible term. Moreover, moving a term labelled
 by $(\fh^\omega, \fz_\omega^\uparrow, \fs_\omega^\uparrow)$ would require introducing additional
 generators for $\cR_\Phi$, in addition to the generators of the form \eqref{gens1m}
$$ T^\fs_{\fc,(1,\fm)}:= \Tree[ .$\fs^\downarrow_{\omega'}$ [ $\fc^\downarrow_\omega$ $(1,\fm)$ ] ] $$
where we assumed that  $\fc$ can only be either a $\fs$ or a $\fz$. However, allowing additional generators of the form
\begin{equation}\label{addgensIMh}
 \Tree[ .$\fs^\downarrow_{\omega'}$ [ {$(\fh^\omega, \fz_\omega^\uparrow, \fs_\omega^\uparrow)$} $(1,\fm)$ ] ] \ \ \ \text{ or } \ \ \ 
\Tree[ .$\fs^\downarrow_{\omega'}$ [ {$(\fh^\omega,\fz_\omega^\uparrow)$} $(1,\fm)$ ] ] \ \ \ \text{ or } \ \ \ 
\Tree[ .$\fs^\downarrow_{\omega'}$ [ {$(\fh^{\omega'},\fs_\omega^\uparrow)$} $(1,\fm)$ ] ]
\end{equation}
has some drawbacks. In the first case, for example, if the trace of the movement at the node
\begin{equation}\label{tracenode}
 \Tree[ .{$(\fs_\omega^\uparrow, \fz_{\omega'}^\downarrow)$} [ {$\text{\sout{ $(\fh^\omega, \fz_\omega^\uparrow, \fs_\omega^\uparrow)$ } }$} $\fz_{\omega}^\downarrow$ ] ] 
\end{equation} 
remains colored by
$(\fh^\omega, \fz_\omega^\uparrow, \fs_\omega^\uparrow)$, which contains a head label $\fh^\omega$ and 
hence participates in the labeling algorithm, this creates a problem 
with the ECP, as formulated in Remark~\ref{ECPrem}, since the trace should not participate in labeling. 
 In fact, a closer inspection shows that it is possible to include additional generators of the form \eqref{addgensIMh}
 without incurring in a violation of ECP. In fact, for IM movement realized via one of the generator \eqref{addgensIMh}
 to take place, the syntactic object has to include a position with an input marked by $\fs_{\omega'}^\downarrow$
 to which the output of a generator as in \eqref{addgensIMh} is grafted. This means that both the maximal projection of
 the head $\fh^\omega$ involved in the generator \eqref{addgensIMh} and that of the higher head
 $\fh^{\omega'}$, whose complement $\fz_{\omega'}^\downarrow$ appears in the output of \eqref{tracenode},
 have to be already present in the syntactic object, for a generator of the form \eqref{addgensIMh} to
 be included in the structure. But the presence of these labelled positions implies that all the internal
 vertices that need to receive a labeling $\fh^\omega$ are already labelled in this syntactic object, hence the trace 
 $\text{\sout{ $(\fh^\omega, \fz_\omega^\uparrow, \fs_\omega^\uparrow)$ } }$ does not participate 
 in any labeling role. In terms of the equivalent colored Merge formulation, this just means that
 when the colored Merge operation implementing this IM operation takes place, all the labeling in which
 the head $\fh^\omega$ participates has already been completed, hence the trace has not labeling role,
 compatibly with ECP.

\subsection{Double object and coloring by theta roles} \label{2ObjThetaSec} 

We now compare the proposed forms \eqref{2ObjForm1}, \eqref{2ObjForm2} and \eqref{2ObjForm3} 
of the syntactic object in terms of coloring by theta roles in $\cL(\bB_\Theta)$. 
As we will see, this coloring in $\cL(\bB_\Phi)$ put a strong constraint on the
possible coloring in $\cL(\bB_\Theta)$, including assumptions about 
theta-hierarchies in the generators in $\cR_\Theta$.

In all cases  \eqref{2ObjForm1}, \eqref{2ObjForm2}, and \eqref{2ObjForm3}, in the
upper part of the structure all nodes from the root until the second ${\mathtt{T}}$-labelled node 
correspond to generators
$$ \Tree[ .$\theta_0$ [ $\theta_0$ $\theta_0$ ] ] $$
while the first $\mathtt{v}^*$ node is a generator
$$ \Tree[ .$\theta_0$ [ {$(\text{\sout{$T_v$}},\theta_E^\downarrow)$} $\theta_E^\uparrow$ ] ] $$
followed by a generator of the form
 $$ \Tree[ .$\theta_E^\uparrow$ [ $\theta_0$ $\theta_E^\uparrow$ ] ] \, . $$
In the lower part of the structure, however, the three cases look different. 

\begin{prop}
Requiring consistent coloring in both $\cR_\Phi$ and $\cR_\Theta$ of the 
candidate syntactic objects \eqref{2ObjForm1}, \eqref{2ObjForm2}, \eqref{2ObjForm3}
and of syntactic objects obtained from \eqref{2ObjForm2} and \eqref{2ObjForm3}
via IM movement, shows that the possibility favored by the coloring rules is that
both  \eqref{2ObjForm2} and \eqref{2ObjForm3} are available prior to Externalization
(hence both hierarchies $\theta_{\mathtt{GO}}\prec \theta_{\mathtt{TH}}$ and
$\theta_{\mathtt{GO}}\succ \theta_{\mathtt{TH}}$ are possible). The other
possibilities 
\begin{itemize}
\item the syntactic object \eqref{2ObjForm1},
\item only one of \eqref{2ObjForm2} or \eqref{2ObjForm3}, together with a syntactic
object obtained from it via IM movement that accounts for remaining planarizations
that are linguistically attested, 
\end{itemize}
are all disfavored by the compatibility of the coloring rules of $\cR_\Phi$ and $\cR_\Theta$. 
\end{prop}

\proof
It is convenient to describe first the case of \eqref{2ObjForm2} and \eqref{2ObjForm3}. 
In the case of \eqref{2ObjForm2} we can give the lower part of the structure a theta coloring
$$ \Tree[ .$\theta_E^\uparrow$  [ .$\theta_{I,2}^\downarrow$  {$(\text{to},\theta_0)$} [ .$\theta_{I,2}^\downarrow$  {$(\mathtt{D},\theta_0)$} {$(\text{John},\theta_{I,2}^\downarrow)$} ] ] [ .{$(\theta_E^\uparrow,\theta_{I,2}^\uparrow)$} [ .$\theta_{I,1}^\downarrow$
 {$(\text{a},\theta_0)$}  {$(\text{book},\theta_{I,1}^\downarrow)$}  ]   {$(\text{gave}, (\theta_E^\uparrow, \theta_{I,1}^\uparrow, \theta_{I,2}^\uparrow))$}  ] ]  $$
In this case the goal/theme theta-hierarchy of the internal theta roles is 
$(\theta_{I,1},\theta_{I,2})=(\theta_{\mathtt{TH}}, \theta_{\mathtt{GO}} )$.
In the case of \eqref{2ObjForm3}, similarly, we can color the structure according to the
coloring rules of $\cR_\Theta$ in the form
$$ 
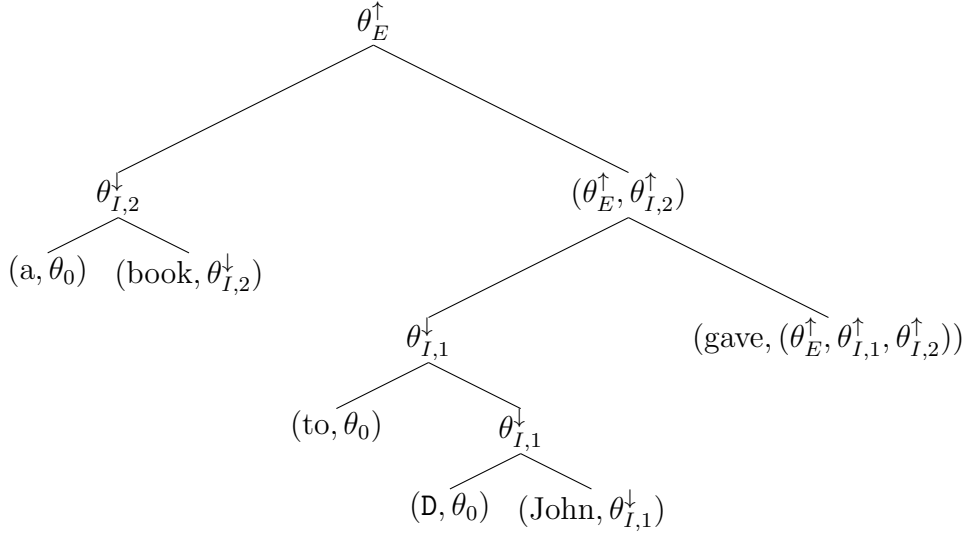
\Tree[ .$\theta_E^\uparrow$ [ .$\theta_{I,2}^\downarrow$ {$(\text{a},\theta_0)$} {$(\text{book},\theta_{I,2}^\downarrow)$} ]  [ .{$(\theta_E^\uparrow, \theta_{I,2}^\uparrow)$} [ .$\theta_{I,1}^\downarrow$ {$(\text{to},\theta_0)$} [ .$\theta_{I,1}^\downarrow$ {$(\mathtt{D},\theta_0)$}  {$(\text{John}, \theta_{I,1}^\downarrow)$} ] ] {$(\text{gave}, (\theta_E^\uparrow, \theta_{I,1}^\uparrow, \theta_{I,2}^\uparrow))$}   ] ]  
$$ 
This is also well colored by generators in $\cR_\Theta$, this time with a goal/theme theta-hierarchy of
the opposite form $(\theta_{I,1},\theta_{I,2})=(\theta_{\mathtt{GO}}, \theta_{\mathtt{TH}} )$.

If we allow for the possibility that both goal/theme theta-hierarchies 
$(\theta_{I,1},\theta_{I,2})=(\theta_{\mathtt{GO}}, \theta_{\mathtt{TH}} )$ and 
$(\theta_{I,1},\theta_{I,2})=(\theta_{\mathtt{TH}}, \theta_{\mathtt{GO}} )$ may be
possible and that either one or the other is filtered out in Externalization (depending
on specific languages), these two possibilities then suffice for all the possible
linear orderings, namely all six permutations of the predicate and its two objects
(the two internal theta roles). 

In a stronger formulation of theta-hierarchies, in which only one {\em or} the other
hierarchy 
$(\theta_{I,1},\theta_{I,2})=(\theta_{\mathtt{GO}}, \theta_{\mathtt{TH}} )$ or 
$(\theta_{I,1},\theta_{I,2})=(\theta_{\mathtt{TH}}, \theta_{\mathtt{GO}} )$
is possible, prior to Externalization, then only one of 
\eqref{2ObjForm2} and \eqref{2ObjForm3} will be accepted by the theta
roles coloring. This means that one of the possible linear ordering known
to occur will not be obtainable as planar embedding of the accepted 
theta-colored syntactic object. This would then imply that the remaining
linear ordering has to be an effect of movement, which, in this case, would
mean movement from a position marked with $(\fh^{\rm lex}, \fz_{\rm lex}^\uparrow, \fs_{\rm lex}^\uparrow)$,
hence this would require the inclusion of additional generators for the form \eqref{addgensIMh},
as discussed above.

For example, consider again the case of \eqref{2Ob4Phi}, with phase coloring
 $$ 
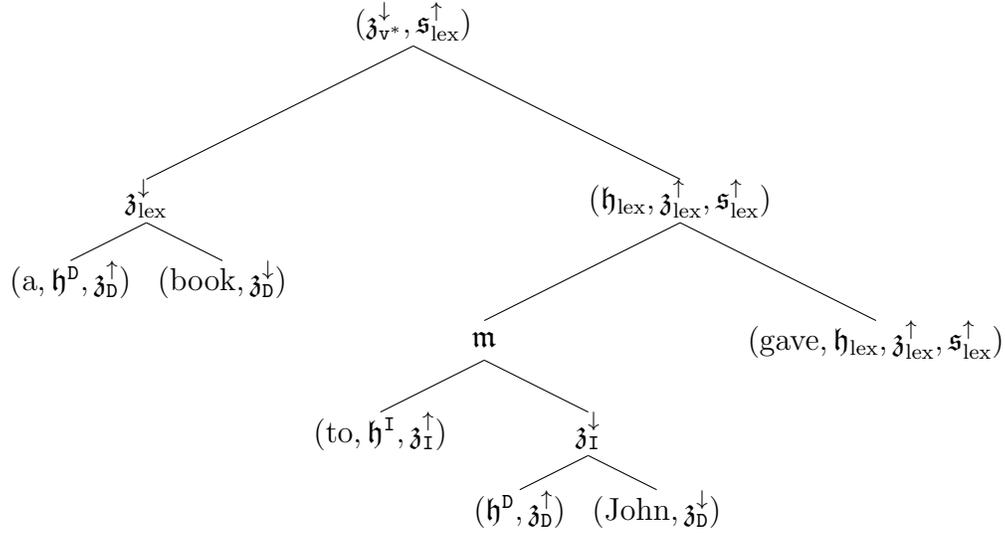
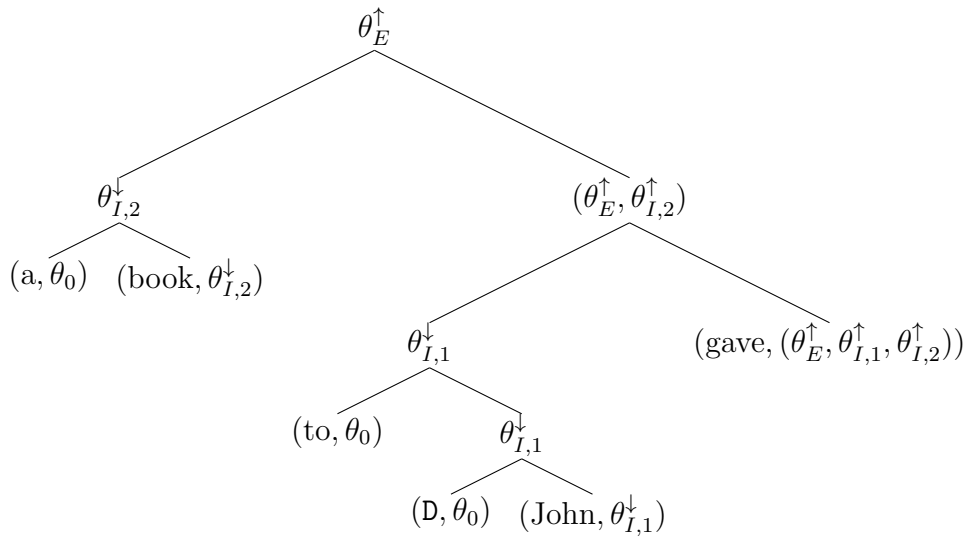
\Tree[ .{$(\fz_{\mathtt{v}^*}^\downarrow, \fs_{{\rm lex}}^\uparrow)$} 
[ .$\fz_{{\rm lex}}^\downarrow$ {$(\text{a},\fh^{\mathtt{D}},\fz_{\mathtt{D}}^\uparrow)$} {$(\text{book},\fz_{\mathtt{D}}^\downarrow)$} ]  [ .{$(\fh_{{\rm lex}}, \fz_{{\rm lex}}^\uparrow, \fs_{{\rm lex}}^\uparrow)$} 
  [ .$\fm$  {$(\text{to}, \fh^{\mathtt{I}}, \fz^\uparrow_{\mathtt{I}})$} [.$\fz^\downarrow_{\mathtt{I}}$ {$(\fh^{\mathtt{D}}, \fz_{\mathtt{D}}^\uparrow)$}  {$(\text{John}, \fz_{\mathtt{D}}^\downarrow)$} ]  ]  
 {$(\text{gave},\fh_{{\rm lex}}, \fz_{{\rm lex}}^\uparrow, \fs_{{\rm lex}}^\uparrow)$} 
 ] ]  
$$ 
and theta coloring
$$ 
\Tree[ .$\theta_E^\uparrow$ [ .$\theta_{I,2}^\downarrow$ {$(\text{a},\theta_0)$} {$(\text{book},\theta_{I,2}^\downarrow)$} ]  [ .{$(\theta_E^\uparrow, \theta_{I,2}^\uparrow)$} [ .$\theta_{I,1}^\downarrow$ {$(\text{to},\theta_0)$} [ .$\theta_{I,1}^\downarrow$ {$(\mathtt{D},\theta_0)$}  {$(\text{John}, \theta_{I,1}^\downarrow)$} ] ] {$(\text{gave}, (\theta_E^\uparrow, \theta_{I,1}^\uparrow, \theta_{I,2}^\uparrow))$}   ] ]  \, , 
$$ 
which can be planarized as ``{\em to John gave a book}" (as in Mandarin) or as `{\em gave (to) John a book}",
but not as ``{\em gave a book to John}". To obtain the linear ordering ``{\em gave a book to John}" requires movement.
In terms of coloring by theta roles, this is compatible with the generators of $\cR_\Theta$, as in \cite{MarLar}, and
would give a syntactic object of the form
$$ 
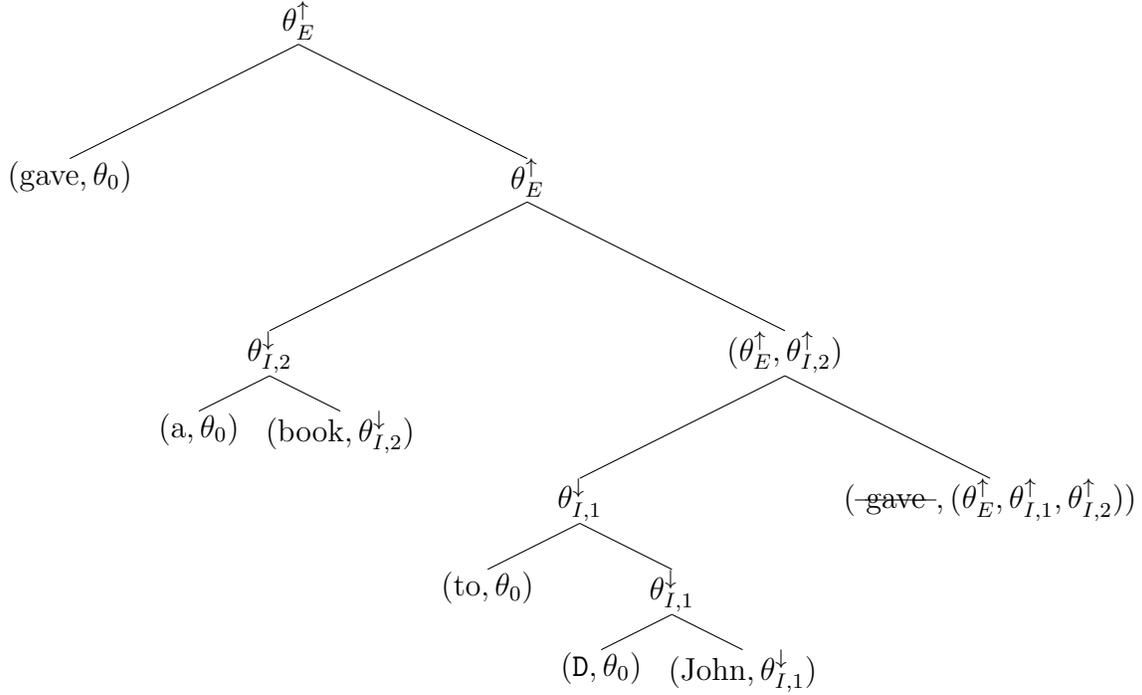
\Tree[ .$\theta_E^\uparrow$  $(\text{gave},\theta_0)$  [ .$\theta_E^\uparrow$ [ .$\theta_{I,2}^\downarrow$ {$(\text{a},\theta_0)$} {$(\text{book},\theta_{I,2}^\downarrow)$} ]  [ .{$(\theta_E^\uparrow, \theta_{I,2}^\uparrow)$} [ .$\theta_{I,1}^\downarrow$ {$(\text{to},\theta_0)$} [ .$\theta_{I,1}^\downarrow$ {$(\mathtt{D},\theta_0)$}  {$(\text{John}, \theta_{I,1}^\downarrow)$} ] ] {$(\text{\sout{ gave }}, (\theta_E^\uparrow, \theta_{I,1}^\uparrow, \theta_{I,2}^\uparrow))$}   ] ]  ]  \, .
$$ 
On the other hand, for coloring according to phases, this movement requires
adding to $\cR_\Phi$ a generator of the form 
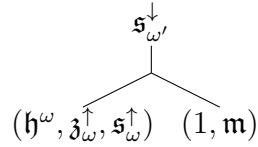
\begin{equation}\label{gentoadd}
 \Tree[ .$\fs^\downarrow_{\omega'}$ [ {$(\fh^\omega, \fz_\omega^\uparrow, \fs_\omega^\uparrow)$} $(1,\fm)$ ] ] \, . 
\end{equation} 
This means that the term $(\text{gave},\fh_{{\rm lex}}, \fz_{{\rm lex}}^\uparrow, \fs_{{\rm lex}}^\uparrow)$ is moved by IM
to a specifier position $\fs^\downarrow_{\omega'}$, which also needs to be labelled $\theta_0$ in the theta roles coloring. 
Because the root output of \eqref{2Ob4Phi} is colored 
$(\fz_{\mathtt{v}^*}^\downarrow, \fs_{{\rm lex}}^\uparrow)$, the next generator above can only be
of the form 
$$ \Tree[ .{$(\fh_{\mathtt{v}^*}, \fs_{\rm lex}^\uparrow)$} [ {$(\fh^{\mathtt{v}^*}, \fz_{\mathtt{v}^*}^\uparrow)$} 
 {$(\fz_{\mathtt{v}^*}^\downarrow, \fs_{\rm lex}^\uparrow)$} ] ]  \, . $$
 
However, this creates a compatibility problem because the output coloring $\fs^\downarrow_{\omega'}$ cannot be
equal to $\fs_{\rm lex}^\downarrow$, since as shown in \eqref{2Ob2Phi} 
that input already receives the root output of the accessible term
$$ T_v = \Tree[ .$\mathtt{D}$ $\mathtt{D}$ Mary ]  $$
as shown in \eqref{2ObjForm3}, which carried the external theta role $\theta_E^\downarrow$.
The $\fs^\downarrow_{\omega'}$ also cannot be $\fs^\downarrow_{\mathtt{T}}$ as that
receives the output of the generator
$$ \Tree[ .{$\fs^\downarrow_{\mathtt{T}}$} [ $\fs_{\rm lex}^\downarrow$ $(1,\fm)$ ] ] $$
in the movement of the accessible term $T_v$, as shown in \eqref{2Ob2Phi}. 
Also $\omega'$ cannot be a higher head above $\fs^\downarrow_{\mathtt{T}}$, 
since that would not result in a planarization of the form ``{\em Mary gave a book to John}" 
(though it could planarize to ``{\em gave Mary a book to John}").
Moreover, movement to the specifier position $\fs^\downarrow_{\omega'}$ of a higher hear 
$\omega'$  may also violate the Phase Impenetrability Condition if the head $\fh^{\omega'}$ 
belongs to the interior of a higher phase. Thus, the position $\fs^\downarrow_{\omega'}$ 
would have to be located between the maximal projection of the head 
$\fh^{\mathtt{v}^*}$ and the head $\fh^{\mathtt{T}}$. This would have the effect of replacing
\eqref{2Ob2Phi} with a structure of the form
{\small
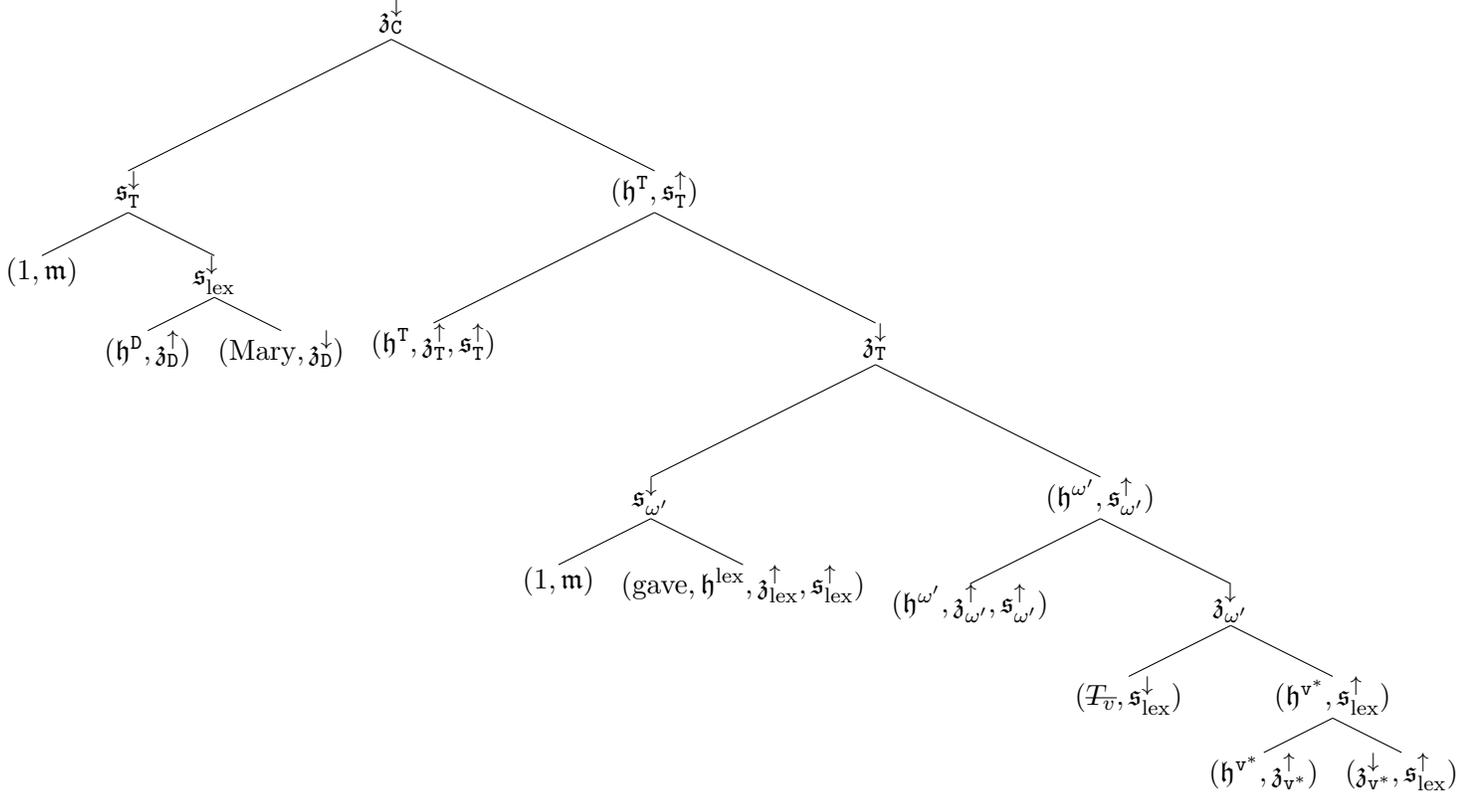
\begin{equation}\label{new2Ob2Phi}
 \Tree[ .$\fz_{\mathtt{C}}^\downarrow$ [ .$\fs_{\mathtt{T}}^\downarrow$ $(1,\fm)$ [ .$\fs_{{\rm lex}}^\downarrow$ 
{$(\fh^{\mathtt{D}}, \fz_{\mathtt{D}}^\uparrow)$} {$(\text{Mary}, \fz_{\mathtt{D}}^\downarrow)$} ] ] [ .{$(\fh^{\mathtt{T}}, \fs_{\mathtt{T}}^\uparrow)$} {$(\fh^{\mathtt{T}}, \fz_{\mathtt{T}}^\uparrow, \fs_{\mathtt{T}}^\uparrow)$} 
[ .$\fz_{\mathtt{T}}^\downarrow$ 
[ .{$\fs_{\omega'}^\downarrow$} {$(1,\fm)$}  {$(\text{gave},\fh^{{\rm lex}}, \fz_{{\rm lex}}^\uparrow, \fs_{{\rm lex}}^\uparrow)$}  ]
[ .{$(\fh^{\omega'}, \fs_{\omega'}^\uparrow)$} 
{$(\fh^{\omega'}, \fz_{\omega'}^\uparrow, \fs_{\omega'}^\uparrow)$}
[ .$\fz_{\omega'}^\downarrow$ {$(\text{\sout{$T_v$}},\fs_{\rm lex}^\downarrow)$} [ .{$(\fh^{\mathtt{v}^*},\fs_{\rm lex}^\uparrow)$} {$(\fh^{\mathtt{v}^*},\fz_{\mathtt{v}^*}^\uparrow)$} 
{$(\fz_{\mathtt{v}^*}^\downarrow, \fs_{\rm lex}^\uparrow)$} ] ] ] ]  ] ]
\end{equation} }
However, this now also creates a problem, because if $\omega'\in \Omega_{\fh,\phi}$ is a phase-head,
the IM movement of $T_v$ would be from the interior of a lower phase and violate the Phase Impenetrability
Condition, and if $\omega'$ is not a phase head, then the IM movement of the term 
$(\text{gave},\fh^{{\rm lex}}, \fz_{{\rm lex}}^\uparrow, \fs_{{\rm lex}}^\uparrow)$ performed by the
added generator would not be movement to Spec-of-Phase, hence by our coloring rules of Proposition
would have to be either $\omega'=\mathtt{INFL}$ (when it is not a phase-head) 
or $\omega'={\rm Root}$. It would have to be
$\omega'=\mathtt{INFL}$ given the position with respect to $\mathtt{v}^*$ in \eqref{new2Ob2Phi},
but this violates the EPP requirement of having a noun or determiner phrase in its specifier position,
as is the case for the $T_v$ term. Thus, the possibility $\omega'=\mathtt{INFL}$ is also not compatible 
with the structure of the Extended Projection and the movement of $T_v$. 

\smallskip

One could also consider changing the generators
of the colored operad $\cR_\Phi$ of phases, adding the possibility of generators where more
than one $\fs^\uparrow$ is flowing upward through the graph, something like
$$ \Tree[ .{$(\fh_{\omega'}, \fs_{\omega'}^\uparrow, \fs_\omega^\uparrow)$} 
[ {$(\fh_{\omega'}, \fz_{\omega'}^\uparrow, \fs_{\omega'}^\uparrow)$} {$(\fz_{\omega'}^\downarrow, \fs_{\omega}^\uparrow)$} ] ] \, . $$
However, this in turn would require a cascade of other changes and additions to the generator to 
still reflect the correct phase structure and allow for the discharge of a $\fs_{\omega'}^\downarrow$ for a higher
head $\omega'$, with $\theta_0$ position, 
before the $\fs_{\omega}^\downarrow =\fs_{\mathtt{T}}^\downarrow$ that receives the
IM movement originating from the external theta role position
$\theta_E^\downarrow$. It seems difficult to obtain a system of bud generator that would have these
properties and would still be compatible with the structure of the Extended Projection. So this does not
look like a viable option. 

\medskip

We can then consider the remaining possibility, namely the case of
a syntactic object of the form  \eqref{2ObjForm1}. According to the choice
of generators for $\cR_\Theta$ in \cite{MarLar}, the internal theta roles
are discharged above the Root position where they are injected, so that
we would again expect, as in the two other cases \eqref{2ObjForm2} and \eqref{2ObjForm3}, 
a structure of the form
\begin{equation}\label{thetaupflow}
 \Tree[ .$\theta_E^\uparrow$ [ $\theta_{I,2}^\downarrow$ [ .{$(\theta_E^\uparrow, \theta_{I,2}^\uparrow)$} $\theta_{I,1}^\downarrow$ {$(\theta_E^\uparrow, \theta_{I,1}^\uparrow, \theta_{I,2}^\uparrow)$} ] ] ] \ , . 
\end{equation} 
However, what we have in \eqref{2ObjForm1} would  instead
require a different coloring with generators that discharge the internal theta roles below the
point of insertion, resulting in a composition of two generators of the form
\begin{equation}\label{thetadownflow}
 \Tree[ .$\theta_E^\uparrow$ [ {$(\theta_E^\uparrow, \theta_{I,1}^\uparrow, \theta_{I,2}^\uparrow)$} [ .{$(\theta_{I,1}^\downarrow, \theta_{I,2}^\downarrow)$} $\theta_{I,1}^\downarrow$ $\theta_{I,2}^\downarrow$ ] ] ] 
\end{equation} 

Thus, a coloring by theta roles of the syntactic object  \eqref{2ObjForm1} would require the
introduction of additional generators in the bud generating system $\cR_\Theta$ that would allow for this
type of coloring as in \eqref{thetadownflow}. This would disfavor  \eqref{2ObjForm1}, with respect to
\eqref{2ObjForm2} and \eqref{2ObjForm3}, as accounting for  \eqref{2ObjForm1} would necessarily complicate 
the structure of the operad of theta roles. Moreover, among the three attested planarizations we considered,
the syntactic object \eqref{2ObjForm1} can only
give ``{\em Mary gave a book to John}" and ``{\em Mary gave (to) John a book}" but not the planarization
as in Mandarin, ``{\em Mary to John gave a book}".  Thus, once again, if only \eqref{2ObjForm1} is viable
before Externalization, we would have to obtain the remaining from from another syntactic object
obtained from \eqref{2ObjForm1}  via IM movement, and this reproduces the same problems
that we discussed in the case where only one of \eqref{2ObjForm2} and \eqref{2ObjForm3}
is viable prior to Externalization. Thus, this also disfavors the possibility of \eqref{2ObjForm1}. 

\smallskip

Thus, both the structure \eqref{2ObjForm1} and the structures obtained by movement 
in \eqref{2ObjForm1}, \eqref{2ObjForm2}, or \eqref{2ObjForm3} (required to achieve the remaining planar
embeddings known to occur if only one of \eqref{2ObjForm1},  \eqref{2ObjForm2}, \eqref{2ObjForm3} is
available) are disfavored with respect to the possibility of having {\em both} 
\eqref{2ObjForm2} and \eqref{2ObjForm3} prior to Externalization, hence having both possible
hierarchies 
$(\theta_{I,1},\theta_{I,2})=(\theta_{\mathtt{GO}}, \theta_{\mathtt{TH}} )$ and 
$(\theta_{I,1},\theta_{I,2})=(\theta_{\mathtt{TH}}, \theta_{\mathtt{GO}} )$ in the $\succeq$ 
preorder of theta-hierarchies. 
Indeed, in this case the simple generators for $\cR_\Theta$ and $\cR_\Phi$
suffice to account for both \eqref{2ObjForm2} and \eqref{2ObjForm3}, with
further selection between them occurring in Externalization. 
\endproof

Thus, we see
that considering both coloring by phases and by theta roles significantly
restricts the possibilities, compared to only considering one of the
two colorings, as in \cite{MarLar}.

\medskip

\section{Combined filtering for phases and theta roles}\label{ComboPhiThetaSec}

The specific cases discussed in the previous section suggest how one can combine
the two colored operads of phases and theta roles into a single filtering system, in the
form of a transduction of colored operads as introduced in Definition~\ref{optransduce}.

As in \cite{MarLar}, for a theta-hierarchy $\theta_{I,1}> \cdots > \theta_{I,n}$, and for $k\leq n$,
we use the notation 
$$ \underline{\theta}^\uparrow_k:= (\theta_E^\uparrow, \theta_{I,1}^\uparrow, \cdots, \theta_{I,k}^\uparrow) \, , $$
with $\theta_{I,i}^\uparrow$ are the internal theta roles and $\theta_E$ is the external theta role.  Thus,
$$ \underline{\theta}^\uparrow=\underline{\theta}^\uparrow_n=(\theta_E^\uparrow, \theta_{I,1}^\uparrow, \cdots, \theta_{I,n}^\uparrow)=(\theta_E^\uparrow, \underline{\theta}_I^\uparrow) $$
is the theta-grid assigned by a predicate. We also write $\theta_{I,j}^\downarrow$ for one of the theta roles
assigned by the predicate that is received by one of its arguments.  

Consider the set $\hat\Omega$ consisting of the colors
\begin{equation}\label{hatOmegaf}
\begin{array}{cccc}
(\fh^\omega, \fz^\uparrow_\omega, \fs^\uparrow_\omega, \underline{\theta}^\uparrow_k) &
(\fh^\omega, \fz^\uparrow_\omega, \fs^\uparrow_\omega, \theta_0) &
(\fh^\omega, \fz_\omega^\uparrow,  \underline{\theta}^\uparrow_k) & 
(\fh^\omega, \fz_\omega^\uparrow, \theta_0) \\
(\fs_\omega^\uparrow, \fz_{\omega'}^\downarrow, \underline{\theta}^\uparrow_k) &
(\fs_\omega^\uparrow, \fz_{\omega'}^\downarrow, \theta_0) & 
(\fh^\omega, \fs^\uparrow_\omega, \theta_E^\uparrow) & (\fh^\omega, \fs^\uparrow_\omega, \theta_0) \\
(\fh^{\omega'}, \fs_\omega^\uparrow, \underline{\theta}^\uparrow_k) & (\fh^{\omega'}, \fs_\omega^\uparrow, \theta_0) &
(\fz_\omega^\downarrow, \theta_{I,j}^\downarrow) & (\fz_\omega^\downarrow, \underline{\theta}^\uparrow_k) \\
(\fz_\omega^\downarrow, \theta_0) & (\fs^\downarrow_\omega, \theta_0) &
(\fs^\downarrow_\omega, \theta_E^\downarrow) & 
(\fm, \theta_0) \\ (\fm, \underline{\theta}^\uparrow_k) &
(\fm, \theta_{I,j}^\downarrow) & (1, \fm, \theta_0) & 
\end{array}
\end{equation}
with $\hat\Omega= \hat\Omega_f \sqcup \hat\Omega_{\rm lex}$, where
$\hat\Omega_f$ consists of colors as above with 
heads $\omega, \omega'\in \Omega_{\fh,f}$ and 
$\hat\Omega_{\rm lex}$ consists of the element $(1, \fm, \theta_0)$
and pairs of the form $(\alpha,\fm)$ of $(\alpha, \fc)$ with $\alpha \in \cS\cO_0$ and 
$\fc$ a color as above with $\omega \in \Omega_{\fh,{\rm lex}}$. 
There are two projection maps
$$ \pi: \hat\Omega \to \Omega_\Phi \ \ \ \text{ and } \ \ \  \pi': \hat\Omega \to \Omega_\Theta $$
where $\Omega_\Phi$ is the set of colors $\Omega_{ext}$ of Definition~\ref{modEPPlang}
of the bud generating system $\bB_\Phi$ of the colored operad of phases, and 
$\Omega_\Theta$ is the set of colors denoted by $\Theta$ in Definition~3.12 of \cite{MarLar},
of the bud generating system $\bB_\Theta$ of the colored operad of theta roles. The
projections $\pi, \pi'$ consist, respectively, of dropping the $\theta$ coordinates in the
colors of \eqref{hatOmegaf} or dropping the phase coordinates $\fh,\fs,\fz,\fm$. 
We set $\cT:=\pi^{-1}(\cT_\Phi)\cap \pi'^{-1}(\cT_\Theta)$ where we write $\cT_\Phi$
and $\cT_\Theta$ for the terminal colors of the bud generating systems $\bB_\Phi$
and $\bB_\Theta$, respectively. We take $\cI:= \hat\Omega \smallsetminus \cT$.

\begin{defn}\label{RPhiTheta}
Consider the colored operad $\bB_{\hat\Omega}(\cM)(n)=\hat\Omega \times \cM(n) \times \hat\Omega^n$
and all the elements of $\bB_{\hat\Omega}(\cM)(2)$ of the form
$$ T^\fc_{\fc',\fc''} := \Tree[ .$\fc$ [ $\fc'$ $\fc''$ ] ] $$
with $\fc,\fc',\fc''$ in $\hat\Omega$. We extend the projection maps to
$$ \pi:  \bB_{\hat\Omega}(\cM)(2) \to \bB_{\Omega_\Phi}(\cM)(2) \ \ \  \text{ and } \ \ \  \pi' : \bB_{\hat\Omega}(\cM)(2) \to \bB_{\Omega_\Theta}(\cM)(2) $$
by setting
$$ \pi (T^\fc_{\fc',\fc''}) := T^{\pi(\fc)}_{\pi(\fc'), \pi(\fc'')} =\Tree[ .$\pi(\fc)$ [ $\pi(\fc')$ $\pi(\fc'')$ ] ] $$
$$ \pi '(T^\fc_{\fc',\fc''}) := T^{\pi'(\fc)}_{\pi'(\fc'), \pi'(\fc'')} =\Tree[ .$\pi'(\fc)$ [ $\pi'(\fc')$ $\pi'(\fc'')$ ] ] $$
We define the subset $\cR_{\Phi\Theta}\subset \bB_{\hat\Omega}(\cM)(2)$ as the subset of
those generators $T^\fc_{\fc',\fc''}$ with the property that
\begin{equation}\label{piRPhiRTheta}
 \pi (T^\fc_{\fc',\fc''}) \in \cR_\Phi \ \ \ \ \text{ and } \ \ \ \  \pi' (T^\fc_{\fc',\fc''}) \in \cR_\Theta \, .  
\end{equation} 
\end{defn} 

We then obtain a proposal for a single filtering system that simultaneously implements
filtering by phases and by theta roles. 

\begin{prop}\label{transductionPhiTheta}
The bud generating system $\bB_{\Phi\Theta}=(\cM_h, \hat\Omega, \cR_{\Phi\Theta}, \cI, \cT)$ determines
a transduction of colored operads, as in Definition~\ref{optransduce} relating the colored operads
of phases and or theta roles. Let $\cL(\bB_{\Phi\Theta})\subset \bB_{\hat\Omega}(\cM)$ 
be the language of the bud generating system $\bB_{\Phi\Theta}$. We can also consider,
under the projections $\pi, \pi'$, the subsets
$\pi^{-1}(\cL(\bB_\Phi))\subset \bB_{\hat\Omega}(\cM)$ and $\pi'^{-1}(\cL(\bB_\Theta))\subset \bB_{\hat\Omega}(\cM)$.
The syntactic objects that are compatibly well formed with respect to both the 
filtering by phases and by theta roles are the elements of
$$ \pi^{-1}(\cL(\bB_\Phi)) \cap \pi'^{-1}(\cL(\bB_\Theta)) \, , $$
which in this case is the same as $\cL(\bB_{\Phi\Theta})$
All the cases discussed in \S \ref{ThetaPhaseMoveSec} of compatible coloring by phases
and theta roles are elements in this set. 
\end{prop}

\proof It suffices to check that we can indeed take $\pi^{-1}(\cL(\bB_\Phi)) \cap \pi'^{-1}(\cL(\bB_\Theta))$
instead of $\pi^{-1}(\cL(\bB_\Phi)) \cap \pi'^{-1}(\cL(\bB_\Theta))\cap \cL(\bB_{\Phi\Theta})$. The syntactic
objects in $\cL(\bB_{\Phi\Theta})$ are composed of generators in $\cR_{\Phi\Theta}$. Under the
projections $\pi,\pi'$ these generators to to generators in $\cR_\Phi$, respectively $\cR_\Theta$, so that
the image of $\cL(\bB_{\Phi\Theta})$ is contained in $\cL(\bB_\Phi)$, respectively $\cL(\bB_\Theta))$
and $\cL(\bB_{\Phi\Theta})\subset \pi^{-1}(\cL(\bB_\Phi)) \cap \pi'^{-1}(\cL(\bB_\Theta))$. If a syntactic
object is in $\pi^{-1}(\cL(\bB_\Phi))$, then it is obtaned from generators of the form  $T^\fc_{\fc',\fc''}$
with $T^{\pi(\fc)}_{\pi(\fc'),\pi(\fc'')}\in \cR_\Phi$ and $\fc=(\pi(\fc), \fc_\Theta)$, $\fc'=(\pi(\fc'), \fc'_\Theta)$,
$\fc''=(\pi(\fc''), \fc''_\Theta)$ with $\fc_\Theta, \fc'_\Theta, \fc''_\Theta$ colors in $\Omega_\Theta$, and
conversely for syntactic objects in $\pi'^{-1}(\cL(\bB_\Theta))$, so that, if a syntactic object is in 
$\pi^{-1}(\cL(\bB_\Phi)) \cap \pi'^{-1}(\cL(\bB_\Theta))$ it is in fact in $\cL(\bB_{\Phi\Theta})$. Thus, the
filtering by both phases and theta roles is achieved by intersecting the preimages of the two
filtering systems, namely the preimages of $\cL(\bB_\Phi)$ and $\cL(\bB_\Theta))$ in the common
space $\bB_{\hat\Omega}(\cM)$, or equivalently the result of this symultaneous filtering is given by 
$\cL(\bB_{\Phi\Theta})$. One can check by direct inspection that all the decompositions into generators
in $\cR_\Phi$ and $\cR_\Theta$ for the cases discussed in \S \ref{ThetaPhaseMoveSec} are
indeed in $\cL(\bB_{\Phi\Theta})$. 
\endproof

\bigskip
\bigskip

\subsection*{Acknowledgment} The first author is supported by NSF grant DMS-2104330,
by Caltech's Center of Evolutionary Science, and by Caltech's T\&C Chen Center for 
Systems Neuroscience.


\bigskip
\begin{thebibliography}{99}

\bibitem{BaJoRo} M.~Baker, K.~Johnson, I.~Roberts, {\em Passive Arguments Raised}, Linguistic. Inquiry 20 (1989), N.2, 219--251.

\bibitem{Chomsky82} N.~Chomsky, {\em Some concepts and consequences of the theory of government and binding}, 
MIT Press, 1982.

\bibitem{Chomsky-bare} N.~Chomsky. {\em Bare Phrase Structure}. In
``Evolution and Revolution in Linguistic Theory", 
edited by H.~Campos and P.~Kempchinsky, 
51--109. Georgetown University Press, 1995.

\bibitem{ChomskyPP} N.~Chomsky, {\em Problems of projection}, Lingua, Vol.130 (2013) 33-49.

\bibitem{ChomskyPPext} N.~Chomsky, {\em Problems of projection: Extension}, in ``Structures, Strategies and Beyond: Studies in honour of Adriana Belletti'' (E.~Di Domenico, C.~Hamann, S.~Matteini, Eds.) John Benjamins Publishing, 2015,  pp.~1--16

\bibitem{ChomskyGK} N.~Chomsky, {\em Minimalism: Where Are We Now, and Where Can We Hope to Go},  Gengo Kenkyu, 160 (2021), 1--41.

\bibitem{ChomskyGE} N.~Chomsky, {\em Genuine Explanation and the Strong Minimalist Thesis}, 
Cognitive Semantics, 8 (2022) 347--365.

\bibitem{Chomsky23} Noam Chomsky.  {\em Working toward the Strong Interpretation of SMT.}
Lecture Series Theoretical Linguistics at Keio-EMU, 2023. 

\bibitem{ChomskyMiracleCreed} Noam Chomsky,  {\em The Miracle Creed and the Strong Minimalist Thesis.}
In {\em A Cartesian Dream: A Geometrical Account of Syntax. In Honor of Andrea Moro,} 
(M.Greco and D.Mocci, Eds.) pp.~17--40. Research in Generative Grammar Monographs. 
Lingbuzz Press, 2023. 

\bibitem{ChomskyGE2} N.~Chomsky, {\em Genuine Explanation}, in: (G.Bocci, D.Botteri, C.Manetti, V.Moscati, Eds.), ``Rich Descriptions and Simple Explanations in Morphosyntax and Language Acquisition", Oxford University Press, 2024, pp.~15--44.

\bibitem{ChomskyElements} N.~Chomsky, T.D.~Seely, R.C.~Berwick, S.~Fong, M.A.C.~Huybregts, H.~Kitahara, A.~McInnerney, Y.~Sugimoto, {\em Merge and the Strong Minimalist Thesis}, Cambridge
Elements, Cambridge University Press, 2023.

\bibitem{CinRi} G.~Cinque, L.~Rizzi,  {\em The Cartography of Syntactic Structures}, 
STiL -- Studies in Linguistics, 2 (2008) 43--59.

\bibitem{Citko} B.~Citko, {\em Phase Theory}, Cambridge University Press, 2014.

\bibitem{Diac} P.~Diaconis, C.Y.A.~Pang, A.~Ram {\em Hopf algebras and Markov chains: Two examples and a theory}, 
J. Algebraic Combin. 39 (2014), no. 3, 527--585.

\bibitem{Giraudo} S.~Giraudo, {\em 
Colored operads, series on colored operads, and combinatorial generating systems}, 
Discrete Math. 342 (2019), no. 6, 1624--1657.

\bibitem{Golmo} M.~Golmohamadian, M.M.~Zahedi, {\em 
Color hypergroup and join space obtained by the vertex coloring of a graph}, 
Filomat 31 (2017), no. 20, 6501--6513.

\bibitem{Lars88} R.K.~Larson, {\em The double object construction}, LI 19, N. 3 (1988) 335--391. 

\bibitem{LarZha} R.K.~Larson \& C.~Zhang, {\em Applied objects in Mandarin and the nature of selection}, in
 ``New Explorations in Chinese Theoretical Syntax: Studies in honor of Yen-Hui Audrey Li",
Linguistik Aktuell/Linguistics Today,  272 (2022) 357--394. 

\bibitem{MCB} M.~Marcolli, N.~Chomsky, R.C.~Berwick, {\em Mathematical Structure of Syntactic Merge},
MIT Press, 2025.

\bibitem{MarLar} M.~Marcolli, R.K.~Larson, {\em Theta Theory: Operads and Coloring}, preprint, lingbuzz/008857, arXiv:2503.06091.

\bibitem{Marty} F. Marty, {\em R\^{o}le de la notion de hypergroupe dans l'\'etude de groupes non abeliens}, 
Comptes Rendus Acad. Sci. Paris 201, (1935), 636--638.

\bibitem{May} J.P.~May, {\it The geometry of iterated loop spaces}, Lecture Notes in Mathematics, Vol.~271, Springer, 1972.

\bibitem{Mit} J.~Mittas, {\em Hypergroupes canoniques}, Math. Balkanica 2 (1972), 165--179. 

\bibitem{NakaRey} S.~Nakamura, M.L.~Reyes, {\em Categories of hypermagmas, hypergroups, and
related structures}, arXiv:2304.09273.

\bibitem{Smith} J.D.H.~Smith, {\em Augmented quasigroups and character algebras}, Adv. Math. 363 (2020) 106983 [42 pages]

\bibitem{DYau}  D.~Yau, {\em Colored Operads}, 
Graduate Studies in Mathematics, 170. American Mathematical Society, 2016. 

\end{thebibliography}
\end{document}